\documentclass[10pt,journal,cspaper,compsoc]{IEEEtran}
\usepackage{epsfig}
\usepackage{graphicx}
\usepackage{amsmath}
\usepackage{amssymb}
\usepackage{array}
\usepackage{color}
\usepackage{multirow}
\usepackage{booktabs}
\usepackage{slashbox}
\usepackage{algorithm}
\usepackage{algorithmic}

\usepackage[pagebackref=false,breaklinks=true,letterpaper=true,colorlinks,citecolor=blue,linkcolor=blue,bookmarks=false]{hyperref}

\newcommand{\ignore}[1]{}
\ifCLASSOPTIONcompsoc
\else
\fi

\ifCLASSINFOpdf
\else
\fi

\begin{document}
%
\title{Learning to Deblur Images with Exemplars}

\author{Jinshan~Pan$^{*}$, Wenqi~Ren$^{*}$, Zhe~Hu$^{*}$,
        and~Ming-Hsuan~Yang
\IEEEcompsocitemizethanks{\IEEEcompsocthanksitem
J. Pan is with School of Computer Science and Engineering, Nanjing University of Science and Technology, Nanjing, China, 210094. E-mail: sdluran@gmail.com.
\IEEEcompsocthanksitem
W. Ren is with State Key Laboratory of Information Security, Institute of Information Engineering, Chinese Academy of Sciences, Beijing, China, 100093. E-mail: rwq.renwenqi@gmail.com.
\IEEEcompsocthanksitem Z.Hu is with Hikvision Research America, Santa Clara, CA, 95054. E-mail: zhe.hu.66@gmail.com.
\IEEEcompsocthanksitem M.-H. Yang are with School of Engineering, University of California, Merced, CA, 95344. E-mail: mhyang@ucmerced.edu.
\IEEEcompsocthanksitem $^*$The authors contributed equally to this work.
}
\thanks{}}

\markboth{IEEE Transactions on Pattern Analysis and Machine Intelligence}{}

\IEEEcompsoctitleabstractindextext{%
\begin{abstract}
Human faces are one interesting object class with numerous applications.
While significant progress has been made in the generic deblurring problem,
existing methods are less effective for blurry face images.
The success of the state-of-the-art image deblurring algorithms
stems mainly from implicit or explicit restoration of salient edges
for kernel estimation.
However, existing methods are less effective as only few edges can be restored
from blurry face images for kernel estimation.
In this paper, we address the problem of deblurring face images by exploiting facial structures.
We propose a deblurring algorithm
based on an exemplar dataset without using coarse-to-fine strategies or heuristic edge selections.
In addition, we develop a convolutional neural network to
restore sharp edges from blurry images for deblurring.
Extensive experiments against the state-of-the-art
methods demonstrate the effectiveness of the proposed
algorithms for deblurring face images.
In addition, we show the proposed algorithms can be applied to image deblurring for other object classes.
%
\end{abstract}

\begin{keywords}
Image deblurring, face image, exemplar-based, edge prediction, deep edge.
\end{keywords}}

\maketitle

\IEEEdisplaynotcompsoctitleabstractindextext

\IEEEpeerreviewmaketitle

\section{Introduction}
\label{sec: introduction}
\IEEEPARstart{T}{he}
goal of image deblurring is to recover the sharp contents and
corresponding blur kernel from one blurry input.
The image formation is usually formulated as
\begin{equation}
\label{eq: blur-model}
B = I*k + \varepsilon,
\end{equation}
where $B$ is the blurred input image,
$I$ is the latent sharp image, $k$ is the blur kernel,
$*$ is the convolution operator, and
$\varepsilon$ is the noise term.
The single image deblurring problem has attracted much attention with
significant advances in recent years~\cite{Fergus/et/al,Levin/CVPR2011,Shan/et/al, Cho/et/al, Xu/et/al,Joshi/et/al, Krishnan/CVPR2011, GoldsteinF12/eccv,Xu/l0deblur/cvpr2013}.
As image deblurring is an ill-posed problem, additional information is
required to constrain the solutions.
One common approach is to exploit statistical priors
of natural images such as heavy-tailed gradient
distributions~\cite{Fergus/et/al,Levin/CVPR2011,Shan/et/al,Levin/CVPR2009},
$L_1/L_2$ prior~\cite{Krishnan/CVPR2011}, and sparsity
constraints~\cite{Cai/tip12}.
While these priors have been shown to be effective for deblurring in general,
they are not designed to capture image properties for specific object
classes.
Recently, numerous methods that exploit specific properties have been developed for
text and low-light images~\cite{Hojin/cho/text/deblur,Jinshan/cvpr/text,cao/tip/text,hu/cvpr2014/lightstreak}.
As human faces are one of the most interesting objects that find
numerous applications, we mainly focus on face image deblurring in this work.
%

\begin{figure}[!t]\footnotesize
\begin{center}
\begin{tabular}{cccc}
\includegraphics[width=0.245\linewidth]{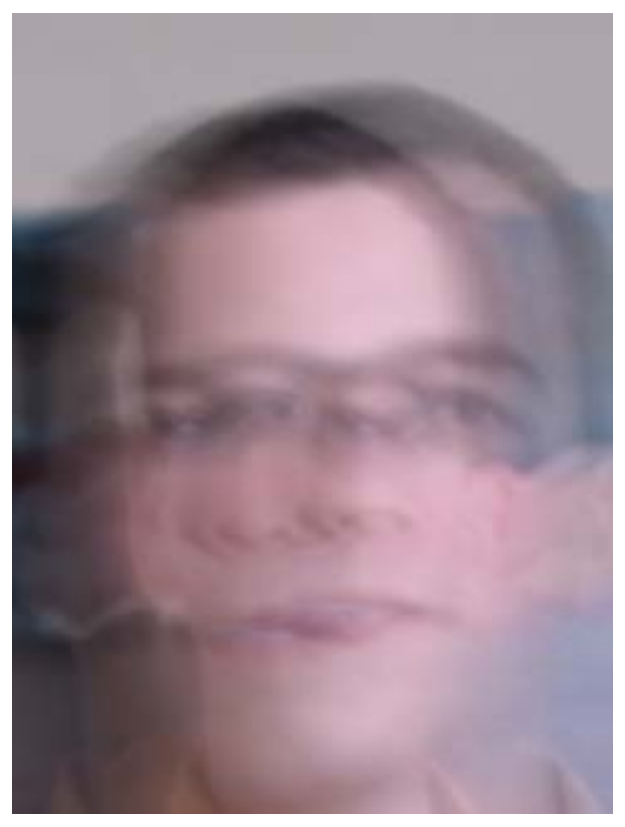} &\hspace{-0.52cm}
\includegraphics[width=0.245\linewidth]{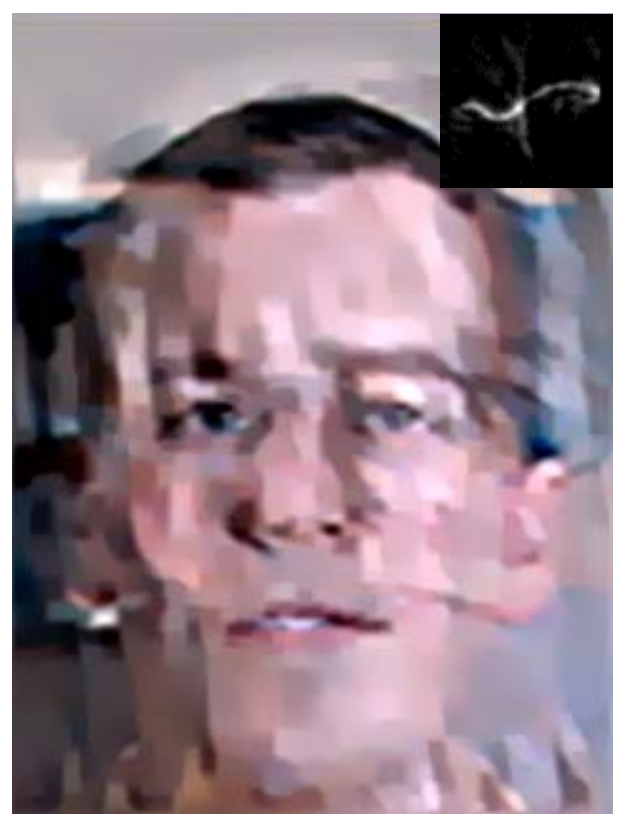} &\hspace{-0.52cm}
\includegraphics[width=0.245\linewidth]{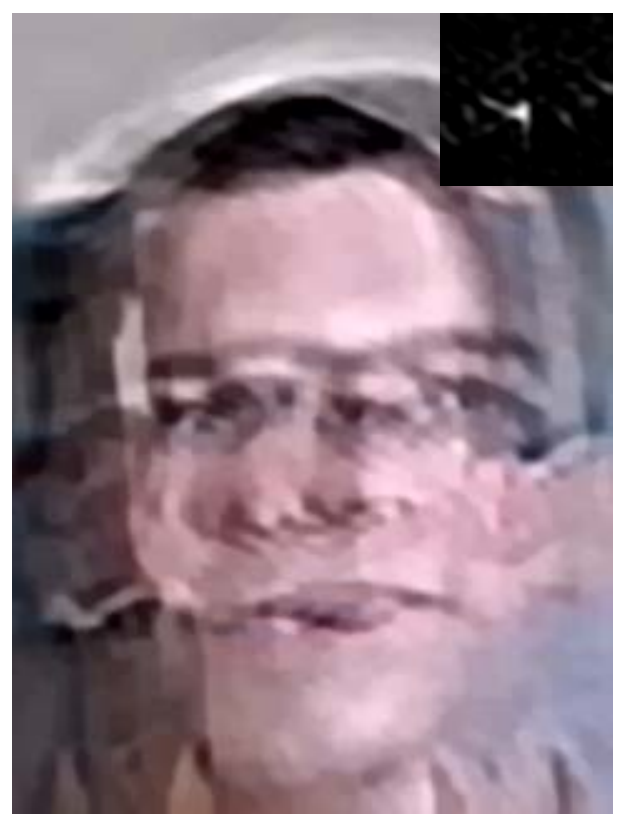} &\hspace{-0.52cm}
\includegraphics[width=0.245\linewidth]{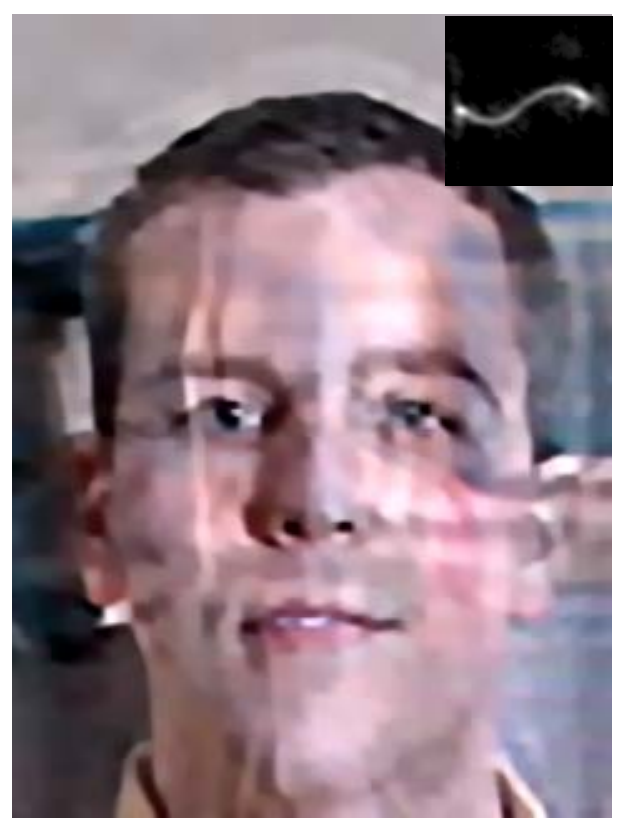} \\
(a) & \hspace{-0.52cm} (b) & \hspace{-0.52cm} (c)  &\hspace{-0.52cm} (d)\\
\includegraphics[width=0.245\linewidth]{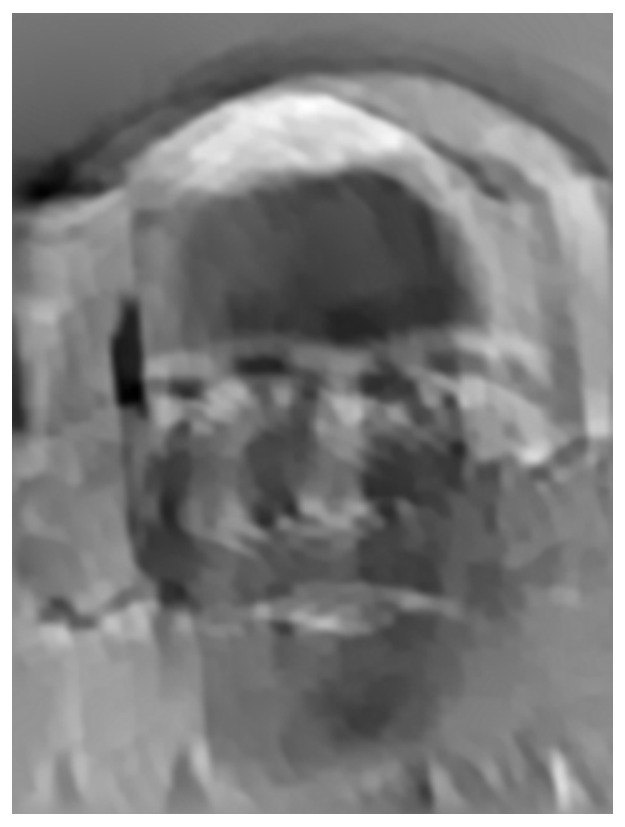} &\hspace{-0.52cm}
\includegraphics[width=0.245\linewidth]{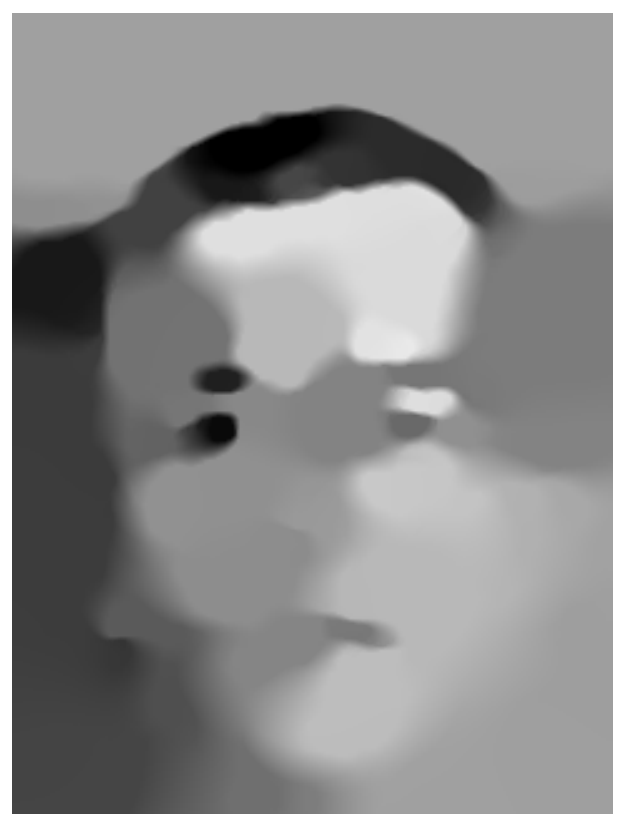} &\hspace{-0.52cm}
\includegraphics[width=0.245\linewidth]{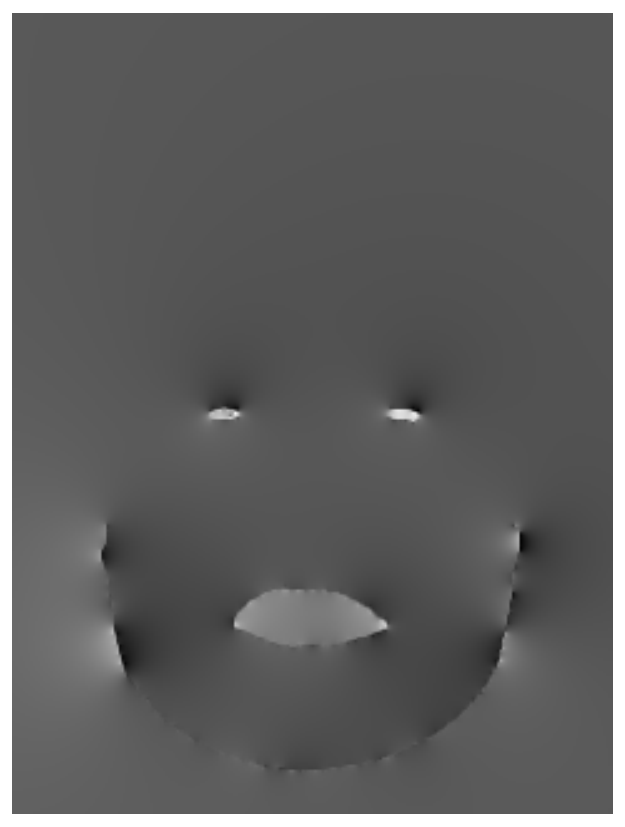} &\hspace{-0.52cm}
\includegraphics[width=0.245\linewidth]{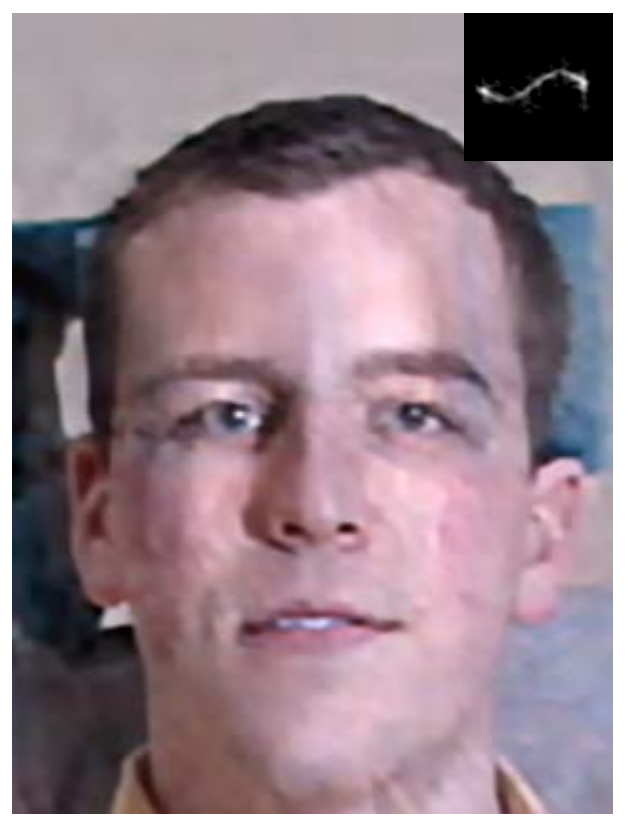} \\
(e) &\hspace{-0.52cm} (f) &\hspace{-0.52cm} (g) & \hspace{-0.52cm} (h)\\
\end{tabular}
\end{center}
\vspace{-0.3cm}
\caption{
A challenging example.
(a) Blurred face image.
(b)-(d) Results of Cho and Lee~\cite{Cho/et/al}, Krishnan~et al.~\cite{Krishnan/CVPR2011}, and Xu~et al.~\cite{Xu/l0deblur/cvpr2013}.
(e)-(f) Intermediate results of Krishnan~et al.~\cite{Krishnan/CVPR2011}
and Xu~et al.~\cite{Xu/l0deblur/cvpr2013}.
(g) Restored salient edges by our exemplar-based method visualized by Poisson reconstruction.
(h) Deblurred image by our method (with the support size of $75\times 75$ pixels).
}
\label{fig:figure1}
\end{figure}

The success of the state-of-the-art image deblurring methods hinges on
implicit or explicit restoration of salient edges
for kernel estimation~\cite{Cho/et/al,Xu/et/al,Joshi/et/al,Xu/l0deblur/cvpr2013}.
Existing algorithms predict sharp edges, mainly based on local image gradients
without considering the structural information of an object class.
Ambiguity inevitably arises in restoring salient edges when only local appearance is considered
due to the ill-posed image deblurring problem.
%
Furthermore, for blurred images without much texture, the edge prediction schemes
require parameter tuning and do not usually perform well.
%
For example, face images have similar components and skin complexion with less
texture than natural images, and existing deblurring methods do not perform well on such inputs.
Fig.~\ref{fig:figure1}(a) shows a blurry face image which
contains scarce texture as a result of large motion blur.
For such images, it is difficult to restore a sufficient number of
sharp edges for kernel estimation using the state-of-the-art methods.
Fig.~\ref{fig:figure1}(b) and (c) show
that the state-of-the-art methods based on sparsity
prior~\cite{Krishnan/CVPR2011} and explicit edge
prediction~\cite{Cho/et/al} do not deblur this image well.

In this work, we first propose an exemplar-based method
to address the above-mentioned issues
for deblurring face images.
To exploit the structural information from one specific class,
we collect an exemplar dataset and restore important visual information for kernel estimation.
For each test image, we use the exemplar with most similar facial structure to restore salient edges and guide the kernel estimation process.
Fig.~\ref{fig:figure1}(g) shows that
the proposed method is able to restore important facial
structures for kernel estimation, and
deblur this blurred image (Fig.~\ref{fig:figure1}(h)).

%

Predicting salient edges based on exemplars entails an effective similarity metric and
search in a large exemplar dataset, which is computationally expensive.
We further develop a deep convolutional neural network (CNN)
to restore salient edges from the blurred input.
The proposed CNN-based algorithm performs favorably against with the exemplar-based  method
and can be carried out in real-time.
In addition, we show that the proposed algorithm can be directly applied to deblur images
of other object classes.
%

\vspace{-3mm}
\section{Related Work}
\vspace{-1mm}
\label{sec: Related Work}
Image deblurring has been studied extensively in
computer vision and machine learning.
In this section we discuss the most relevant algorithms and put this
work in proper context.

{\noindent \bf Statistical Priors.} Since blind image deblurring is an ill-posed problem, it requires certain
assumptions or prior knowledge to constrain the solution space.
Early approaches, e.g.,~\cite{Yitzhaky/adirect},
assume simple parametric blur kernels to deblur images,
which cannot deal with complex motion blur.
%
As image gradients of natural images can be modeled well by a
heavy-tailed distribution, Fergus~et al.~\cite{Fergus/et/al} use
a mixture of Gaussians to learn the statistical prior for deblurring.
Similarly, Shan~et al.~\cite{Shan/et/al}
use a parametric model to approximate the heavy-tailed prior for
natural images.
In~\cite{Cai/tip12}, Cai~et al. assume
that the latent images and kernels can be sparsely represented by an
over-complete dictionary based on wavelets.
On the other hand, it has been shown that the most favorable solution
for a maximum a posteriori (MAP) deblurring method with sparsity prior
is usually a blurred image rather than a sharp one~\cite{Levin/CVPR2009}.
As~\cite{Levin/CVPR2009} is usually computationally expensive,
an efficient algorithm for approximation of marginal likelihood
is developed~\cite{Levin/CVPR2011} for image
deblurring.

%
{\noindent \bf Image Priors in Favor of Clear Images.}
Different image priors that favor clear images instead of blurred images have been introduced for image
deblurring.
Krishnan~et al.~\cite{Krishnan/CVPR2011}
present a normalized sparsity prior, and
Xu~et al.~\cite{Xu/l0deblur/cvpr2013} use the $L_0$ constraint on
image gradients for kernel estimation.
Non-parametric
patch priors that model edges and corners
have also been proposed~\cite{libin/sun/patchdeblur_iccp2013} for blur
kernel estimation.
%
We note that although the use of sparse priors facilitates
kernel estimation, it is likely to fail when the blurred images do not
contain rich texture.
%
In~\cite{tomer/eccv/MichaeliI14}, Michaeli and Irani exploit internal patch recurrence for image deblurring.
This method performs well when images contain repetitive patch patterns, but may fail otherwise.
%
Class-specific image prior~\cite{Anwariccv15/class/specific/deblurring} has been shown to be
effective for certain object categories and less effective for scenes with complex background.
%
Recently, Pan et al.~\cite{darkchannel/deblurring/cvpr16} develop an
image prior based on the dark channel prior~\cite{dark/channel/he/cvpr09} for blur kernel estimation.
However, this method does not perform well when clear images do not contain zero-intensity pixels
or the blurred images contain noise.

\begin{figure*}[!t]\footnotesize
\begin{center}
\begin{tabular}{ccccccc}
\includegraphics[width=0.13\linewidth]{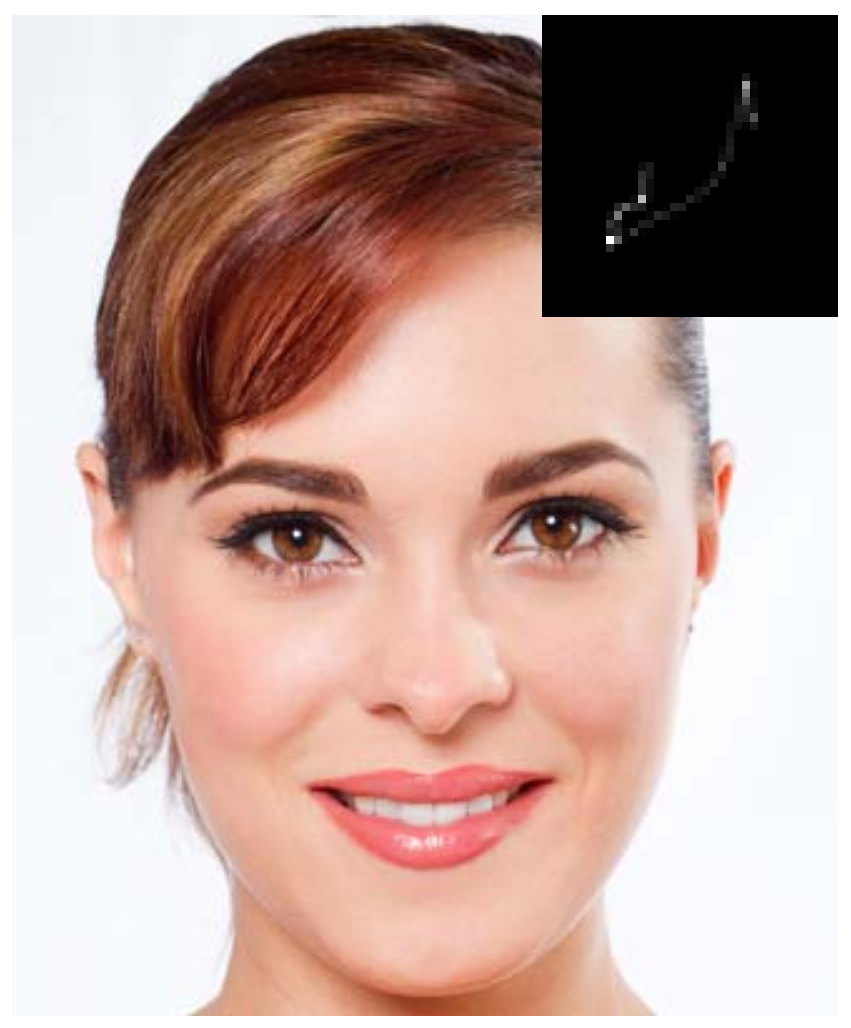} & \hspace{-0.52cm}
\includegraphics[width=0.13\linewidth]{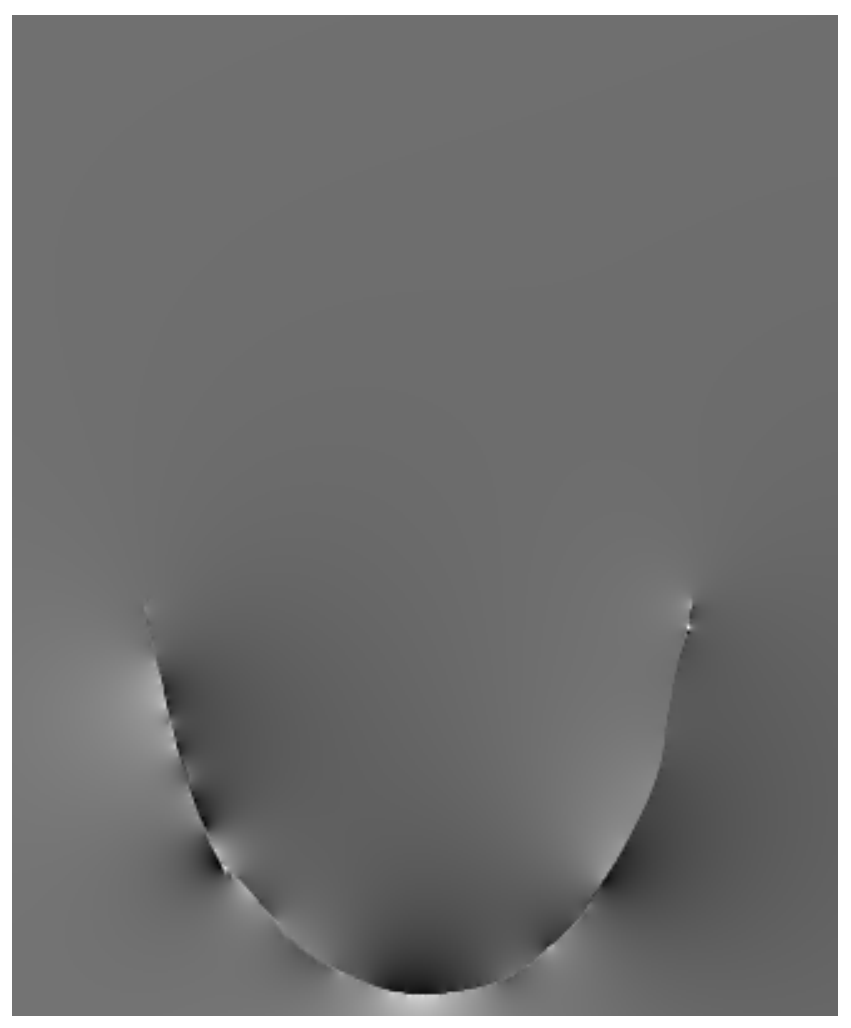} &\hspace{-0.52cm}
\includegraphics[width=0.13\linewidth]{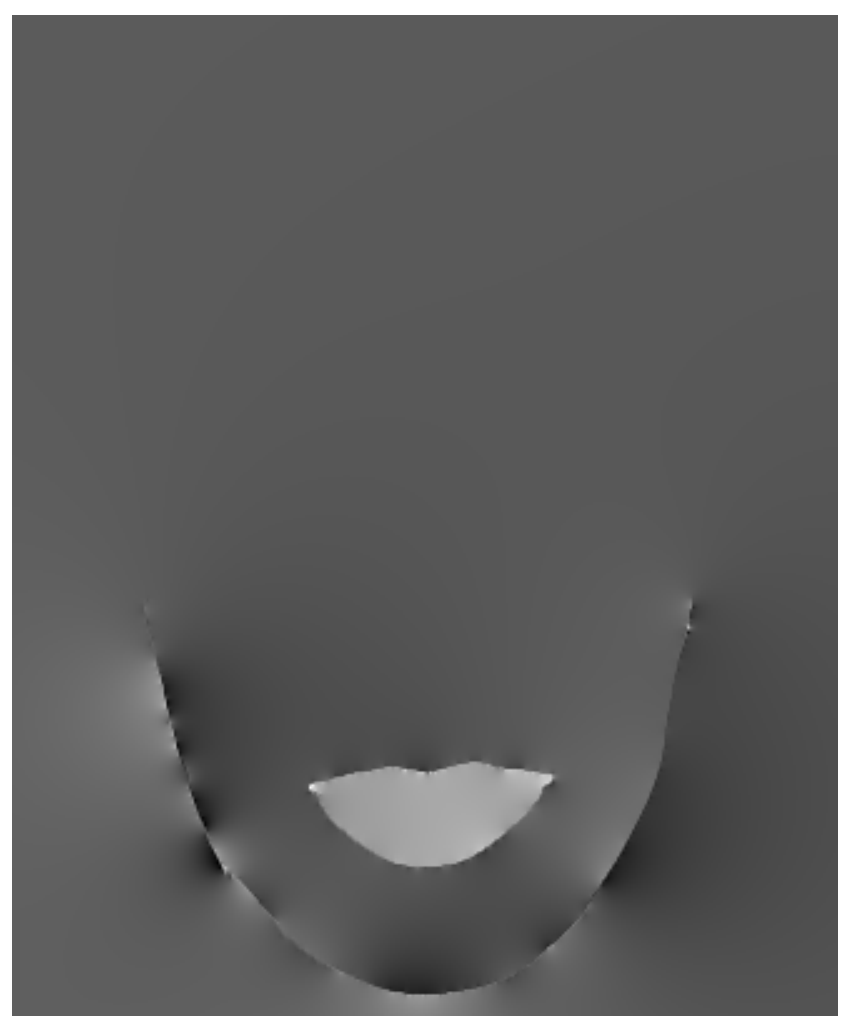} &\hspace{-0.52cm}
\includegraphics[width=0.13\linewidth]{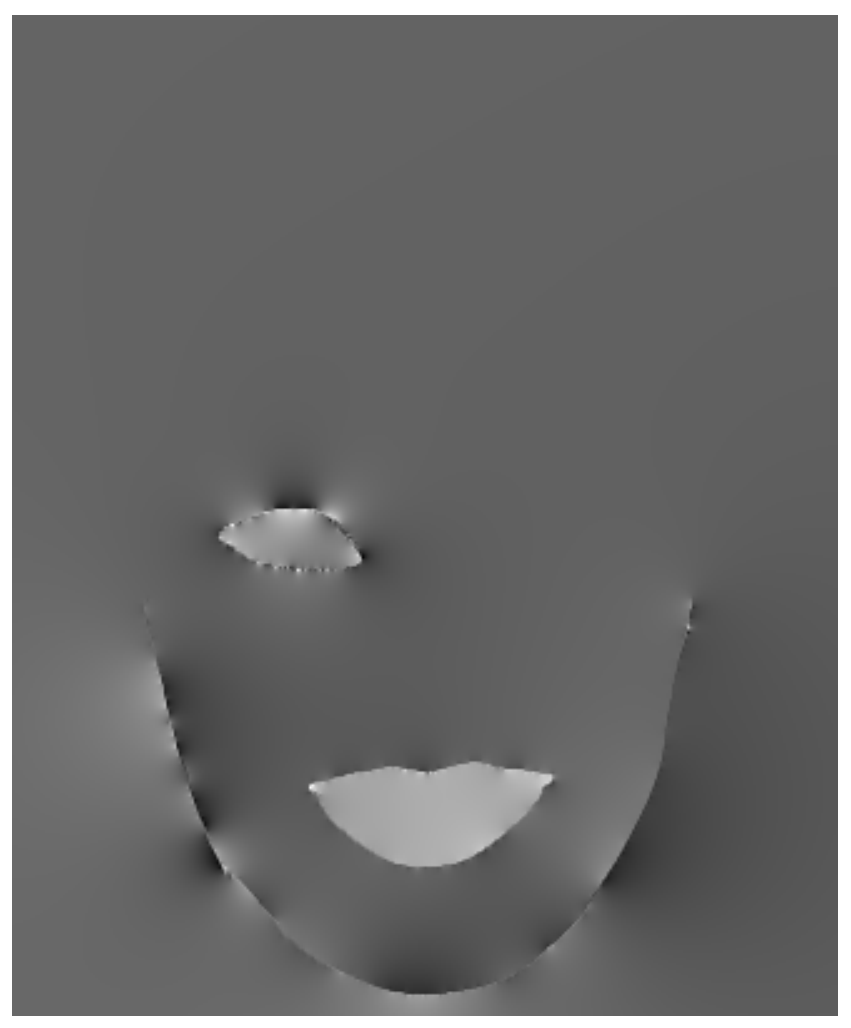} &\hspace{-0.52cm}
\includegraphics[width=0.13\linewidth]{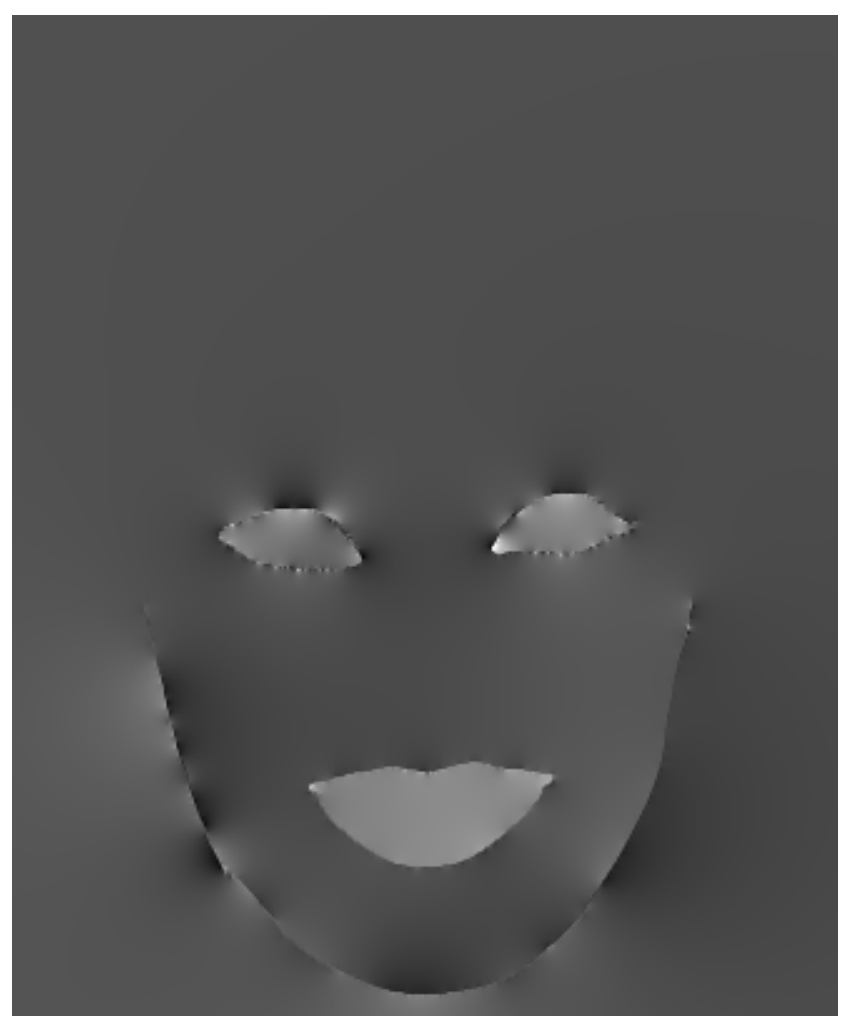} &\hspace{-0.52cm}
\includegraphics[width=0.13\linewidth]{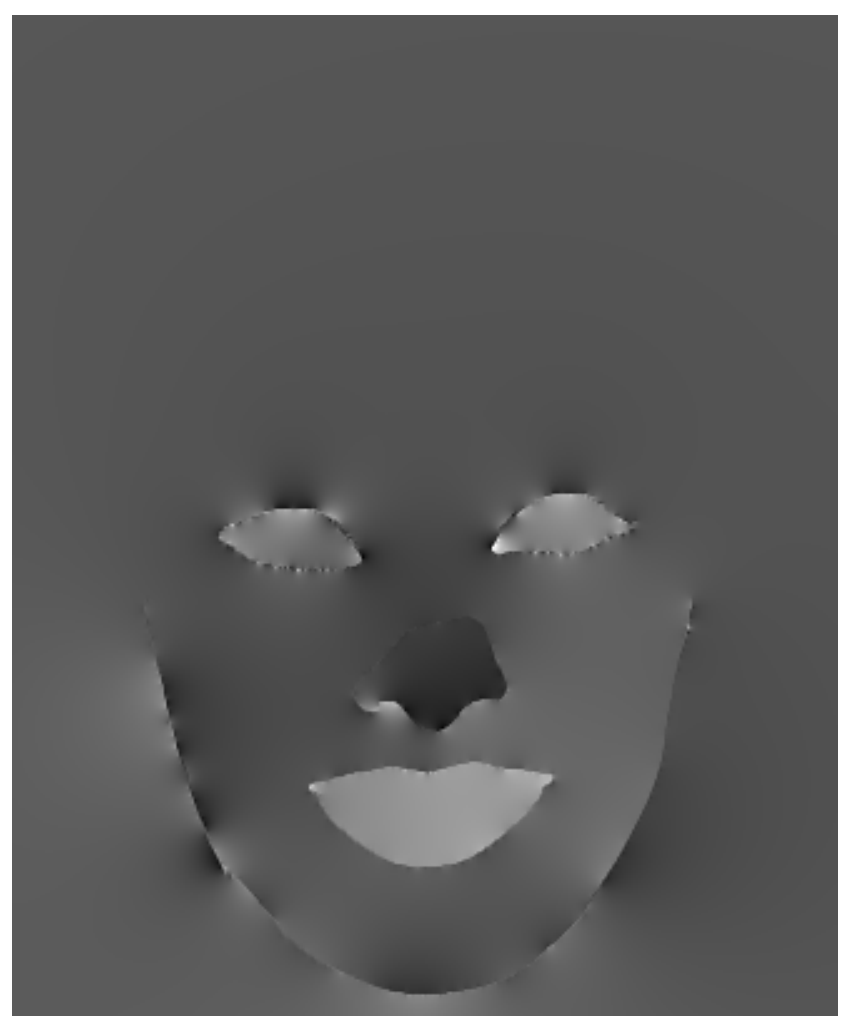} &\hspace{-0.52cm}
\includegraphics[width=0.13\linewidth]{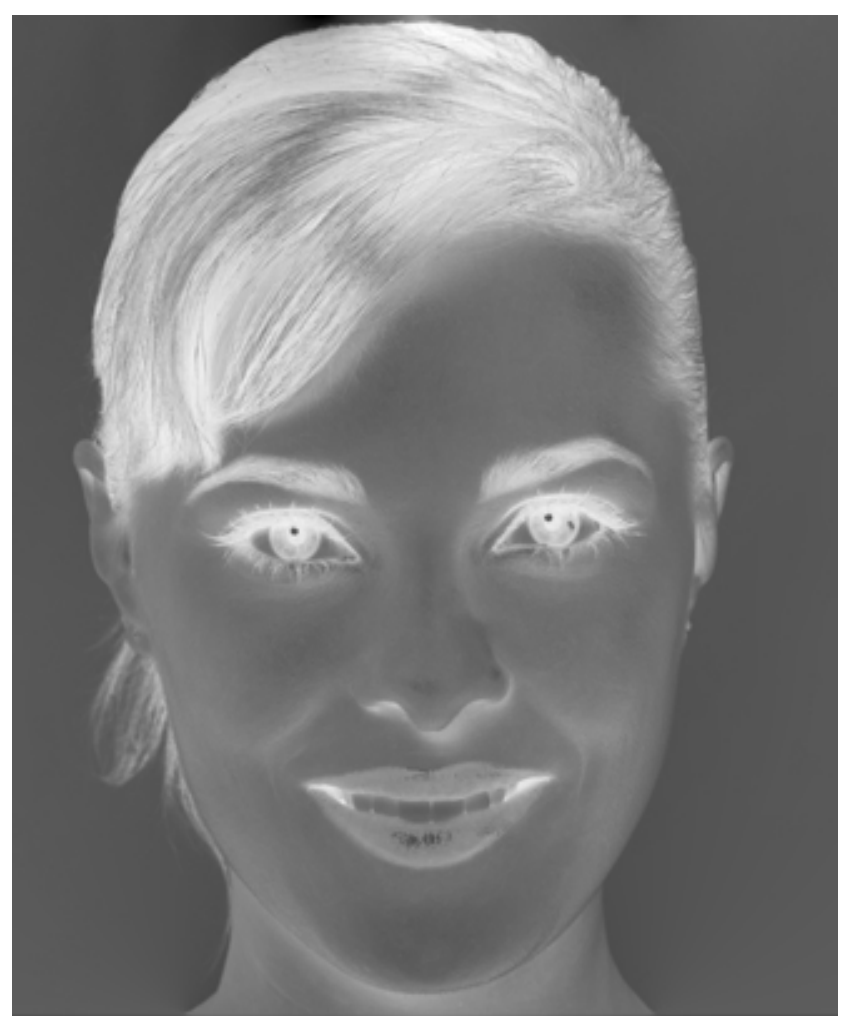} \\
(a) & (b) & (c)  & (d) & (e) & (f) & (g) \\
\includegraphics[width=0.13\linewidth]{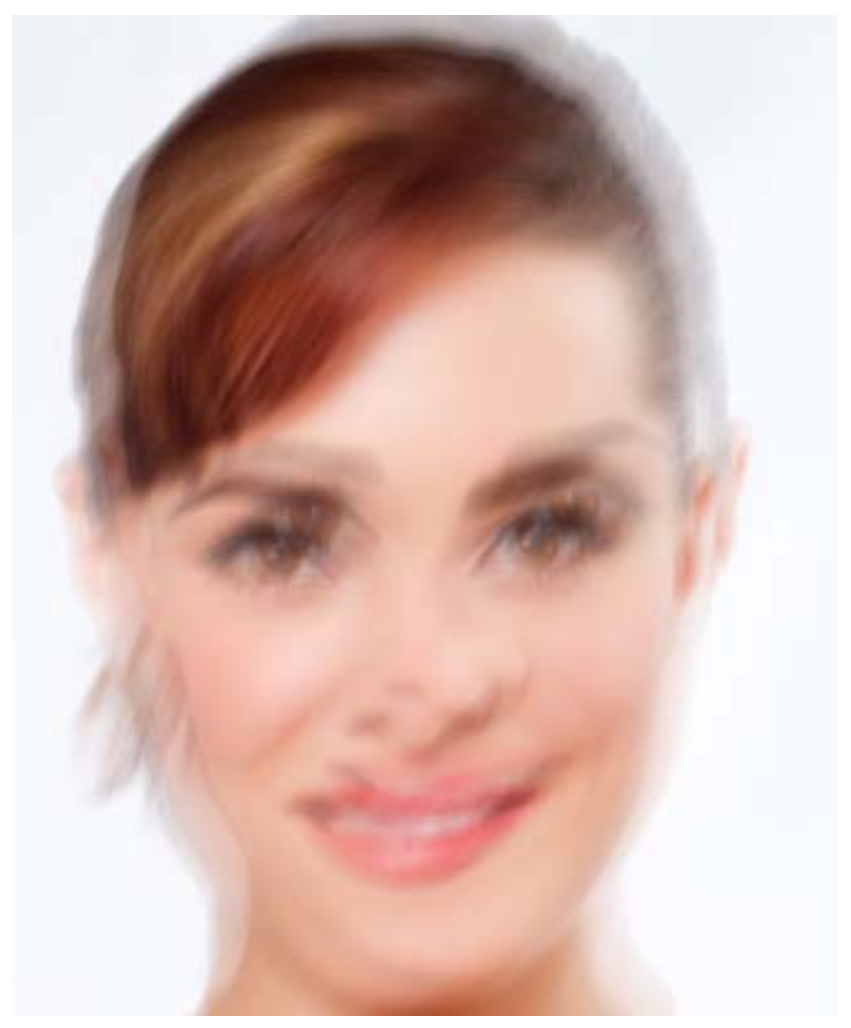} & \hspace{-0.52cm}
\includegraphics[width=0.13\linewidth]{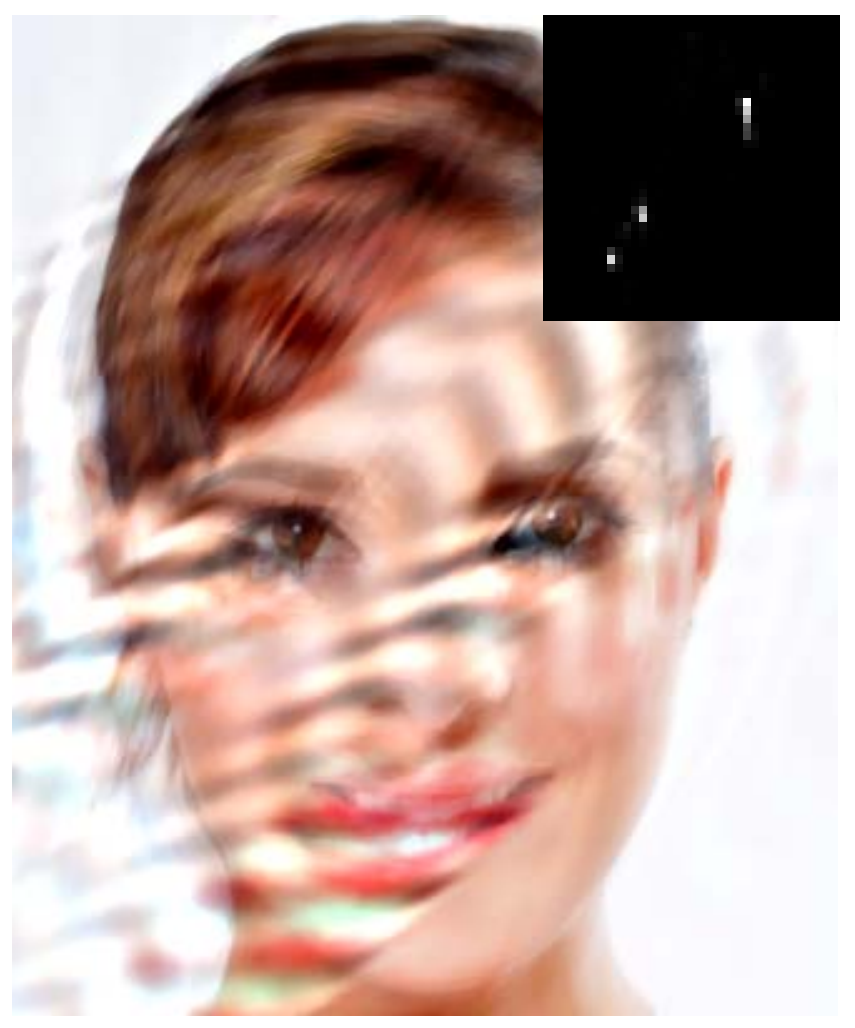} & \hspace{-0.52cm}
\includegraphics[width=0.13\linewidth]{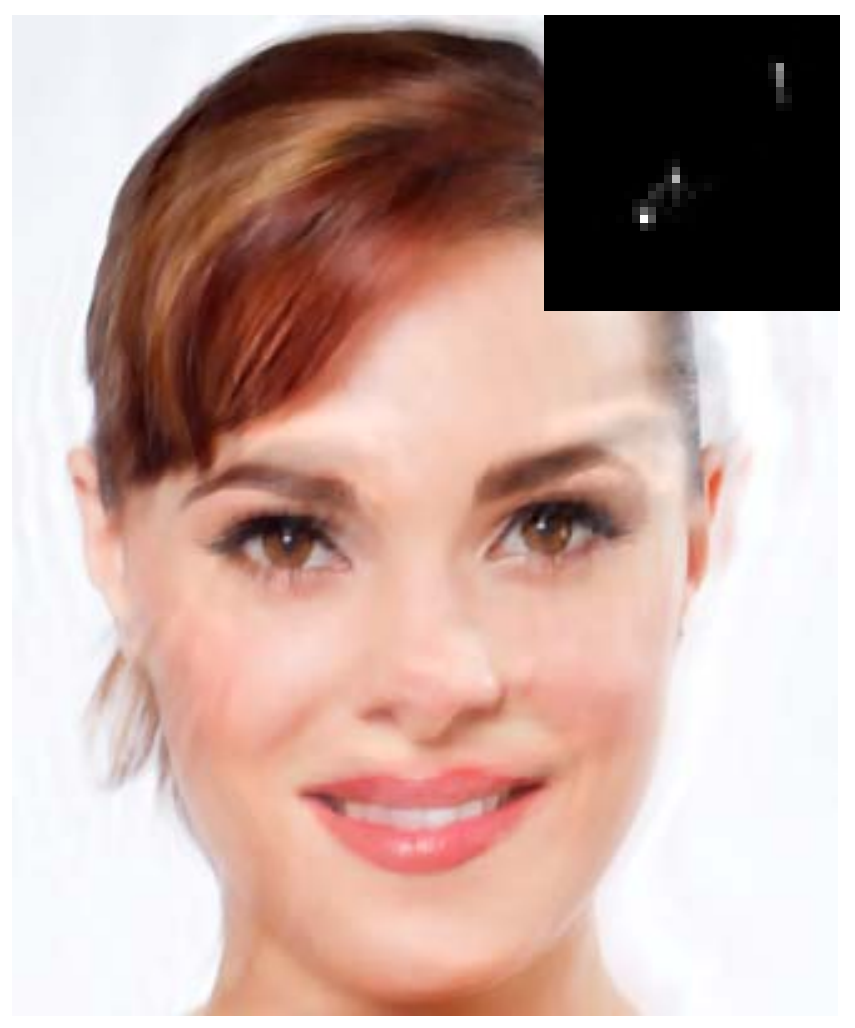} & \hspace{-0.52cm}
\includegraphics[width=0.13\linewidth]{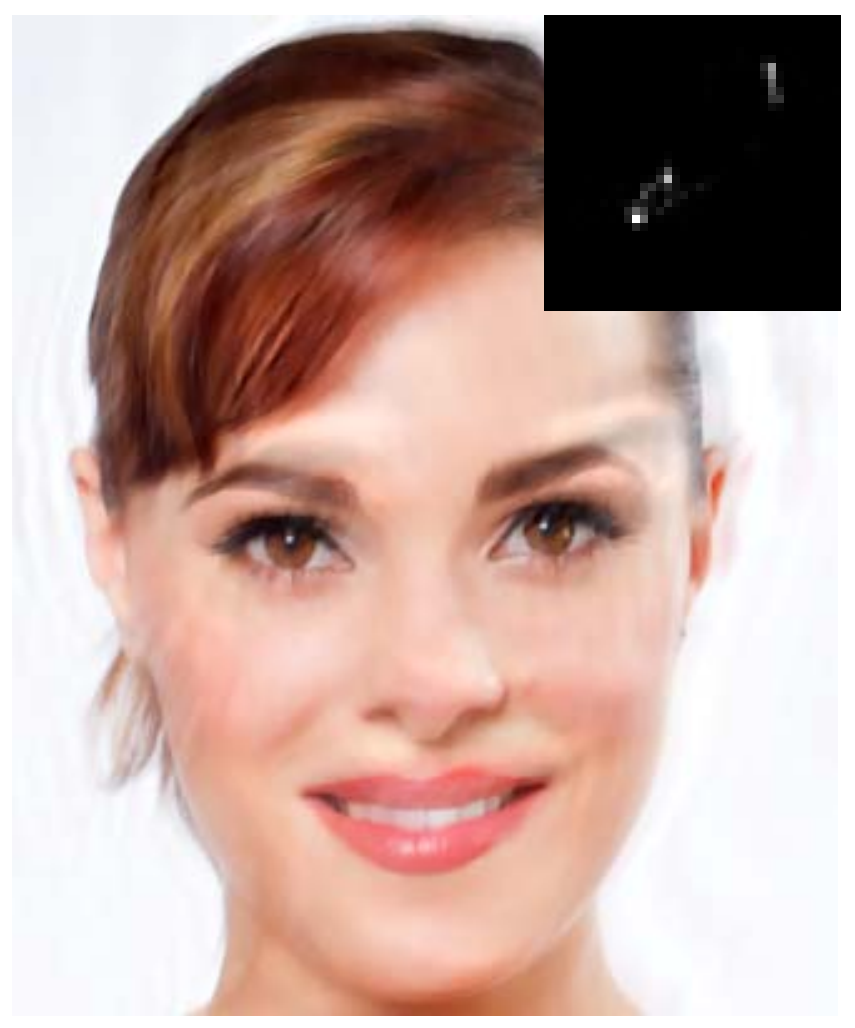} & \hspace{-0.52cm}
\includegraphics[width=0.13\linewidth]{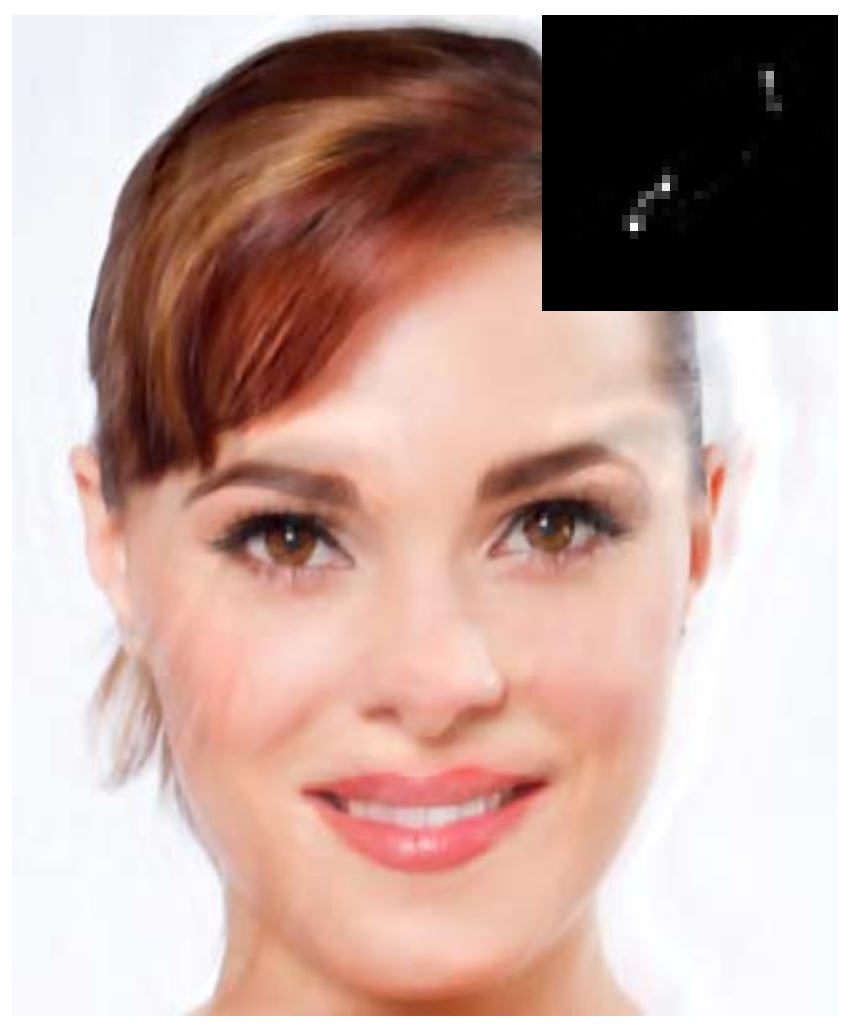} & \hspace{-0.52cm}
\includegraphics[width=0.13\linewidth]{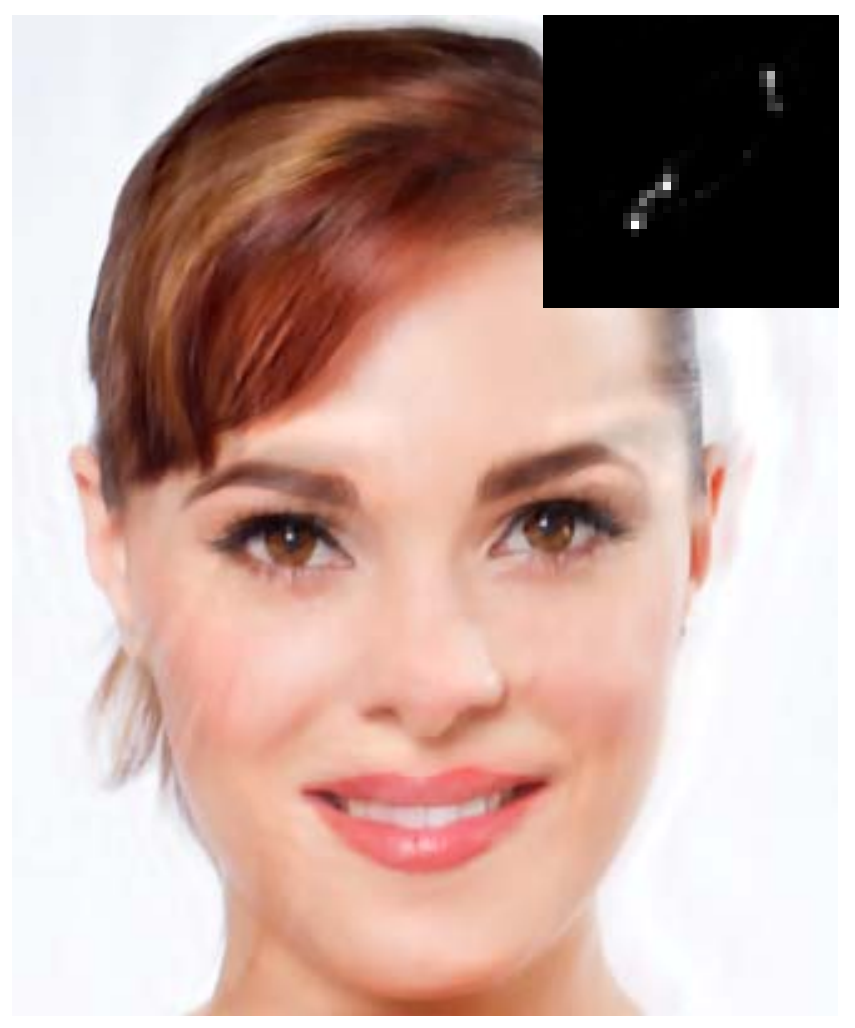} & \hspace{-0.52cm}
\includegraphics[width=0.13\linewidth]{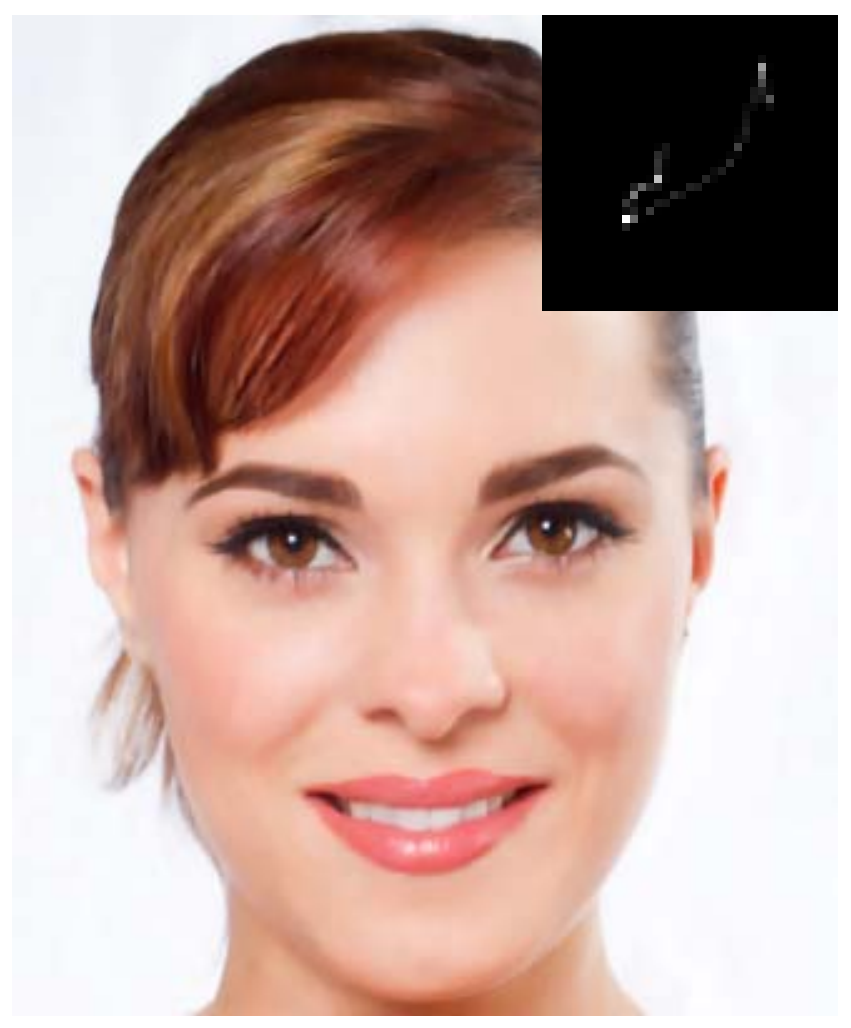} \\
 (h) & (i) & (j)  & (k) & (l) & (m) & (n) \\
\end{tabular}
\end{center}
\vspace{-0.3cm}
\caption{Effect of salient edges in kernel estimation.
(a) True image and kernel.
(h) Blurred image.
(b)-(f) Extracted salient edges of facial components
from the clear images visualized by Poisson reconstruction.
(g) The ground-truth edges of (a).
(i)-(n) Deblurred results by using edges (b)-(g), respectively.
}
\label{fig: a-number-salient-edges}
\end{figure*}

{\noindent \bf Edge Selection.}  In addition to statistical priors, numerous blind image deblurring
methods explicitly exploit edges for kernel
estimation~\cite{Cho/et/al,Xu/et/al,Joshi/et/al,cho/cvpr/radon}.
Joshi~et al.~\cite{Joshi/et/al} and Cho~et al.~\cite{cho/cvpr/radon}
use the restored sharp edges from a blurred image for kernel estimation.
In~\cite{Cho/et/al}, Cho and Lee utilize bilateral and shock filters
to predict sharp edges.
The blur kernel is determined by alternating between restoring
sharp edges and estimating blur kernels in a coarse-to-fine
manner.
As strong edges restored from a blurred image are not necessarily
useful for kernel estimation, Xu and Jia~\cite{Xu/et/al} develop a method to
select informative ones for deblurring.
Despite demonstrated success, these
methods rely largely on image filtering methods (e.g.,
shock and bilateral filters) and heuristics for restoring sharp edges,
which are less effective for objects with specific geometric structures.

{\noindent \bf Face Deblurring.}
A few algorithms have been developed to deblur face images
for the recognition task.
Nishiyama~et al.~\cite{face/deblur/pami/NishiyamaHTSKY11} learn
subspaces from blurred face images with known blur kernels for
recognition.
As the set of blur kernels is pre-defined, the application domain of
this approach is limited.
Zhang~et al.~\cite{zhang11/et/al/iccv} propose a joint image
restoration and recognition method based on sparse representations.
However, this method is most effective for well cropped and aligned face images
with simple motion blurs.

{\noindent \bf Example-based Deblurring.} Recently, HaCohen et al.~\cite{HaCohen/iccv13} propose a deblurring method which uses sharp reference examples for guidance.
The method requires a reference image with the same contents as the input
to obtain dense correspondence for reconstruction.
Although it has been shown to deblur specific images well,
the assumption of using reference images with same contents limit its application domain.
%
In contrast, the proposed methods do not require the exemplar to have the same
or closely similar contents of the input.
The blurred face image can be of different identity and background
when compared to exemplar images.
The proposed methods only require the matched example to have similar
structures (in terms of image gradients)
for kernel estimation instead of using dense corresponding pixels.
As such, the proposed algorithms can be applied to class specific image deblurring with
fewer constraints.
%

%

{\noindent \bf Convolutional Neural Networks.}  Convolutional neural networks
have been widely used in low-level vision tasks
including image denoising~\cite{jain2009natural}, super-resolution~\cite{dong2014learning,wang2015deep},
non-blind deconvolution~\cite{sun2015learning,xu2014deep}, blind image deblurring~\cite{SchulerPAMI15} and image filtering~\cite{Xu/ICML15/deepedge,ren2015shepard}.
Schuler~et al.~\cite{SchulerPAMI15} incorporate a sharpening convolutional neural network
into an iterative blind deconvolution method to estimate the blur kernel.
However, this method needs to re-train different networks for kernels of
different sizes, which limits the application domains.
In~\cite{Xu/ICML15/deepedge}, Xu~et al. propose a method to learn
edge-aware filters using a deep convolutional neural network.
However, we note that this method can
only be applied to approximate edge-aware filters for clear images.
This method cannot be directly
applied to restore salient edges from blurry images for kernel estimation.
%

\vspace{-3mm}
\section{Proposed Algorithms}
\vspace{-1mm}
\label{sec: Learning Good Structures for Kernel Estimation}
As the kernel estimation problem is
non-convex~\cite{Fergus/et/al,Levin/CVPR2011},
most state-of-the-art deblurring methods use coarse-to-fine
approaches to refine the results.
Furthermore, explicit or implicit edge selection schemes are adopted to
constrain and converge to feasible solutions.
Notwithstanding the demonstrated success in deblurring images, these
methods are less effective for face images that contain fewer textured contents.
To address these issues, we first propose an exemplar-based algorithm
to estimate blur kernels for face images.
The proposed method restores important structural information from
exemplars to facilitate accurate kernel estimation.
%
%
To reduce the computational cost,
we further propose a CNN-based algorithm which can predict sharp edges
more effectively than the exemplar-based method.

\begin{figure}[t]\footnotesize
\begin{center}
\begin{tabular}{ccc}
\includegraphics[width = 0.32\linewidth]{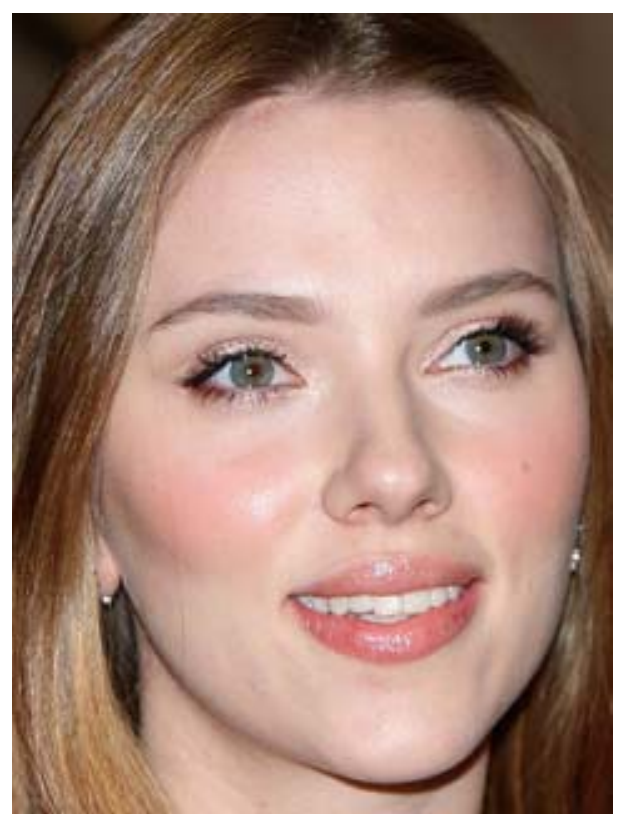} & \hspace{-0.52cm}
\includegraphics[width=0.32\linewidth]{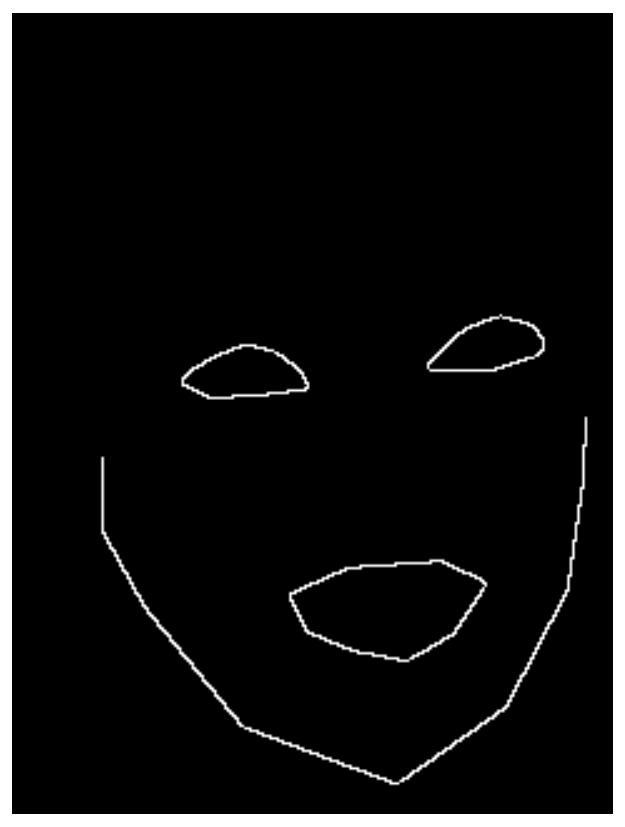} & \hspace{-0.52cm}
\includegraphics[width=0.32\linewidth]{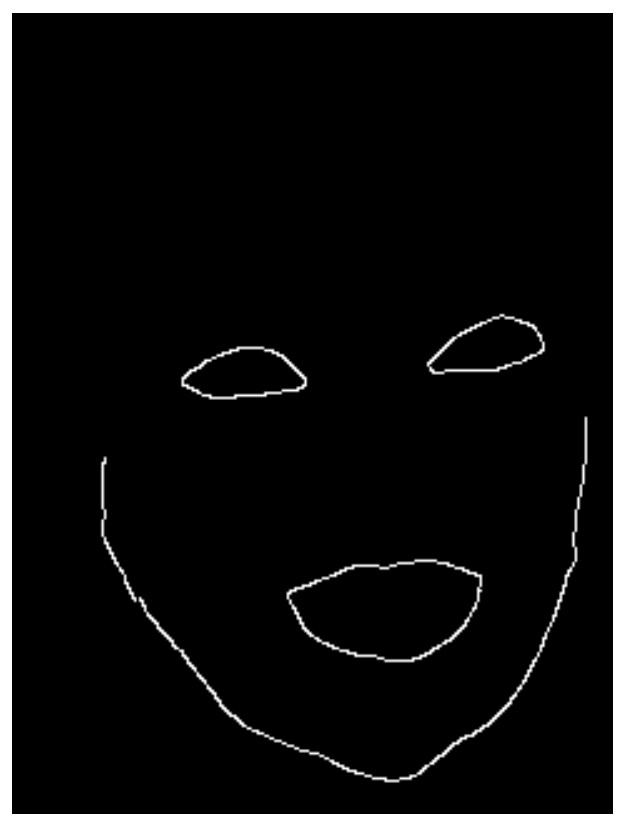} \\
(a)  &\hspace{-0.52cm} (b)  &\hspace{-0.52cm} (c) \\
\end{tabular}
\end{center}
\vspace{-0.3cm}
\caption{Extracted salient edges (see Section~\ref{ssec: Learning Framework} for details).
(a) Input image.
(b) Initial contour.
(c) Refined contour.}
\label{fig: edges-mask}
\end{figure}

\begin{figure}[!t]\footnotesize
\begin{center}
\begin{tabular}{c}
\includegraphics[width=0.65\linewidth]{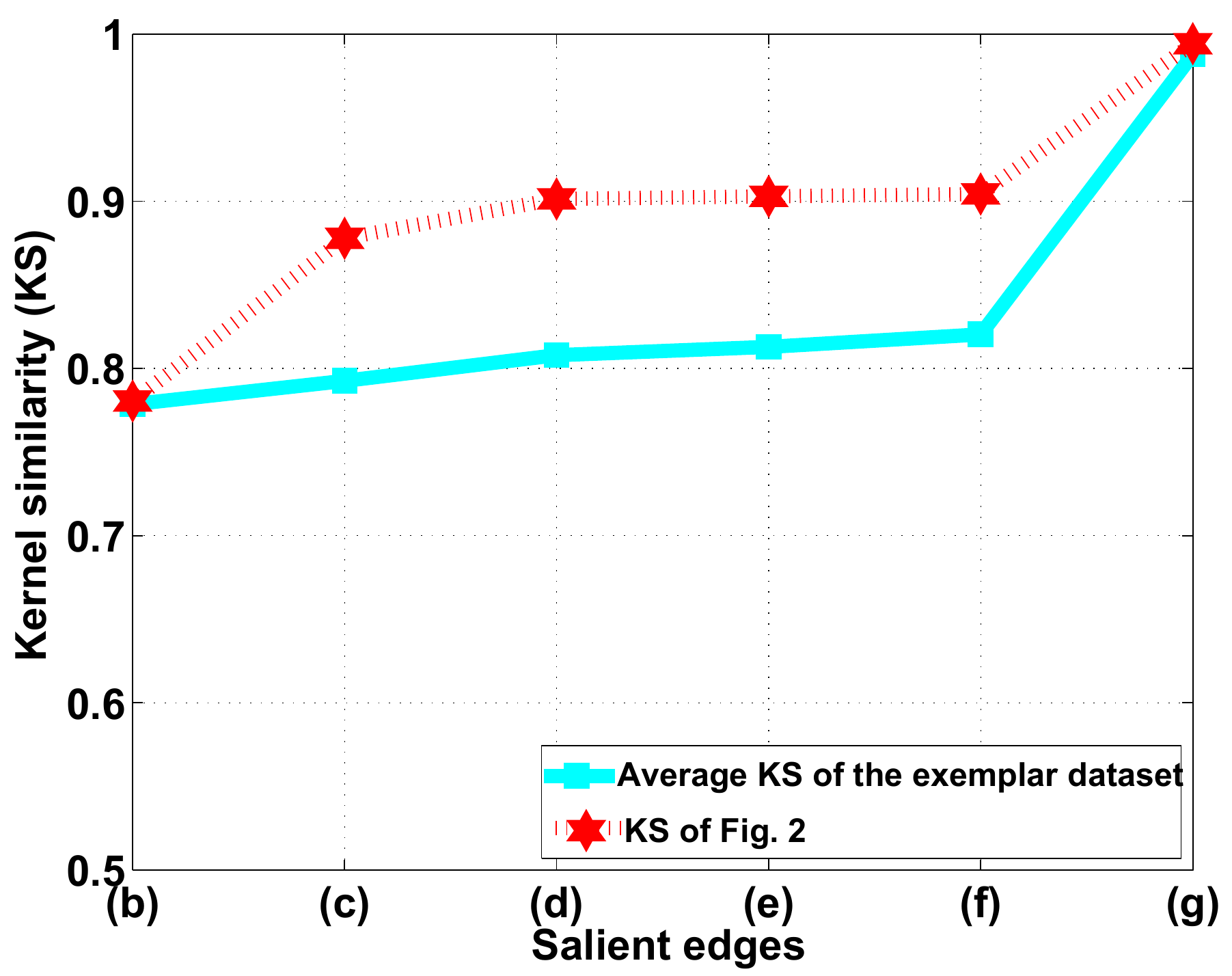} \\
\end{tabular}
\end{center}
\vspace{-0.4cm}
\caption{
Kernel estimation accuracy (KS stands for kernel similarity) with respect to
restored salient edges from different facial components.
The $x$-axis (b)-(g) represent 6 facial components
in Fig.~\ref{fig: a-number-salient-edges}(b)-(g).
}
\label{fig:onecol-1}
\end{figure}

\vspace{-4mm}
\subsection{Structure of Face Images}
\vspace{-1mm}
\label{ssec: Salient Edges in Kernel Estimation}
We first determine the types and number of salient edges from exemplars
for kernel estimation within the context of face deblurring.
For face images, the salient edges that capture the object structure
may be the lower contour, mouth, eyes, nose, eyebrows and hair.
As eyebrows and hair have small edges with large variations which
may be less effective for kernel estimation~\cite{Xu/et/al,Huzhe/eccv2012},
we do not consider them as useful structures.
Fig.~\ref{fig: a-number-salient-edges} shows
several components restored from a clear face image as
approximations of the latent image for kernel estimation.

To extract salient edges as shown in Fig.~\ref{fig: a-number-salient-edges}(b)-(g),
we manually locate the initial contours of the informative components
(Fig.~\ref{fig: edges-mask}(b)), and use the guided
filter~\cite{Kaiming/eccv10/guidedfilter} for refinement.
The optimal threshold, computed by the Otsu method~\cite{otsu}, is
applied to each filtered image to obtain the refined binary
contour mask~$\mathcal{M}$ of the facial components (Fig.~\ref{fig: edges-mask}(c)).
As such, the salient edge is defined by
\begin{equation}
\label{eq: definition-salient-edge}
\nabla S(x) = \left\{\begin{array}{ll} \nabla T(x), & \mbox{if } x \in \{x| \mathcal{M}(x) = 1\},\\0, & \mbox{}\ $otherwise$, \end{array}\right.
\end{equation}
where $T(x)$ is the clear image and $\nabla$ is the gradient operator.
We use the horizontal $[-1, 1]$ and vertical $[-1, 1]^{\top}$ derivatives to compute image gradients.

%
We evaluate these edges by considering them as the predicted salient edges
in the deblurring framework and estimate the blur kernels
according to~\cite{Levin/CVPR2011} by
\begin{equation}
\label{eq: general-kernel-estimation-model}
k^* = \arg\min_{k} \| \nabla S*k - \nabla B\|_2^2 + \alpha\|k\|^{0.5},
\end{equation}
where $\nabla S$ is the gradient of the salient edges restored
from an exemplar image
%
as shown in Fig.~\ref{fig:  a-number-salient-edges}(b)-(g),
$\nabla B$ is the gradient computed from the blurred input
(Fig.~\ref{fig: a-number-salient-edges}(h)),
$k$ is the blur kernel, and $\alpha$ is a weight (e.g., 0.005 in this
work) for the regularization term.
%
The sparse deconvolution method~\cite{Levin/CVPR2011} with a
hyper-Laplacian prior $L_{0.8}$ is employed to restore latent images
(Fig.~\ref{fig: a-number-salient-edges}(i)-(n)).
The deblurred results using the above-mentioned components
(e.g., Fig.~\ref{fig: a-number-salient-edges}(l) and (m)),
are comparable to that using the ground-truth edges (Fig.~\ref{fig: a-number-salient-edges}(n)),
which provide the ideal case for salient edge prediction of the input blurry image.

To support the above-mentioned observations,
we collect a set of 160 images generated from 20 images (10 images from the CMU PIE
dataset~\cite{Multi/PIE/datasets} and 10 images from the Internet)
convolving with 8 blur kernels,
and restore the corresponding edges from different combinations of
components (i.e., Fig.~\ref{fig: a-number-salient-edges}(b)-(g)).
We conduct the same experiment as Fig.~\ref{fig: a-number-salient-edges},
and compute the average accuracy of the estimated kernels in terms of kernel similarity~\cite{Huzhe/eccv2012}.
The red dashed curve in Fig.~\ref{fig:onecol-1} shows the relationship
between the edges of facial components and accuracy of estimated kernels.
As shown in the figure, the metric converges when all the mentioned components
(e.g., Fig.~\ref{fig: a-number-salient-edges}(e)) are included,
and the set of edges is sufficient (kernel similarity value of $0.9$ in Fig.~\ref{fig:onecol-1}) for accurate kernel estimation.

For real-world applications, the ground-truth edges are not available.
Recent methods adopt thresholding and similar techniques to select salient
edges for kernel estimation and this inevitably introduces some incorrect edges from a blurred image.
Furthermore, the edge selection strategies, either explicitly or
implicitly, consider only local edges rather than structural
information of a particular object class, e.g., facial components and
contour.
In contrast, we consider important geometric structures of a
face image for kernel estimation.
From the experiments with different facial components, we
determine that the set of lower face contour, mouth and eyes is sufficient
to achieve accurate kernel estimation
and deblurred results.
More importantly, these components can also be robustly
restored~\cite{Facedection/CVPR12/Zhu} unlike the other parts (e.g.,
eyebrows or nose in Fig.~\ref{fig: a-number-salient-edges}(a)).
Thus, we use these three components as the informative
structures for face image deblurring.

\vspace{-4mm}
\subsection{Structure Prediction}
\label{ssec: Structure Prediction}
Based on above discussions, we propose two structure prediction methods
for blur kernel estimation.
%

\vspace{-2mm}
\subsubsection{Structure Prediction by Exemplars}
\label{ssec: Learning Framework}
We use a set of $2,435$ face images from the CMU PIE
dataset~\cite{Multi/PIE/datasets} as our exemplars for deblurring.
The selected face images are from different identities with
varying facial expressions and poses.
For each exemplar, we restore the informative structures (i.e., lower face contour, eyes and
mouth) as discussed in Section~\ref{ssec: Salient Edges in
  Kernel Estimation}.
%
%
%
As such, a set of $2,435$ exemplar structures are generated as the potential facial structure
for kernel estimation.

Given a blurred image $B$, we search for its best matched
exemplar structure.
We use the maximum response of normalized cross-correlation to
find the best candidate based on image gradients,
\begin{equation}
\label{eq: normalized-cross-correlation}
v_i = \max_t\left\{\frac{\sum_x \nabla B(x) \nabla T_i(x+t)}{\|\nabla
    B(x)\|_2\|\nabla T_i(x + t)\|_2}\right\},
\end{equation}
where $i$ is the index of the exemplar, $T_i(x)$ is the $i$-th exemplar, and $t$ is the possible shift
between image gradients $\nabla B(x)$ and $\nabla T_i(x)$.
The value of $v_i$ is large if $\nabla B(x)$ is similar to $\nabla T_i(x)$.
%
To deal with face images of different scales,
we resize each exemplar with sampled scaling factors in
the range [1/2, 2]
at a sampling step size of 0.5.
before using~\eqref{eq: normalized-cross-correlation}.
%
Similarly,  we rotate each exemplar with the rotation angle in [-10, 10] degree before
using~\eqref{eq: normalized-cross-correlation} to deal with rotated face images,
where the sampling step size is 1.

The predicted salient edges $\nabla S$  for kernel estimation is defined by
\begin{equation}
\label{eq: final-structure}
\nabla S = \nabla S_{i^*},
\end{equation}
where $i^* = \text{arg}\max_i{v_i}$, and $\nabla S_{i^*}(x)$ is computed by
\begin{equation}
\label{eq: structure-component}
\nabla S_{i^*}(x) = \left\{\begin{array}{ll} \nabla T_{i^*}(x), & \mbox{if } x \in \{x| \mathcal{M}_{i^*}(x) = 1\},\\0, & \mbox{}\ $otherwise$. \end{array}\right.
\end{equation}
Here $\mathcal{M}_{i^*}$ is the contour mask for ${i^*}$-th exemplar.
In the experiments, we find that the method
using the edges of exemplars $\nabla  T_{i^*}(x)$ as
the predicted salient edges performs similarly as that of the input image $\nabla B(x)$,
(see Section~\ref{sec: Experimental Results}).
The reason is that $\nabla  T_{i^*}(x)$ and $\nabla B(x)$ share similar structures
due to the matching step, and
thus the results using either of them as the guidance are similar.

\begin{figure}[!t]\footnotesize
\begin{center}
\begin{tabular}{cc}
\includegraphics[width = 0.49\linewidth]{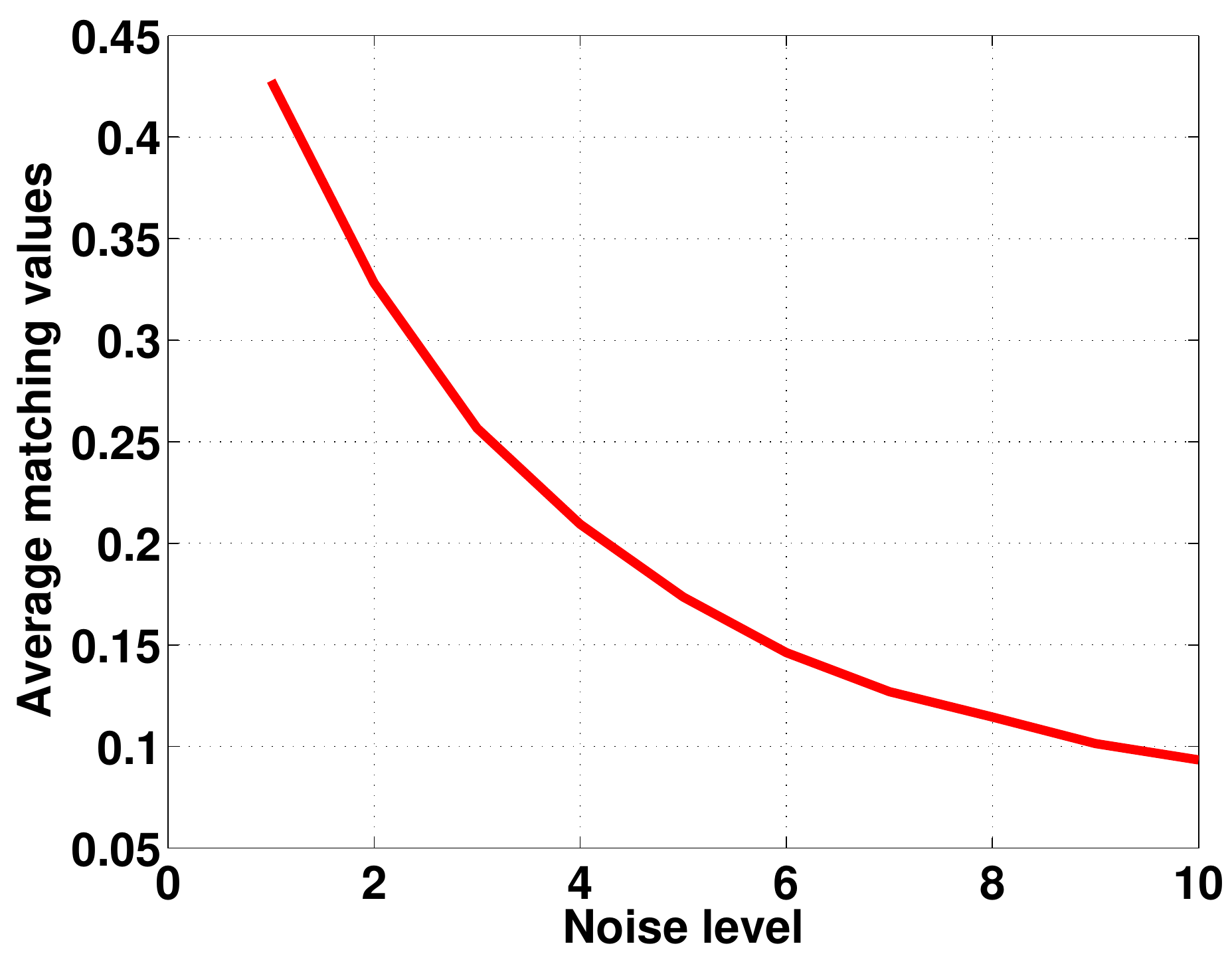} & \hspace{-0.52cm}
\includegraphics[width = 0.49\linewidth]{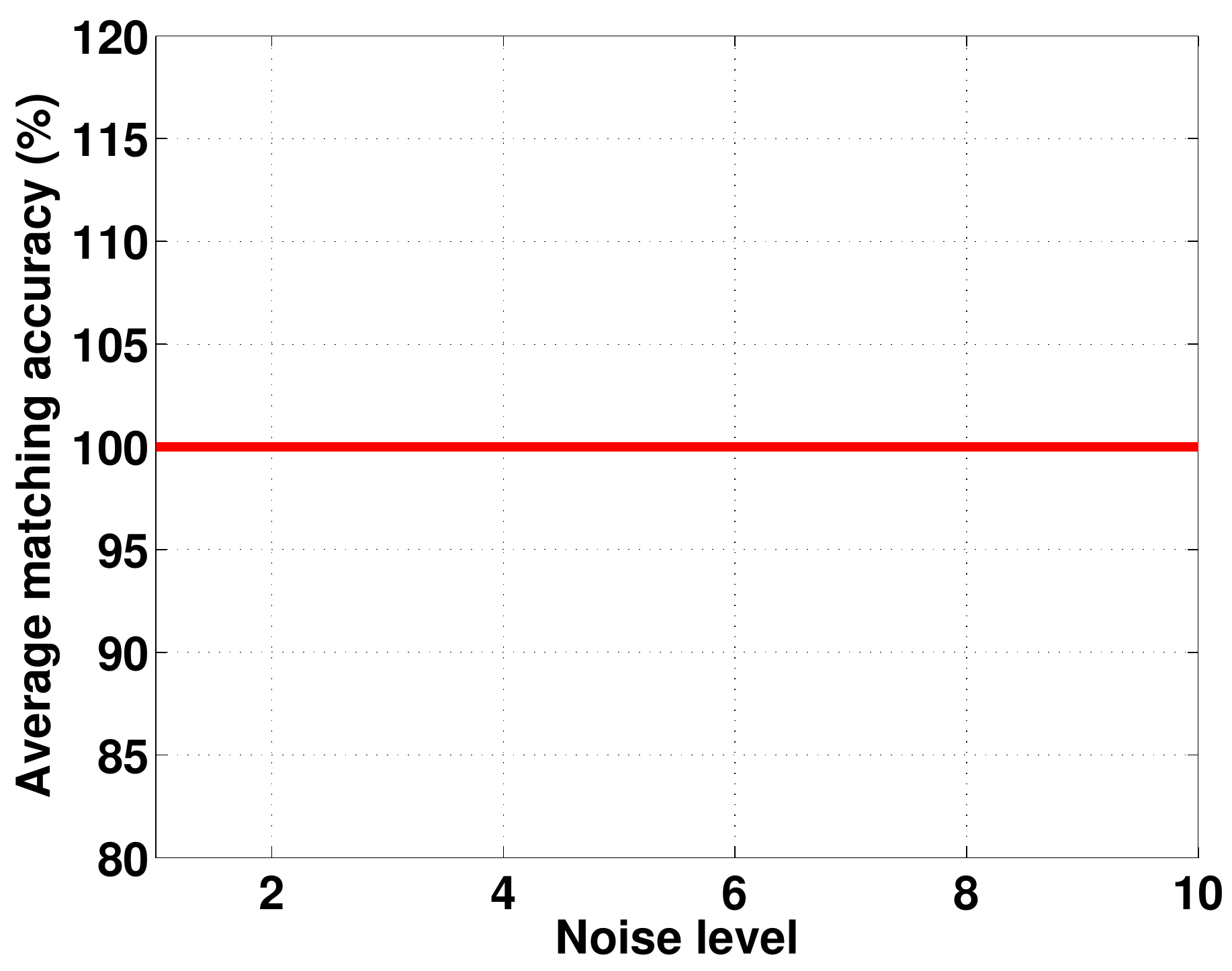} \\
 (a) &\hspace{-0.52cm} (b) \\
\end{tabular}
\end{center}
\vspace{-0.3cm}
\caption{Effect of noise on the proposed matching criterion.}
\label{fig: noise-influence-match}
\end{figure}

We conduct experiments with the quantitative evaluations to
demonstrate the effectiveness and robustness of our matching criterion.
We collect 100 clear images from 50 identities, with 2 images for each.
The images from the same person are different in terms of facial expression and background.
In the test phase, we blur one image with random noise as the test image,
and use the others as exemplars.
If the matched exemplar is the image from the same person, we
consider that as a success.
We evaluate each images with 8 blur kernels and 11 noise levels (0-10\%)
and show the matching accuracy in Fig.~\ref{fig: noise-influence-match}(b).
We note that although noise decreases the average matching values
(see Fig.~\ref{fig: noise-influence-match}(a)),
it does not affect the matching accuracy (Fig.~\ref{fig: noise-influence-match}(b)).
%
%

\vspace{-2mm}
\subsubsection{Structure Prediction by a Deep CNN}
\vspace{-1mm}
\label{sec: Deep Learning Framework}
Although the above-mentioned structure prediction method can effectively predict $\nabla S$,
searching for the best matched exemplar in the large dataset is computationally expensive.
In this section, we propose an approach to predict structure information from a
blurry face image based on a CNN, which has similar effect for edge prediction
but achieves 3,000 times acceleration against the exemplar-based method
(see Section~\ref{sec: Experimental Results}).

%



\begin{table}[t]
\vspace{-3mm}
	\caption{CNN architecture for structure prediction.}
	\centering
\vspace{-3mm}
\begin{tabular}{ccccccc}
		\toprule
		Layer & 1 & 2 & 3 & 4 & 5 & 6\\
		\midrule
		Filter size & $15\times15$ & $1\times1$ & $1\times1$ & $1\times1$ & $1\times1$ & $1\times1$\\
		Channel     & 64 & 64 & 64 & 64 & 64 & 1 \\
		\bottomrule
	\end{tabular}
	\label{tab: architecture}
\end{table}

\vspace{-2mm}
{\flushleft \textbf{Proposed Network.}} Given a blurred face image, our goal is to predict the salient structure
by a CNN, where each layer contains convolution operations
followed by non-linear activations.
The network architecture and parameters are shown in Table~\ref{tab: architecture}.
%
%
We assume that the input blurred face image $B$ is of size $p \times q \times 1$ for 1 gray channel, where $p \times q$ is the spatial resolution.
In the structure prediction network, the first convolution layer
takes a large filter ($15 \times 15$) to capture large spatial information.
The subsequent layers take the output from the previous layer
by applying an $1 \times 1 \times k$ filter.
The response of each convolution layer is given by
\begin{equation}
	f_n^{l+1} = \sigma(\sum_{m}(f_m^{l}*k_{m,n}^{l+1})+b_n^{l+1}),
	\label{eq:conv}
\end{equation}
where $f_n^l$ and $f_m^{l+1}$ are the feature maps of layer $l$ and $l+1$, respectively.
In addition, $k$ is the convolution kernel, indices $(m, n)$ denote the mapping
from the $m$-th feature map of one layer to the $n$-th feature map of the next layer.
The function $\sigma(\cdot)$ denotes the Rectified Linear Unit (ReLU)~\cite{nair2010rectified}
and $b$ is the bias.

\begin{figure}[t]\footnotesize
\begin{center}
\begin{tabular}{cccc}
\includegraphics[width=0.24\linewidth]{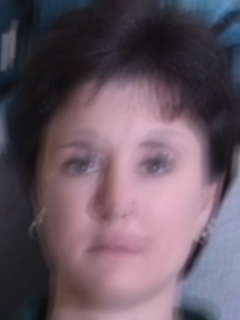} & \hspace{-0.45cm}
\includegraphics[width=0.24\linewidth]{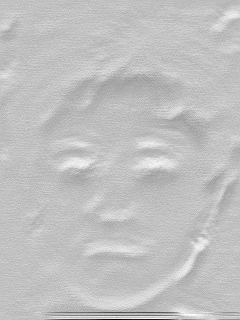} & \hspace{-0.45cm}
\includegraphics[width=0.24\linewidth]{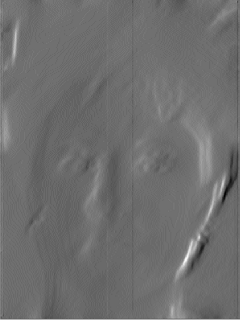} & \hspace{-0.45cm}
\includegraphics[width=0.24\linewidth]{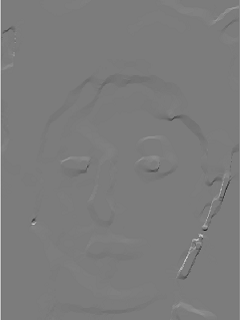} \\
\includegraphics[width=0.24\linewidth]{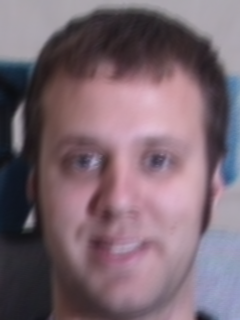} & \hspace{-0.45cm}
\includegraphics[width=0.24\linewidth]{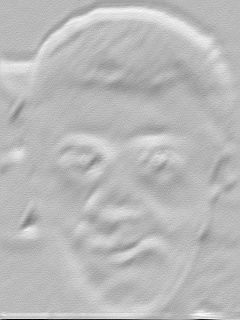} & \hspace{-0.45cm}
\includegraphics[width=0.24\linewidth]{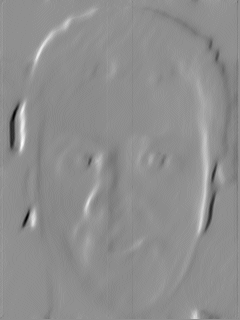} & \hspace{-0.45cm}
\includegraphics[width=0.24\linewidth]{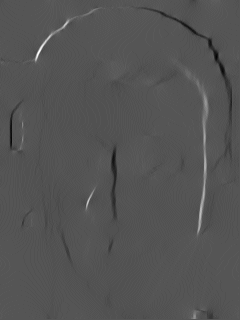} \\
\includegraphics[width=0.24\linewidth]{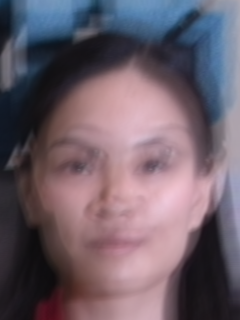} & \hspace{-0.45cm}
\includegraphics[width=0.24\linewidth]{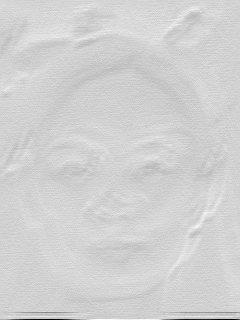} & \hspace{-0.45cm}
\includegraphics[width=0.24\linewidth]{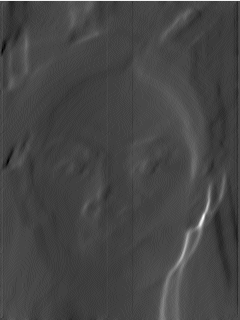} & \hspace{-0.45cm}
\includegraphics[width=0.24\linewidth]{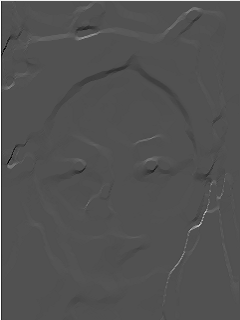} \\
(a) Inputs &\hspace{-0.45cm} (b) Feature maps &\hspace{-0.45cm} (c) Using \cite{Xu/ICML15/deepedge}  &\hspace{-0.45cm} (d) Our results \\
\end{tabular}
\end{center}
\vspace{-0.3cm}
\caption{Extracted salient edges by \cite{Xu/ICML15/deepedge} and our method.
(a) Input blurred images.
(b) Some intermediate feature maps generated by proposed network.
(c) Restored edges by \cite{Xu/ICML15/deepedge}.
(d) Restored edges by the proposed CNN.}
\label{fig: L0vsOur}
\end{figure}

The proposed network is motivated by the state-of-the-art edge prediction approaches~\cite{Cho/et/al,Xu/et/al} which rely on heuristic filtering methods to select sharp edges.
These edge prediction methods usually contain two main steps: 1) suppression of minor details by a smoothing filter and 2) enhancement of strong structures by the shock filter.
In this work, we propose a deep CNN to restore sharp edges from blurred images, where the first few layers are designed to remove details (see Fig.~\ref{fig: L0vsOur}(b)) and the following layers are used to restore sharp edges (see Fig.~\ref{fig: L0vsOur}(d)).

We note that Xu~et al.~\cite{Xu/ICML15/deepedge} propose a CNN to approximate
various image filters.
However, this network architecture cannot
learn sharp structure information when the input image is blurred
as the mapping function between the blurred images and sharp structures
are more complex.
%
Fig. \ref{fig: L0vsOur} shows the structure prediction results
by the CNN \cite{Xu/ICML15/deepedge} and our network.
As the method by Xu~et al.~\cite{Xu/ICML15/deepedge} is designed
to restore edges from clear images, it does perform well on blurry inputs
as shown in Fig. \ref{fig: L0vsOur}(c).
In contrast, the proposed network restores sharp edges (Fig. \ref{fig: L0vsOur}(d)) from the blurred input images, especially at face contour, nose and eyes regions.

\vspace{-2mm}
{\flushleft \textbf{Training.}}
Learning the mapping function between a blurred face image $B$ and the corresponding structure
$S$ is achieved by minimizing the loss between the gradient of the reconstructed structure $S_i$ and the corresponding gradient of the ground-truth structure $S_i^*$,
\vspace{-2mm}
\begin{equation}
\label{eq: loss}
L(\nabla S_i) = \frac{1}{D}\sum_{i=1}^{D}\{\frac{1}{2}||\nabla S_i-\nabla S_i^*||_1+\eta\phi(\nabla S_i)\},
\vspace{-2mm}
\end{equation}
where $D$ is the number of blurred face images in training set, $\phi(z)=\sqrt{(z^2+\epsilon^2)}$ is the sparse regularization to enforce sparsity on gradients and $\eta$ is the parameter for the regularization.
For the ground-truth structure $S_i^*$, we use the $L_0$ smoothing filter~\cite{Xu/L0/smooth} to remove extraneous details in the clear face image $I_i$.
Then the $L_0$ smoothed result $L(I_i)$ can be considered as the desired sharp edges $S_i^*$.

\begin{figure}[!t]\footnotesize
	\begin{center}
		\begin{tabular}{c}
			\hspace{-0.15cm}
			\includegraphics[width=0.98\linewidth]{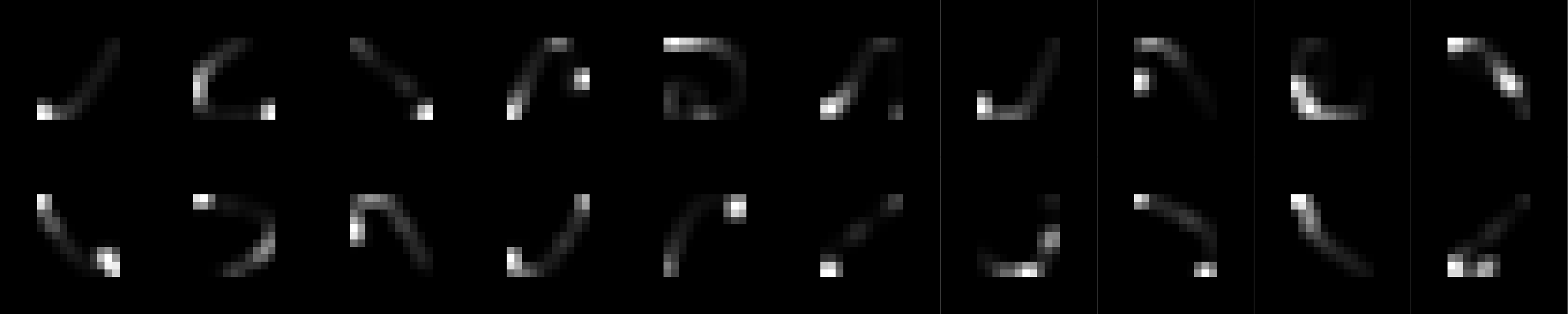} \\
		\end{tabular}
	\end{center}
	\vspace{-0.3cm}
	\caption{Examples of synthetic blur kernels.
	}
	\label{fig:kernel-examples}
\end{figure}

To generate blurred face images, we synthesize blur kernels that appear realistic to real
scenarios by sampling random 3D trajectories used
in~\cite{schmidt/cvpr2013/discriminative}, the obtained trajectories are projected and
rasterized to random square kernel sizes
in the range from $13\times13$ up to $27\times27$ pixels.
Some examples of the generated blur kernels are shown in Fig.~\ref{fig:kernel-examples}.
We synthetically generate blurred face images by convolving each clean image
with 50 generated blur kernels.
%
%
With the blurred face images and corresponding clear images,
the network parameters are learned by minimizing the energy function~\eqref{eq: loss} using the stochastic gradient descent (SGD) scheme.
In the test stage, we apply the trained network to a
blurred face image to generate the salient edges contained in $\nabla S$.

\vspace{-4mm}
\subsection{Kernel Estimation from Exemplar Structure}
\vspace{-1mm}
\label{ssec: Kernel Estimation with Learned Structures}
After obtaining salient edges
by the exemplar-based or CNN-based method, we estimate the blur
kernel by alternately solving
\begin{equation}
\label{eq: deblurring-model-x}
\min_{I} \|I*k-B\|_2^2 + \lambda \|\nabla I\|_0 + \theta \|\nabla I-\nabla S\|_2^2,
\end{equation}
and
\begin{equation}
\label{eq: deblurring-model-k}
\min_{k} \|\nabla S*k - \nabla B\|_2^2 + \gamma \|k\|_2^2,
\end{equation}
where $\lambda$, $\theta$ and $\gamma$ are parameters for the regularization terms.
Here the $L_0$-norm is employed to restore $I$ and remove
ringing artifacts in $I$ as shown by~\cite{Xu/L0/smooth}, and the last term
in \eqref{eq: deblurring-model-x} enforces the gradient of $I$ is similar to the
predicted $\nabla S$.
In~\eqref{eq: deblurring-model-k}, the $L_2$-norm based regularization
is employed to stabilize the blur kernel estimation with a fast solver.

We use the half-quadratic
splitting $L_0$ minimization method~\cite{Xu/L0/smooth}
to solve~\eqref{eq: deblurring-model-x}.
By introducing the auxiliary variable $\textbf{w} = (w_x, w_y)^{\top}$
corresponding to $\nabla I$,
we rewrite~\eqref{eq: deblurring-model-x} as
\begin{equation}
\label{eq: deblurring-model-x-modified}
\min_{I,\textbf{w}} \|I*k-B\|_2^2 + \beta\|\textbf{w} - \nabla I\|_2^2 + \lambda
\|\textbf{w}\|_0 + \theta \|\nabla I-\nabla S\|_2^2,
\end{equation}
where $\beta$ is a scalar weight and increased by a factor of 2 over iterations.
When $\beta$ is close to infinity, the solution
of~\eqref{eq: deblurring-model-x-modified} approaches
that of~\eqref{eq: deblurring-model-x}.

We note that~\eqref{eq: deblurring-model-x-modified} can be
efficiently solved by alternately minimizing $I$ and
$\textbf{w}$.
At each iteration, the solution of $I$ can be obtained by
\begin{equation}
\label{eq: deblurring-model-x-modified-I}
\min_{I} \|I*k-B\|_2^2 + \beta\|\textbf{w} - \nabla I\|_2^2 + \theta \|\nabla I-\nabla S\|_2^2,
\end{equation}
which has a closed-form solution computed in the frequency domain by
\begin{equation}
\small
\label{eq: deblurring-model-x-modified-I-sol-fft}
I =
\mathcal{F}^{-1}\left(\frac{\overline{\mathcal{F}(k)}\mathcal{F}(B)+\beta
    \overline{\mathcal{F}(\nabla)}\mathcal{F}(\textbf{w}) + \theta
    \mathcal{F}_s}
{\overline{\mathcal{F}(k)}\mathcal{F}(k)
  +(\beta+\theta)(\overline{\mathcal{F}(\nabla)}\mathcal{F}(\nabla)}\right).
\end{equation}
Here $\mathcal{F}(\cdot)$ and $\mathcal{F}^{-1}(\cdot)$ denote the
Discrete Fourier Transform (DFT) and inverse DFT, respectively;
$\overline{\mathcal{F}(\cdot)}$ is the complex conjugate operator;
and $\mathcal{F}_s=
\overline{\mathcal{F}(\partial_x)}\mathcal{F}(\partial_x S) + \overline{\mathcal{F}(\partial_y)}\mathcal{F}(\partial_y S)$
where $\partial_x$ and $\partial_y$ denote the
vertical and horizontal derivative operators.

Given $I$, the solution of $\textbf{w}$ in~\eqref{eq: deblurring-model-x-modified} can be obtained by
\begin{equation}
\label{eq: deblurring-model-x-modified-g-sol-w}
\textbf{w} = \left\{\begin{array}{ll}\nabla I, & \mbox{}\ |\nabla I|^2
    \geqslant \frac{\lambda}{\beta},\\0, & \mbox{}\
    $otherwise$. \end{array}\right.
\end{equation}
The main steps for solving~\eqref{eq: deblurring-model-x} are shown in
Algorithm~\ref{alg:latent-image-estimation-algorithm}.

\begin{algorithm}[!t]
\caption{Solving~\eqref{eq: deblurring-model-x}}
\label{alg:latent-image-estimation-algorithm}
\begin{algorithmic}
\STATE {\textbf{Input:} Blurred image $B$ and estimated kernel $k$.}
\STATE $I\gets B$, $\beta\gets 2\lambda$.
\REPEAT
\STATE solve  $\textbf{w}$ using~\eqref{eq:
  deblurring-model-x-modified-g-sol-w}.
\STATE solve $I$ using~\eqref{eq: deblurring-model-x-modified-I-sol-fft}.
\STATE $\beta\gets 2\beta$.
\UNTIL $\beta > 10^5$
\STATE \textbf{Output:} Latent image $I$.
\end{algorithmic}
\end{algorithm}

Based on the above analysis,
the main steps for the proposed kernel estimation algorithm are
summarized in Algorithm~\ref{alg:kernel-estimation-algorithm}.
We use the conjugate gradient method to solve the least squares problem~\eqref{eq: deblurring-model-k}.

In Algorithm~\ref{alg:kernel-estimation-algorithm}, we update the initial predicted $\nabla S$ to remove extraneous weak
edges generated by inaccurate estimation of the CNN-based or exemplar-based method
(see Section~\ref{sec: Property Analysis} for more analysis).
%

\begin{algorithm}[t]
\caption{Blur kernel estimation algorithm}
\label{alg:kernel-estimation-algorithm}
\begin{algorithmic}
\STATE {\textbf{Input:} Blurred image $B$ and predicted salient edges $\nabla S_0$ by the exemplar or CNN-based method.}
\FOR {$l = 1 \to n$}
\STATE solve $k$ using~\eqref{eq: deblurring-model-k}.
\STATE solve  $I$ using
Algorithm~\ref{alg:latent-image-estimation-algorithm}.
\STATE $\nabla S \gets \nabla I$. \/// \verb"Update the salient edges"
\ENDFOR
\STATE \textbf{Output:} Blur kernel $k$.
\end{algorithmic}
\end{algorithm}

\vspace{-3mm}
\subsection{Recovering Latent Images}
\vspace{-1mm}
\label{sec: Final Image Restoration}
Once the blur kernel is determined, the latent image can be estimated by
a number of non-blind deconvolution methods.
In this work, we use the method with a
hyper-Laplacian prior $L_{0.8}$~\cite{Levin/07} to recover the latent image.

\vspace{-3mm}
\section{Experimental Results}
\vspace{-2mm}
\label{sec: Experimental Results}
%

We evaluate the proposed algorithm against the state-of-the-art image deblurring methods on face images.
In addition, we show that the proposed algorithm can be applied to other deblurring tasks by using exemplars of specific classes with categorical structures.
Implemented in MATLAB, it takes about $27$ seconds for the exemplar-based method
to process a blurred image of $320 \times 240$ pixels on an Intel Xeon CPU with 12 GB RAM.
The code and dataset are available on the authors' websites and more results can be found in the supplementary document, available online.
As the method~\cite{HaCohen/iccv13} requires a reference image with
same contents as the blurred image, this is not included in performance evaluation.
However, for completeness we provide some comparisons in the supplementary material.
\vspace{-2mm}
{\flushleft \textbf{Parameter setting.}}
In all the experiments, the parameters $\lambda$, $\theta$, $\gamma$ and $n$ are set
to be $0.002$, $0.001$, $1$ and $50$, respectively.
The sensitivity analysis on these parameters is presented in Section~\ref{sec: Property Analysis}.
\vspace{-2mm}
{\flushleft \textbf{Dataset.}}
For the exemplar-based method, we use a set of $2,435$ face images from the CMU PIE
dataset~\cite{Multi/PIE/datasets} (which contains face images in different poses and expressions) as our dataset.
%
%
To train the proposed network, we use $2,435$ exemplar images and $50$ blur kernels as the training dataset.
That is, a set of $121,750$ blurred images is used in the training process.
The identities of exemplar and test sets are not overlapped in all the experiments.
%

\vspace{-4mm}
\subsection{Synthetic Dataset using Frontal Faces}
\begin{figure}[t]\footnotesize
	\begin{center}
		\begin{tabular}{cc}
			\includegraphics[height=0.4\linewidth]{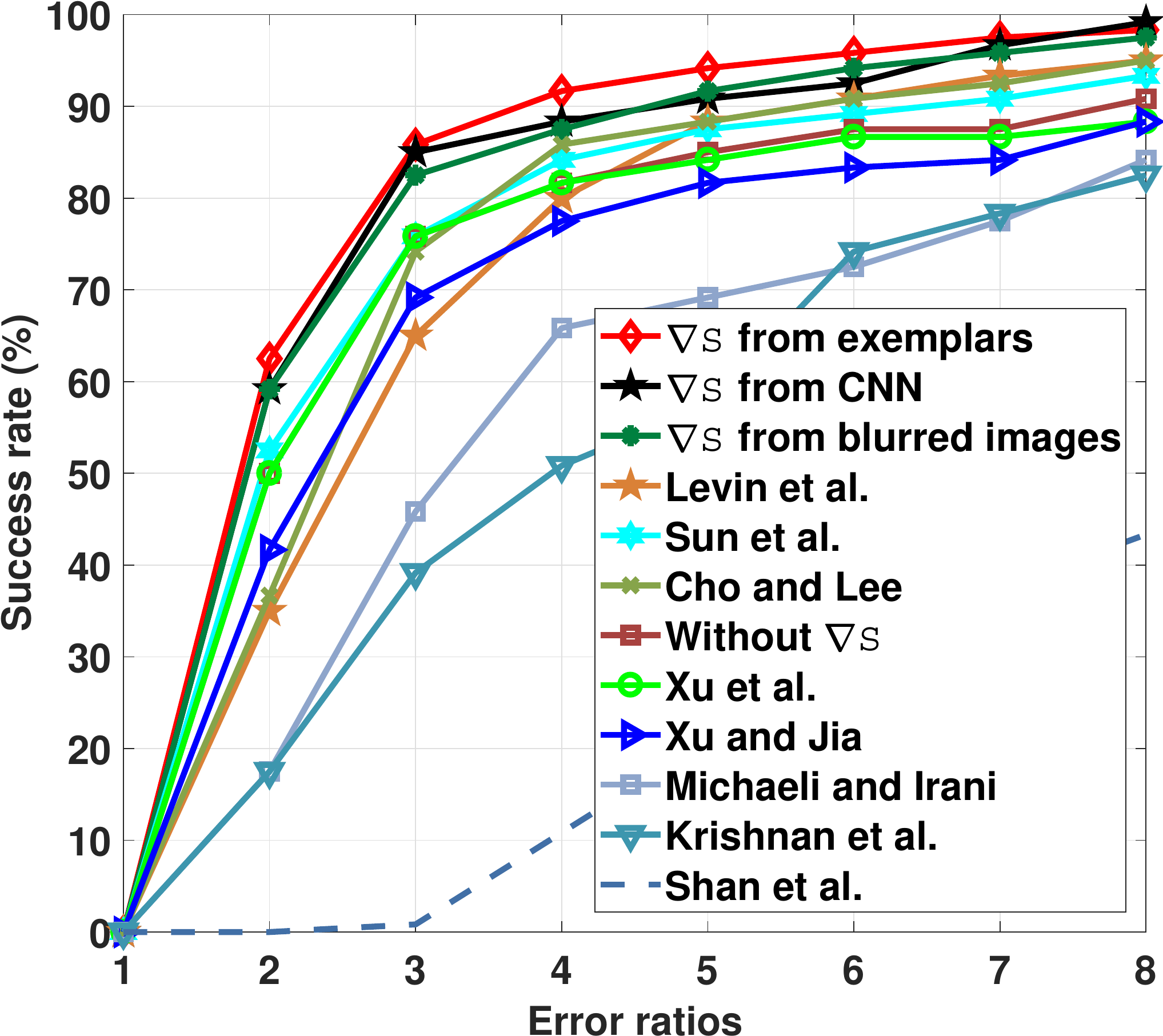} & \hspace{-0.52cm}
			\includegraphics[height=0.4\linewidth]{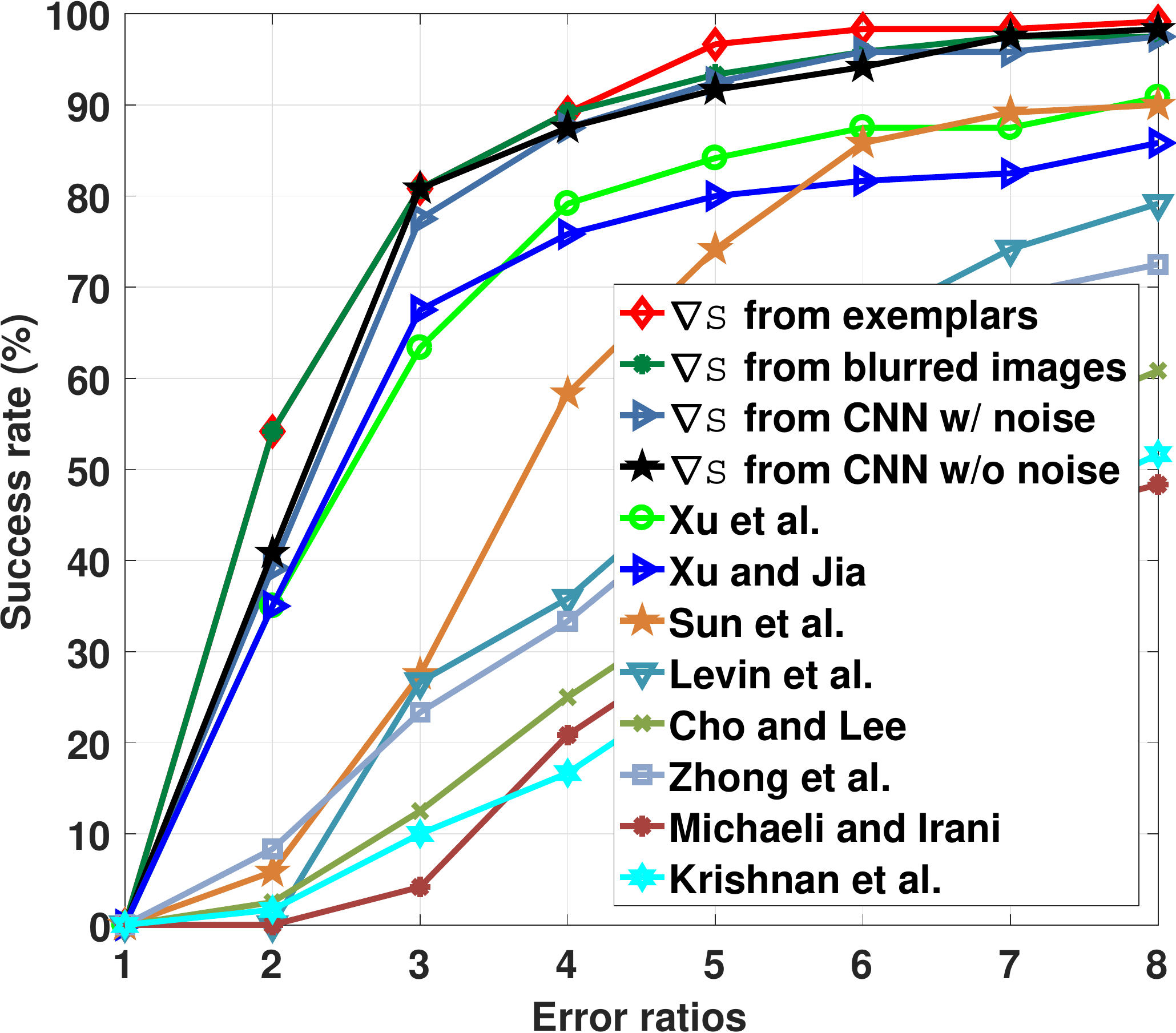} \\
			\scriptsize (a) Results on noise-free images & \scriptsize (b) Results on noisy images \\
		\end{tabular}
	\end{center}
	\vspace{-0.3cm}
	\caption{Quantitative comparisons with several state-of-the-art
		single-image blind deblurring methods:
		Shan~et al.~\cite{Shan/et/al}, Cho and Lee~\cite{Cho/et/al}, Xu and
		Jia~\cite{Xu/et/al}, Krishnan~et al.~\cite{Krishnan/CVPR2011},
		Levin~et al.~\cite{Levin/CVPR2011}, Zhong~et al.~\cite{zhong/lin_cvpr2013/noise/deblur}, Xu~et al.~\cite{Xu/l0deblur/cvpr2013}, Sun et al.~\cite{libin/sun/patchdeblur_iccp2013}, and Michaeli and Irani~\cite{tomer/eccv/MichaeliI14}.
	}
	\label{fig: error-ratios}
\end{figure}

\vspace{-1mm}
\label{sec: synthetic frontal}
For quantitative evaluations, we collect a
dataset of 60 clear face images and 8 ground-truth
kernels in a way similar to~\cite{Levin/CVPR2009} to generate a
test set of 480 blurred inputs.
We evaluate the proposed algorithms against state-of-the-art methods
based on edge selection~\cite{Cho/et/al,Xu/et/al} and
sparsity priors~\cite{Shan/et/al,Krishnan/CVPR2011,Levin/CVPR2011,Xu/l0deblur/cvpr2013}.
We use the non-blind deconvolution method~\cite{epll/iccv} and adopt the error metric proposed by
Levin~et al.~\cite{Levin/CVPR2009} for fair comparison.
Fig.~\ref{fig: error-ratios} shows the cumulative error ratio where
higher curves indicate more accurate results.
The proposed algorithms generate better results than
state-of-the-art methods for face image deblurring.
The results show the advantages of using facial structures as the guidance
over those using local edge selection methods~\cite{Cho/et/al,Xu/et/al,Xu/l0deblur/cvpr2013}.

We evaluate different schemes to predict edges $\nabla S$:
1) using the edges of exemplars $\nabla T_{i^*}(x)$ as $\nabla S$ (original);
2) using the edges predicted by the CNN;
3) using the edges of the input image $\nabla B(x)$ as $\nabla S$ (i.e., using $\nabla B(x)$ instead of $\nabla T_{i^*}(x)$ to compute $\nabla S$ in~\eqref{eq: structure-component});
4) not using $\nabla S$ at all.
Fig.~\ref{fig: error-ratios}(a) shows
the first three approaches perform similarly as the matched $\nabla T_{i^*}(x)$,
predicted edges $\nabla S$  by the CNN, and $\nabla B(x)$
share similar structures, which also demonstrates the effectiveness of
the proposed exemplar-based and CNN-based methods.
On the other hand,
the schemes using the predicted edges perform significantly better than the one
without using predicted edges.

\begin{figure*}[!t]\footnotesize
\begin{center}
\begin{tabular}{ccccc}
\includegraphics[width=0.19\linewidth]{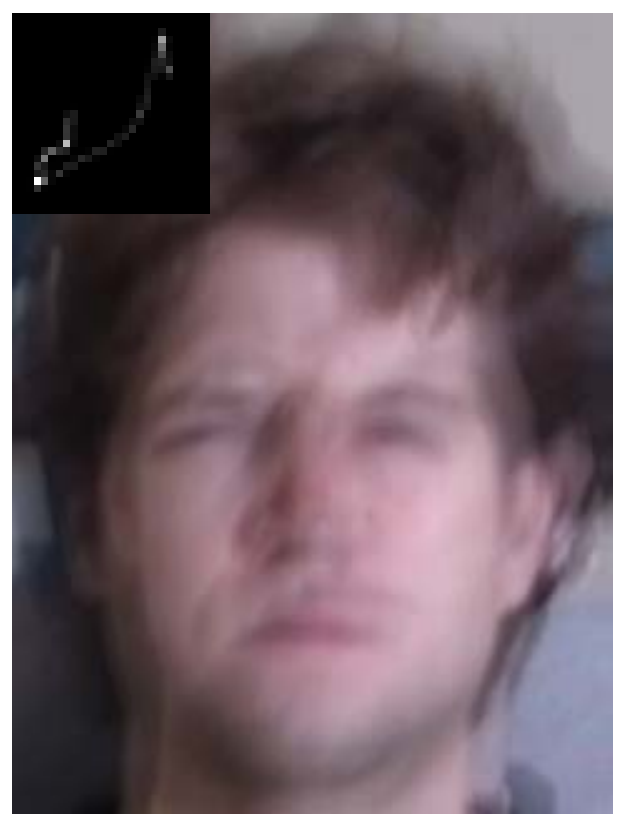} & \hspace{-0.48cm}
\includegraphics[width=0.19\linewidth]{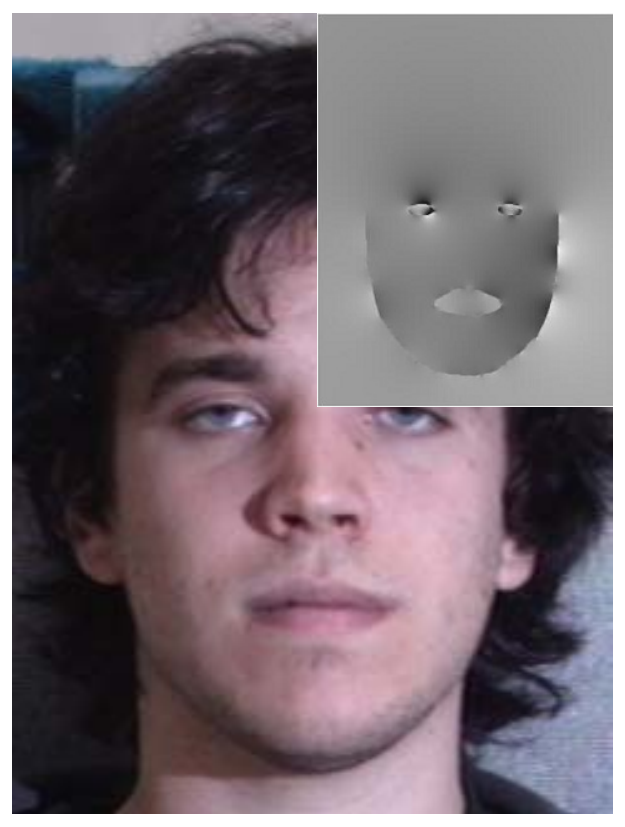} & \hspace{-0.48cm}
\includegraphics[width=0.19\linewidth]{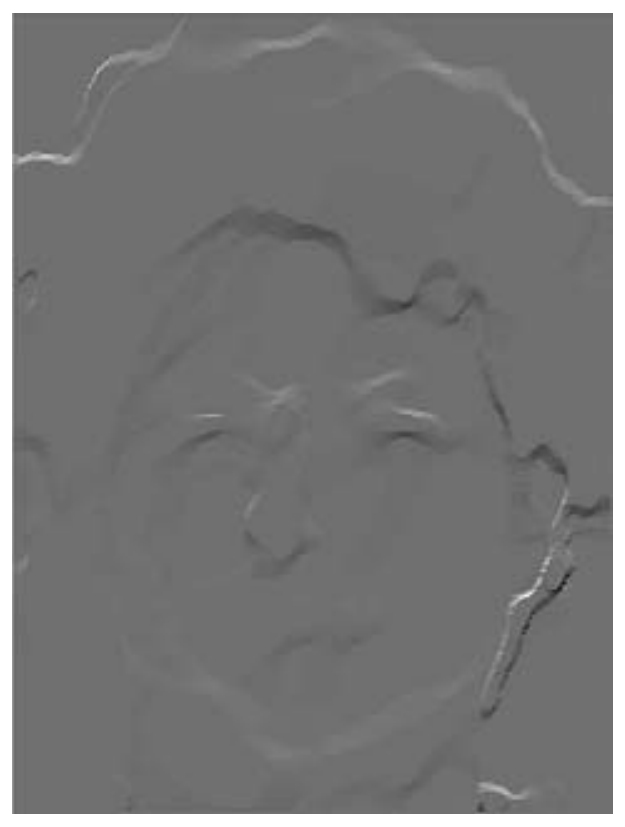} & \hspace{-0.48cm}
\includegraphics[width=0.19\linewidth]{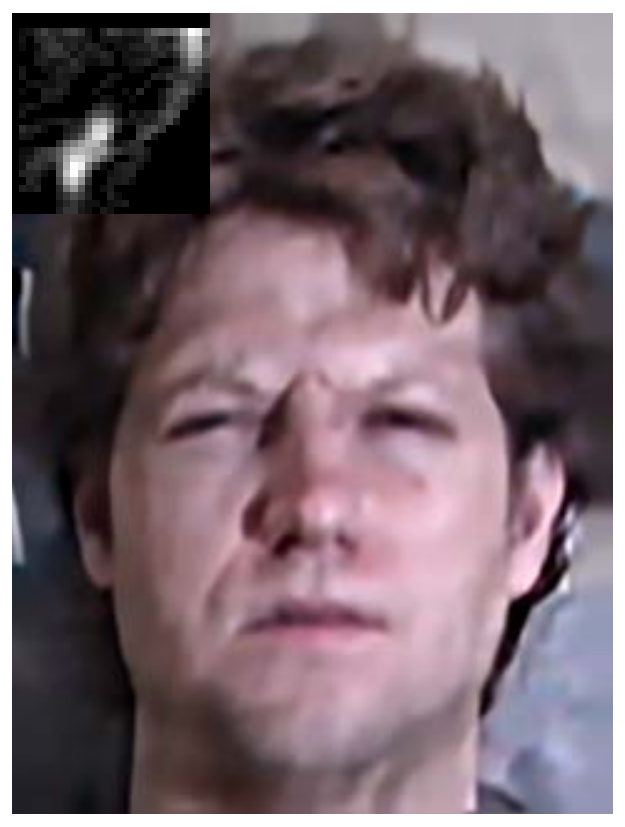} & \hspace{-0.48cm}
\includegraphics[width=0.19\linewidth]{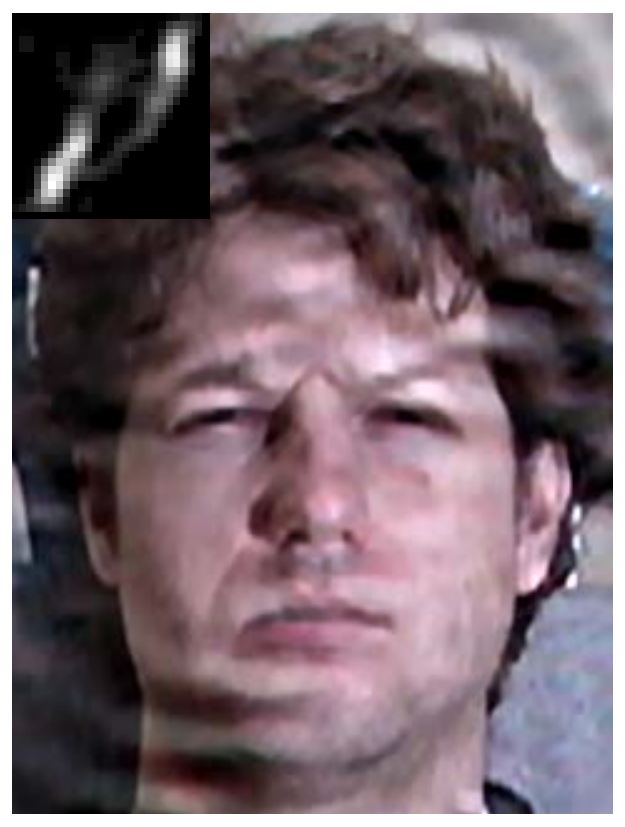} \\
(a) Input and kernel & \hspace{-0.48cm} (b) $\nabla S$ from exemplar & \hspace{-0.48cm} (c) CNN-based $\nabla S$ & \hspace{-0.48cm} (d) Sun~et al.~\cite{libin/sun/patchdeblur_iccp2013} & \hspace{-0.48cm} (e) Xu and Jia~\cite{Xu/et/al} \\
\hfill
\includegraphics[width=0.19\linewidth]{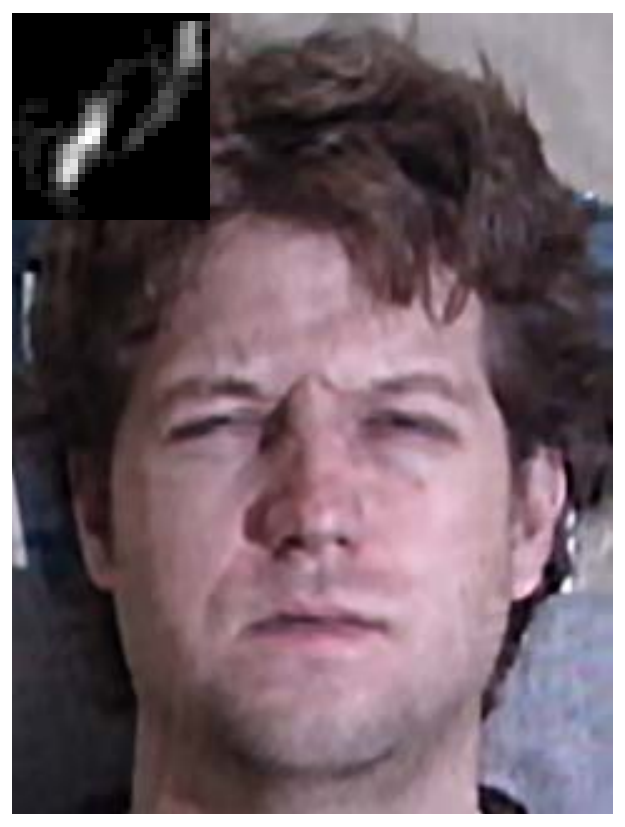} & \hspace{-0.48cm}
\includegraphics[width=0.19\linewidth]{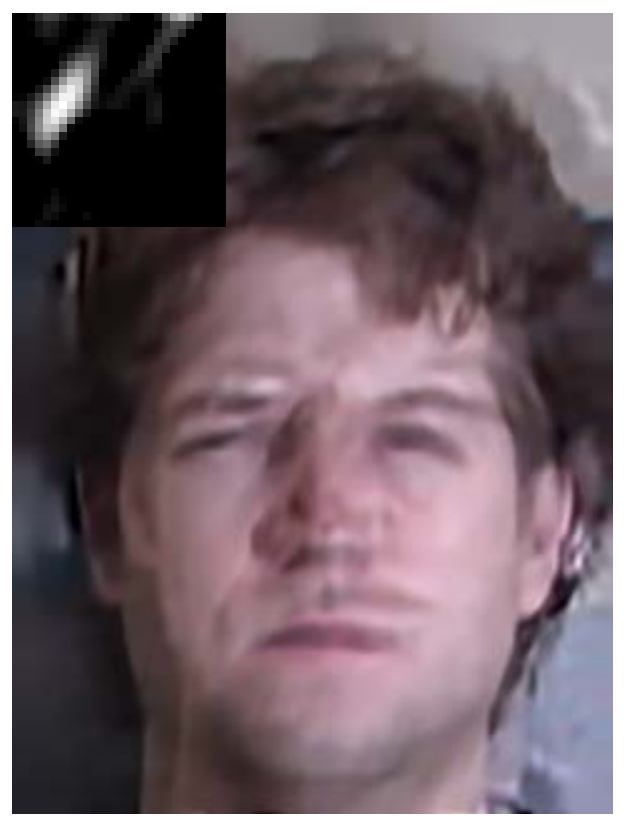} & \hspace{-0.48cm}
\includegraphics[width=0.19\linewidth]{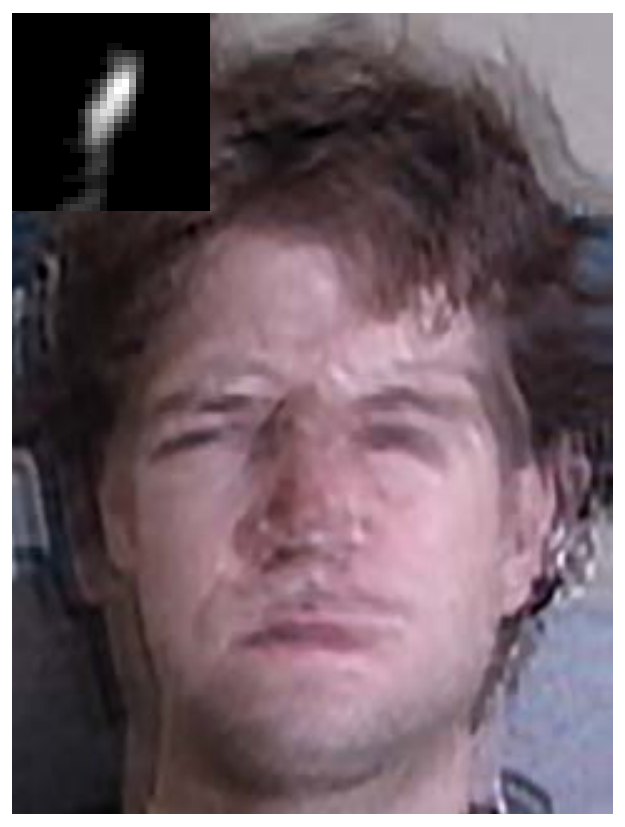} & \hspace{-0.48cm}
\includegraphics[width=0.19\linewidth]{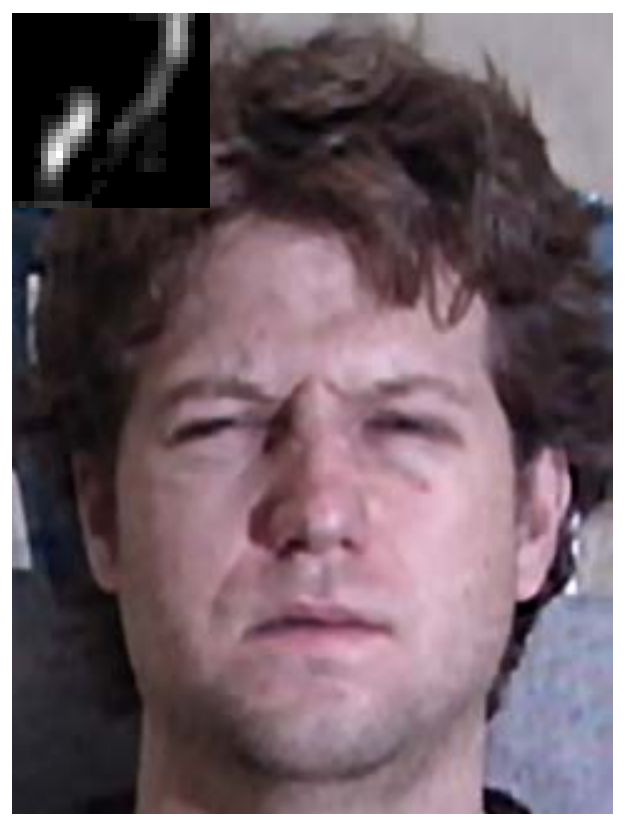} & \hspace{-0.48cm}
\includegraphics[width=0.19\linewidth]{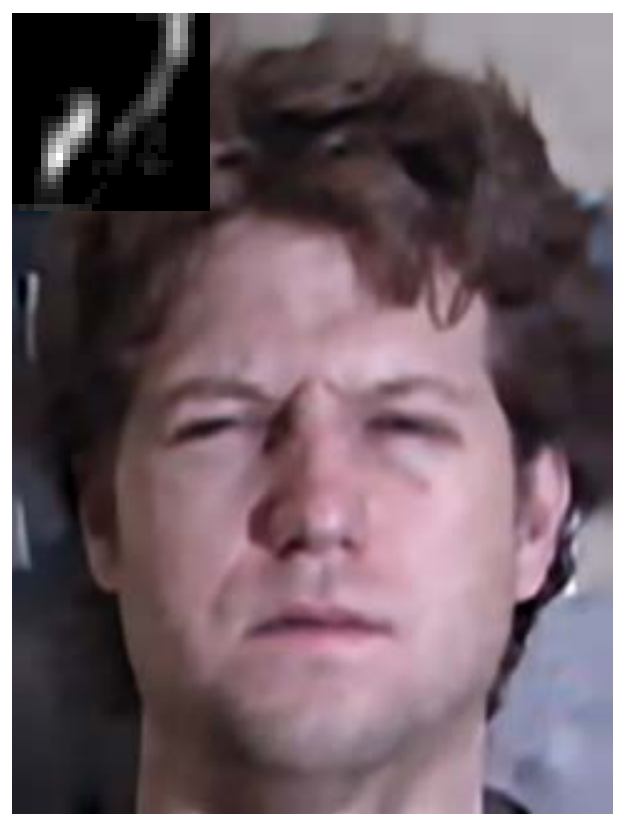} \\
(f) Xu~et al.~\cite{Xu/l0deblur/cvpr2013} & \hspace{-0.48cm} (g) Michaeli and Irani~\cite{tomer/eccv/MichaeliI14} & \hspace{-0.48cm} (h) Ours without $\nabla S$ & \hspace{-0.48cm} (i) Our exemplar-based & \hspace{-0.48cm} (j) Our CNN-based\\
\end{tabular}
\end{center}
\vspace{-0.2cm}
	\caption{An example from the synthesized frontal face test dataset.
	}
	\label{fig: visualization-examples1}
\end{figure*}

\begin{figure*}[!t]\footnotesize
	\begin{center}
		\begin{tabular}{ccccc}
			\includegraphics[width=0.23\linewidth]{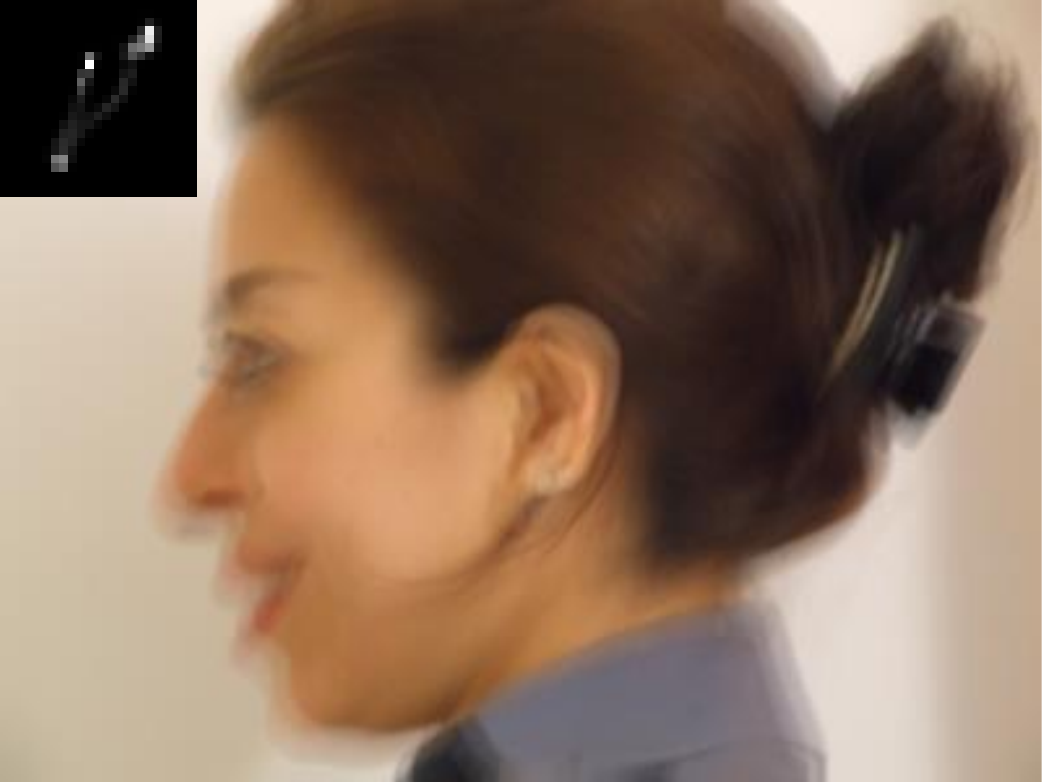} & \hspace{-0.45cm}
			\includegraphics[width=0.23\linewidth]{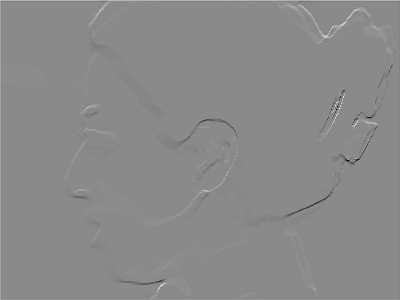} &\hspace{-0.45cm}
			\includegraphics[width=0.23\linewidth]{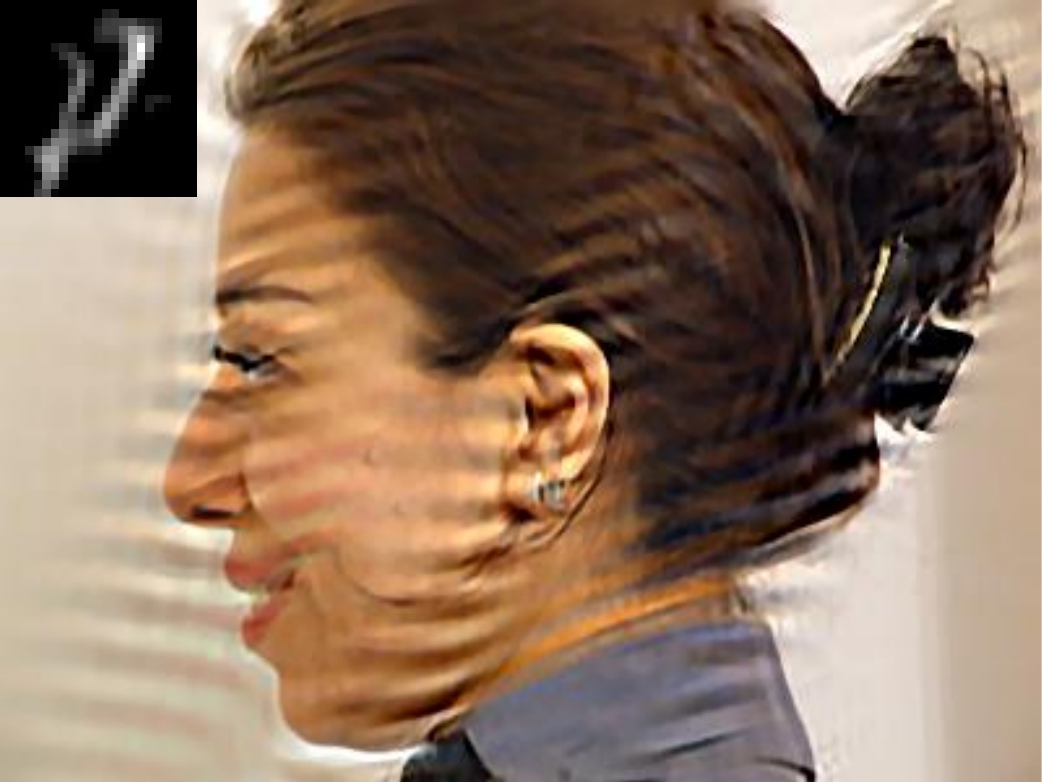} & \hspace{-0.45cm}
			\includegraphics[width=0.23\linewidth]{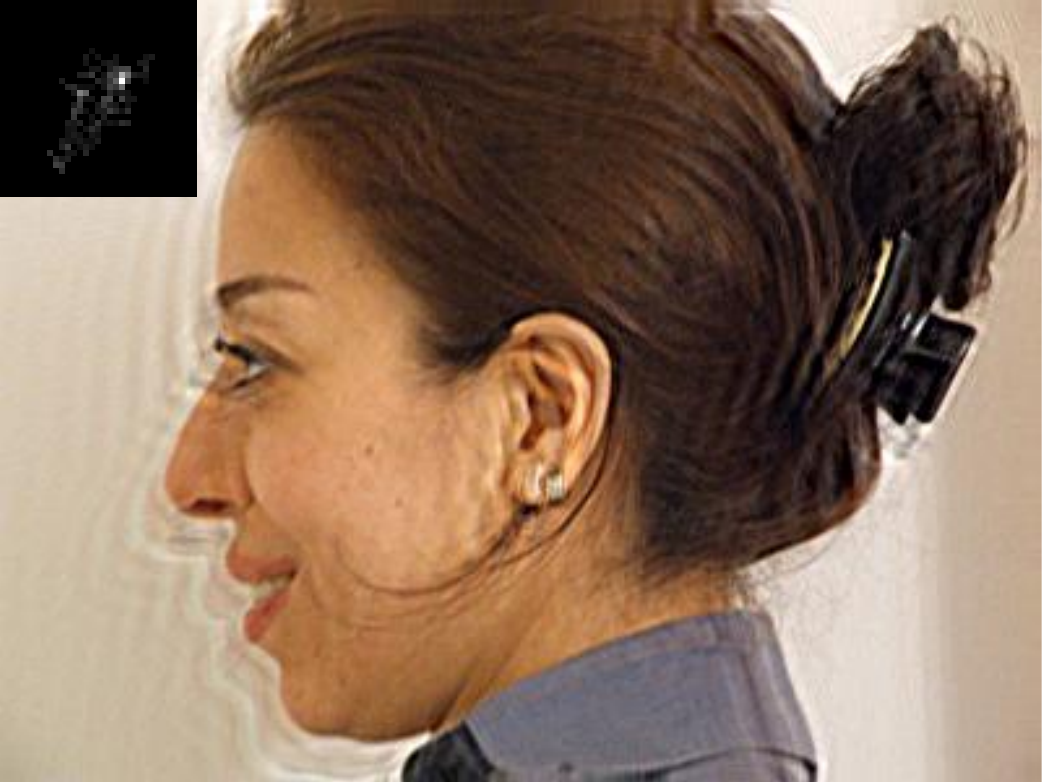} \\
			(a) Input and kernel &\hspace{-0.45cm} (b) Predicted $\nabla S$ &\hspace{-0.45cm} (c) Shan~et al.~\cite{Shan/et/al}  &\hspace{-0.45cm} (d)  Cho and Lee~\cite{Cho/et/al} \\
			\hfill
			\includegraphics[width=0.23\linewidth]{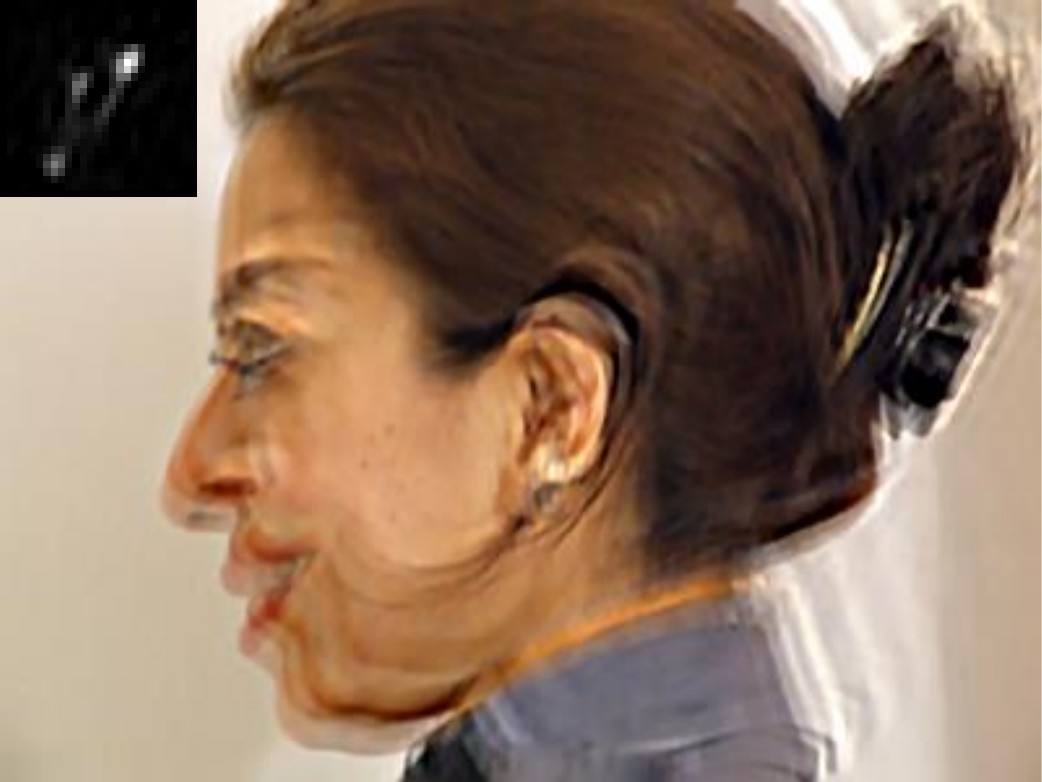} & \hspace{-0.45cm}	
			\includegraphics[width=0.23\linewidth]{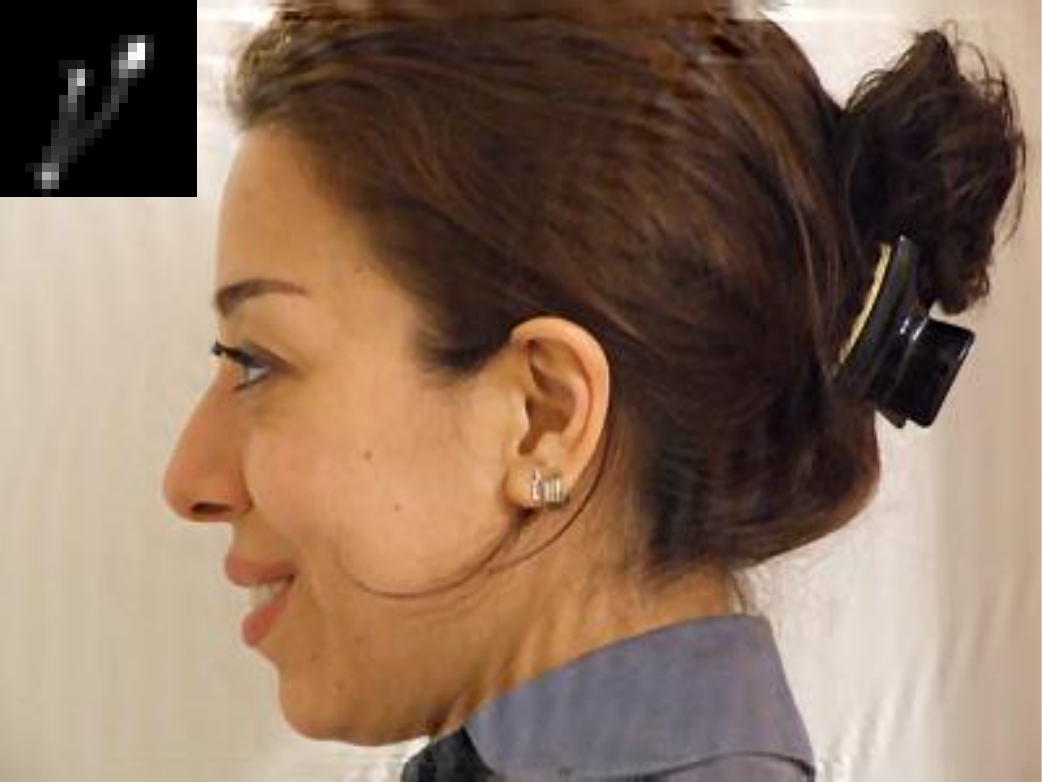} & \hspace{-0.45cm}
			\includegraphics[width=0.23\linewidth]{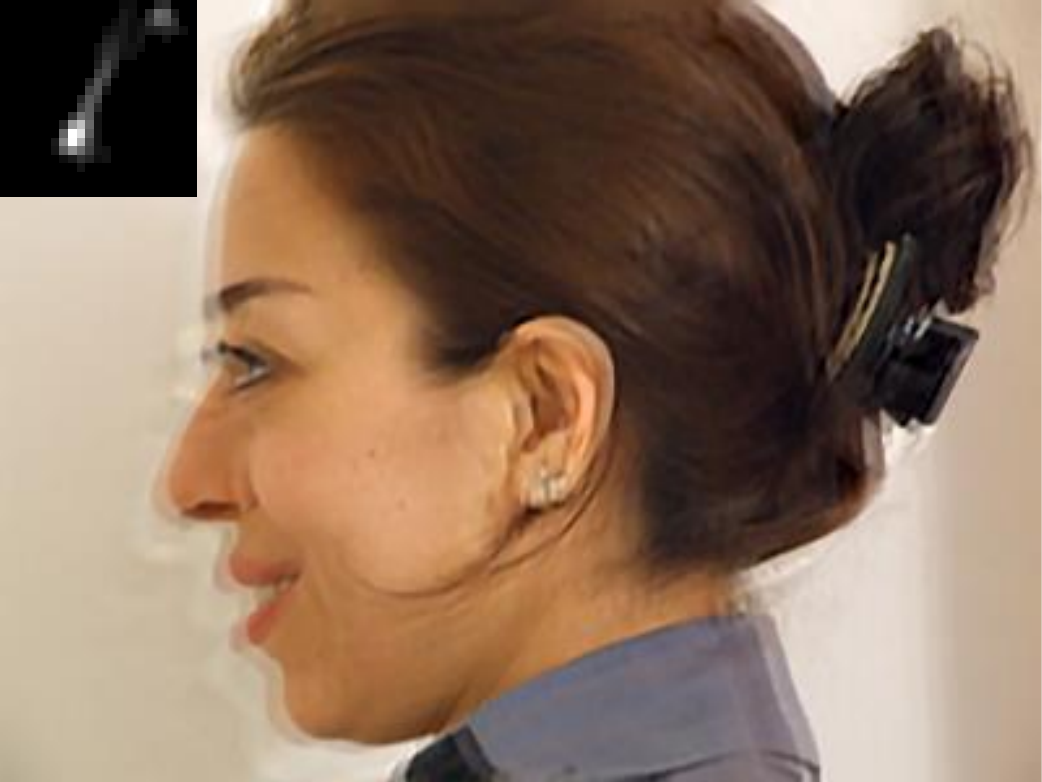} & \hspace{-0.45cm}
			\includegraphics[width=0.23\linewidth]{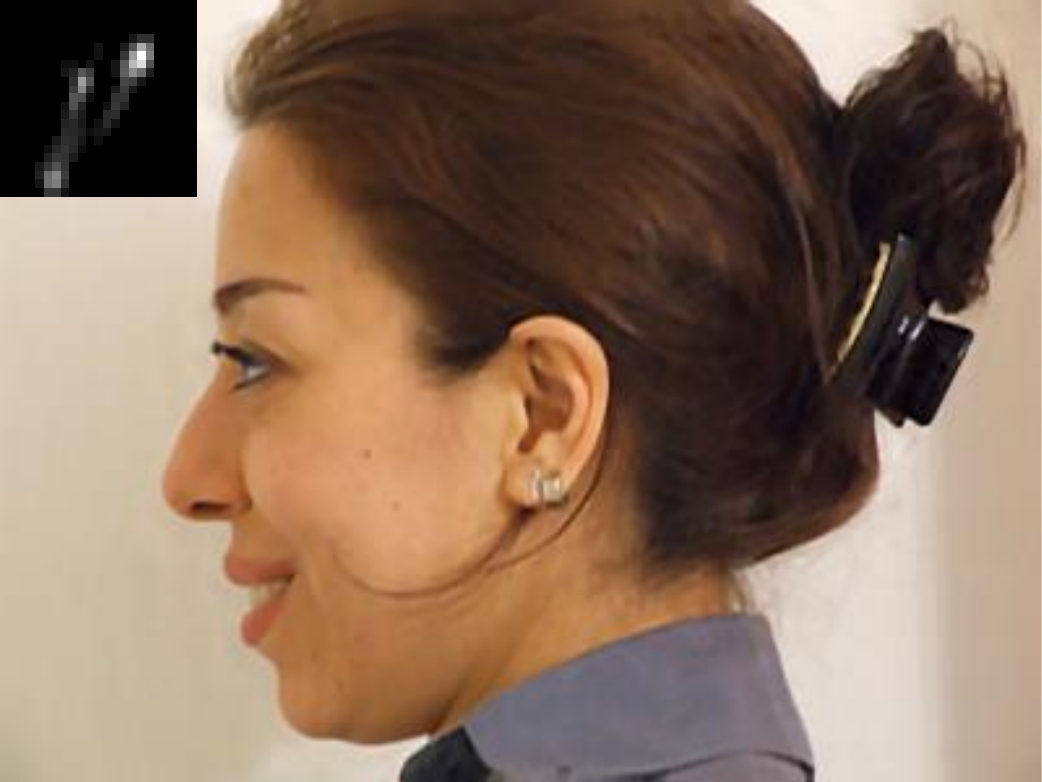} \\
			(f) Krishnan~et al.~\cite{Krishnan/CVPR2011} &\hspace{-0.45cm} (h) Xu~et al.~\cite{Xu/l0deblur/cvpr2013} &\hspace{-0.45cm} (i) Ours without $\nabla S$ &\hspace{-0.45cm} (j) Our CNN-based\\
		\end{tabular}
	\end{center}
	\vspace{-0.3cm}
	\caption{An example from the synthesized profile face test dataset.
	}
	\label{fig: pose1}
\end{figure*}

We note the proposed method without predicted $\nabla S$ does not use coarse-to-fine strategies and generates similar results to~\cite{Xu/l0deblur/cvpr2013},
which indicates that the coarse-to-fine strategy does not help kernel estimation
on blurry face images with less texture.
In addition, we note that the results generated by the
exemplar-based method are slightly better than those by the CNN-based one as shown in Fig.~\ref{fig: error-ratios}(a).
One of the main reasons is that the exemplar-based method directly uses the structures of the clear exemplars, while the CNN-based algorithm
uses the structures predicted from blurred inputs via regression.
Thus, edges from exemplars are much sharper than
those of the CNN-based method, which accordingly lead to better kernel estimates.
However, the run time of the prediction step by the CNN-based algorithm is
significantly less than that of the exemplar-based method as shown in Table~\ref{tab: run time}.
%
%
The average run time of the prediction step by the exemplar-based method is 4260 seconds.
In contrast, the average run time of the prediction step by the CNN-based method is
only 0.95 seconds.
\begin{table}[t]
	\caption{Average run time (/s) on the 480 test images.}
	\centering
\vspace{-3mm}
\begin{tabular}{cccc}
		\toprule
		Time & Prediction step & Deblurring step & Total \\
		\midrule
		Exemplar-based & $4260$ & $27$ & $4287$\\
		Deep CNN-based & $0.95$ & $27$ & $27.95$ \\
		\bottomrule
	\end{tabular}
	\label{tab: run time}
\end{table}

Fig.~\ref{fig: error-ratios}(b)  shows the quantitative comparisons when
1\% random noise is added to the test images for examples.
%
For the CNN-based algorithm, we train the proposed network using noise-free images (denoted as
$\nabla{S}$ from CNN w/o noise in Fig.~\ref{fig: error-ratios}(b))
and images with random noise (denoted as $\nabla{S}$ from CNN w/ noise
in Fig.~\ref{fig: error-ratios}(b)) to evaluate the deblurring performance under noise.
%
Compared to other state-of-the-art methods, the proposed algorithms perform well on blurry images with noise.
We note that the results on noisy images show higher curves than those with
noise-free images.
The reason is that a noisy input increases the denominator value
of the measure~\cite{Levin/CVPR2009}.
Thus the error ratios from noisy images are usually smaller than those
from noise-free inputs, under the same blur kernel.

We show one example from the test set in Fig.~\ref{fig: visualization-examples1}.
The method based on the patch recurrence prior~\cite{tomer/eccv/MichaeliI14}
generates deblurred images with significant blur residual
as the statistical models are designed for generic objects without exploiting categorical structures.
%
The edge based methods~\cite{Xu/et/al,libin/sun/patchdeblur_iccp2013} do not perform well for
face deblurring as the assumption that there exist a sufficient
number of sharp edges in the latent images does not hold.
Compared to the method based on an $L_0$-regularization~\cite{Xu/l0deblur/cvpr2013},
the results by the proposed algorithms contain significantly fewer artifacts.

In Fig.~\ref{fig: visualization-examples1}(b), although the best matched exemplars are from different
identities with different facial expressions, the main structures of (a) and (b) are similar, e.g., the lower face contours and upper eye contours.
In addition, the learned sharp edges capture the main structures of the blurred inputs
as shown in Fig.~\ref{fig: visualization-examples1}(c).
The deblurred results also indicate that our search approach~\eqref{eq:
  normalized-cross-correlation} is able to find
the image with similar structure, and the learning scheme~\eqref{eq: loss}
is able to restore the sharp latent edge from an input.
The results shown in Fig.~\ref{fig: visualization-examples1}(i) and (j) demonstrate that
the predicted salient edges significantly improve the accuracy of
kernel estimation,
while the results without predicted salient edges are similar to delta functions.
Although our method is also developed within the MAP framework,
the predicted salient edges based on the matched exemplar or CNN
provide good initialization for kernel estimation such that the
issue with delta kernel solution (e.g., Fig.~\ref{fig: visualization-examples1}(h))
is addressed effectively.

\vspace{-4mm}
\subsection{Synthetic Dataset using Profile Faces}
\vspace{-1mm}
\label{sec: synthetic profile}
We collect a dataset of 50 clear profile face images from the PICS dataset
(\href{http://pics.psych.stir.ac.uk/}{http://pics.psych.stir.ac.uk/}) and
8 ground-truth kernels from~\cite{Levin/CVPR2009} to generate a test set of 400 blurred face images.
%
%
As the proposed algorithms perform similarly as discussed in Section~\ref{sec: synthetic frontal},
we only compare the CNN-based method with the state-of-the-arts~\cite{Levin/CVPR2011,Cho/et/al,Xu/et/al,Krishnan/CVPR2011,Xu/l0deblur/cvpr2013,libin/sun/patchdeblur_iccp2013}.
One example from this profile face dataset and the deblurred
results are shown in Fig.~\ref{fig: pose1}.

Fig.~\ref{fig: pose1}(b) show the predicted structures by the proposed
CNN method for the blurry profile face images.
Note that most blurred edges are not included in the predicted salient structures.
Similar to the results presented in Section~\ref{sec: synthetic frontal},
the estimated kernels and restored images by Cho and Lee~\cite{Cho/et/al} contain
a significant amount of noise as shown in Fig.~\ref{fig: pose1}(d).
The deblurred results by the method based on the sparsity priors~\cite{Shan/et/al,Krishnan/CVPR2011}
contain ringing artifacts as shown in Fig.~\ref{fig: pose1}(c) and (f).

\begin{figure}[!t]
	\begin{center}
		\begin{tabular}{c}
			\includegraphics[width=0.76\linewidth]{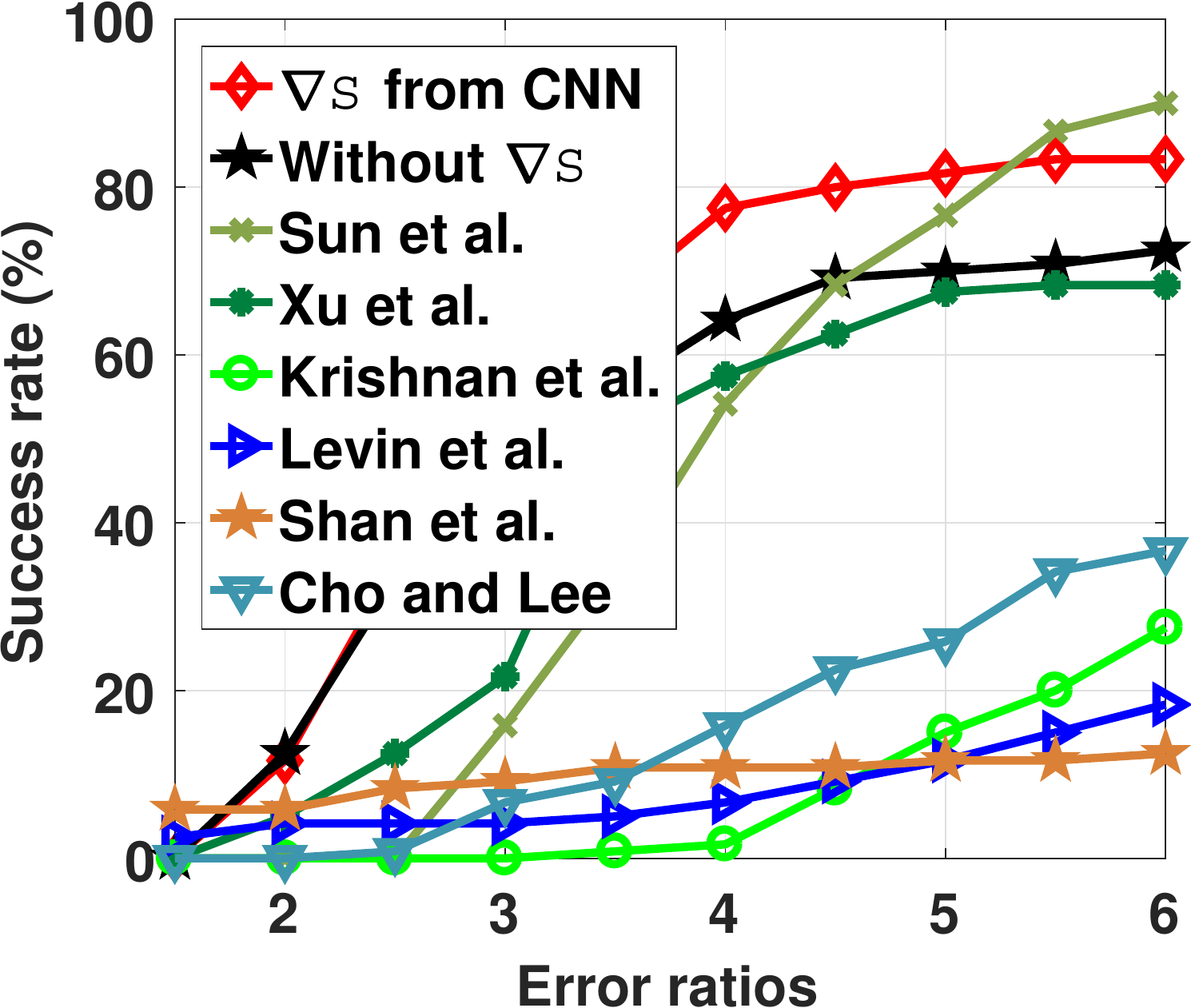} \\
		\end{tabular}
	\end{center}
	\vspace{-0.3cm}
	\caption{Quantitative comparisons on profile faces with several state-of-the-art single-image blind deblurring methods:
		Shan~et al.~\cite{Shan/et/al},
		Cho and Lee~\cite{Cho/et/al},
		Krishnan~et al.~\cite{Krishnan/CVPR2011},
		Levin~et al.~\cite{Levin/CVPR2011},
		Xu~et al.~\cite{Xu/l0deblur/cvpr2013},
and Sun et al.~\cite{libin/sun/patchdeblur_iccp2013}.
	}
	\label{fig: errorratio-pose}
\end{figure}

Quantitatively, Fig.~\ref{fig: errorratio-pose} shows that the proposed algorithm based on the
CNN performs well against the state-of-the-art methods on this dataset of profile face images
based on the cumulative error ratio~\cite{Levin/CVPR2009}.

\vspace{-3mm}
\subsection{Real Images}
\begin{figure*}[!t]\footnotesize
	\begin{center}
		\begin{tabular}{ccccc}
			\includegraphics[width=0.19\linewidth]{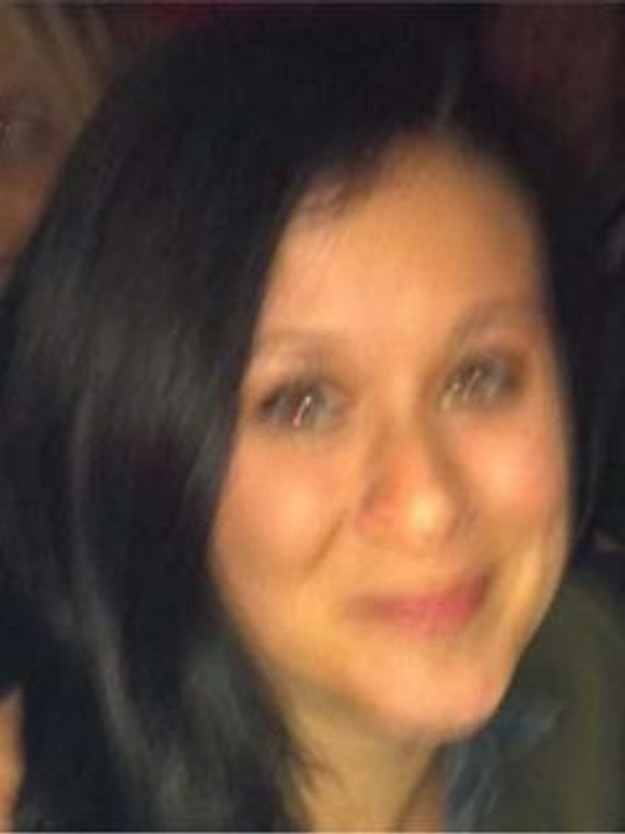} & \hspace{-0.4cm}
			\includegraphics[width=0.19\linewidth]{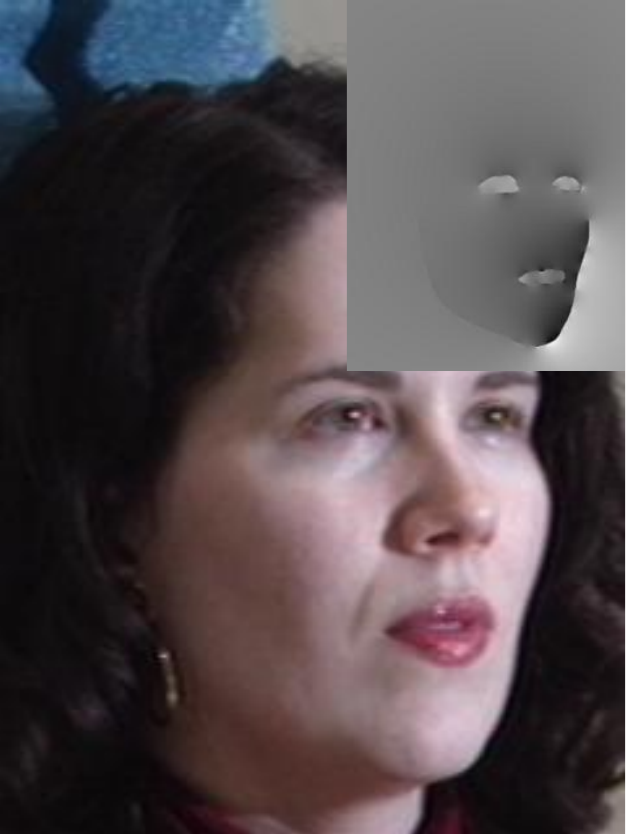} & \hspace{-0.4cm}
			\includegraphics[width=0.19\linewidth, height = 0.254\linewidth]{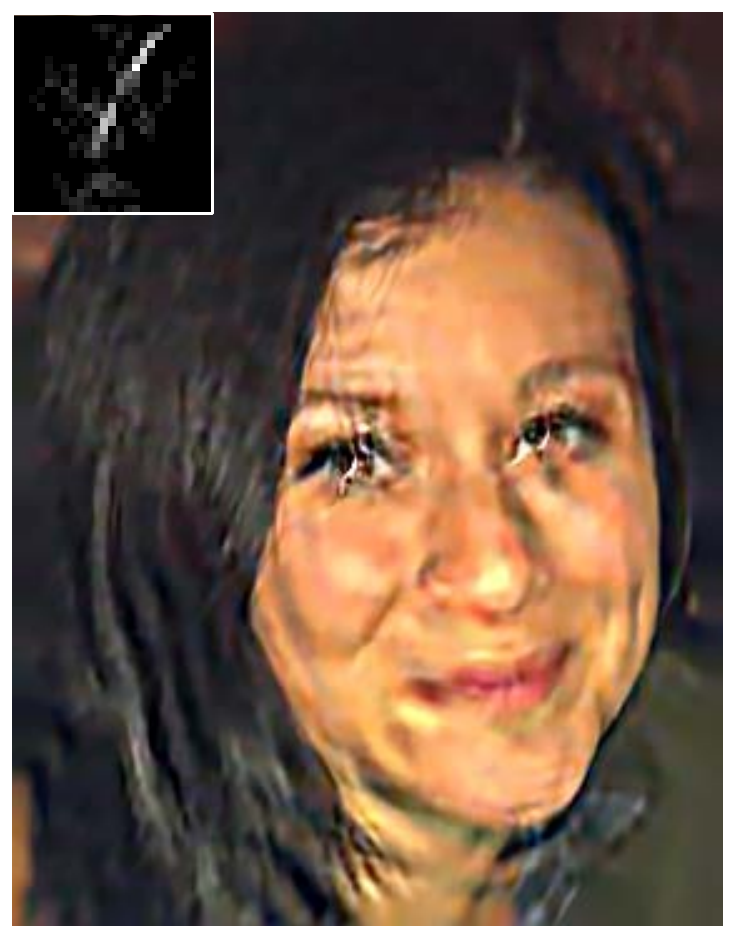} & \hspace{-0.4cm}
			\includegraphics[width=0.19\linewidth]{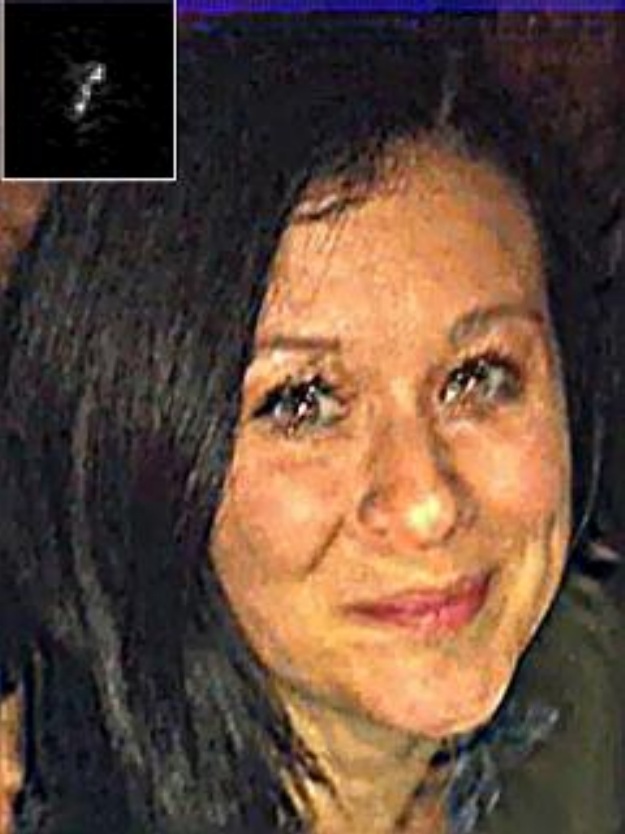} & \hspace{-0.4cm}
			\includegraphics[width=0.19\linewidth]{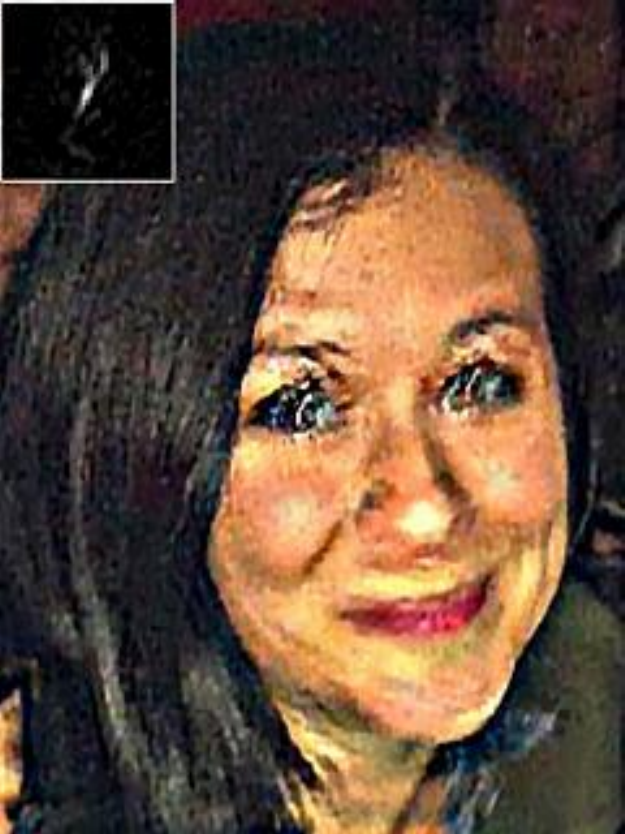} \\
			(a) Input &\hspace{-0.4cm} (b) Exemplar-based $\nabla S$ &\hspace{-0.4cm} (c) Sun~et al.~\cite{libin/sun/patchdeblur_iccp2013} &\hspace{-0.4cm} (d) Cho and Lee~\cite{Cho/et/al} &\hspace{-0.4cm} (e) Xu and Jia~\cite{Xu/et/al} \\
			\hfill
			\includegraphics[width=0.19\linewidth]{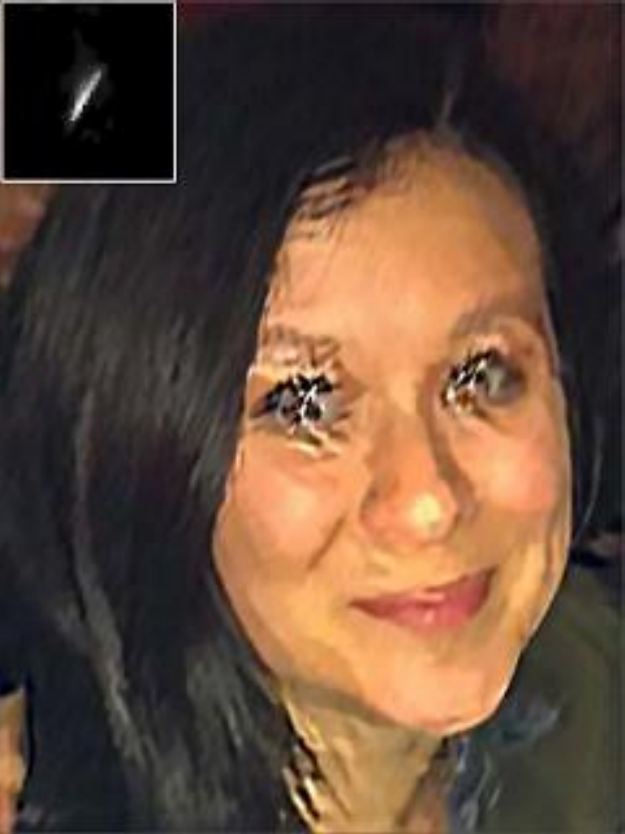} & \hspace{-0.4cm}
			\includegraphics[width=0.19\linewidth]{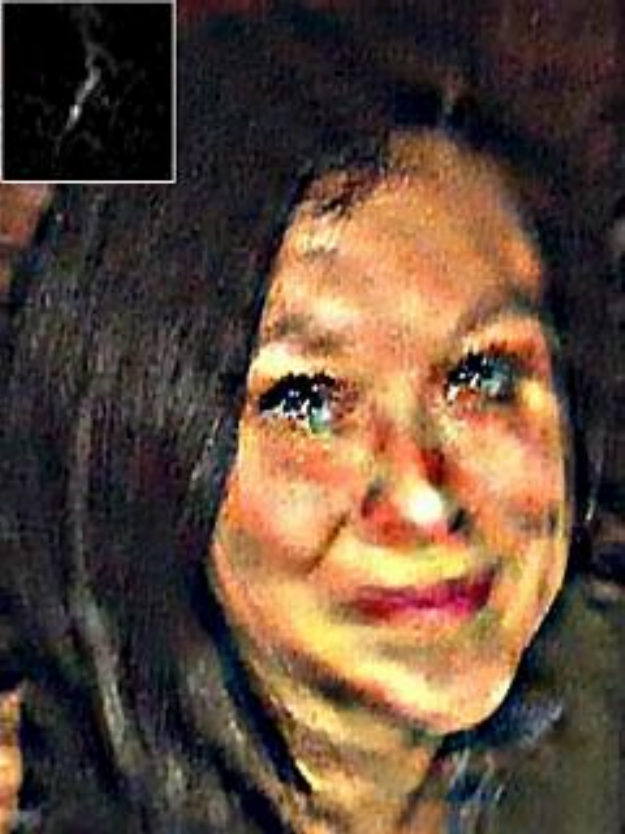} & \hspace{-0.4cm}
            \includegraphics[width=0.19\linewidth, height = 0.254\linewidth]{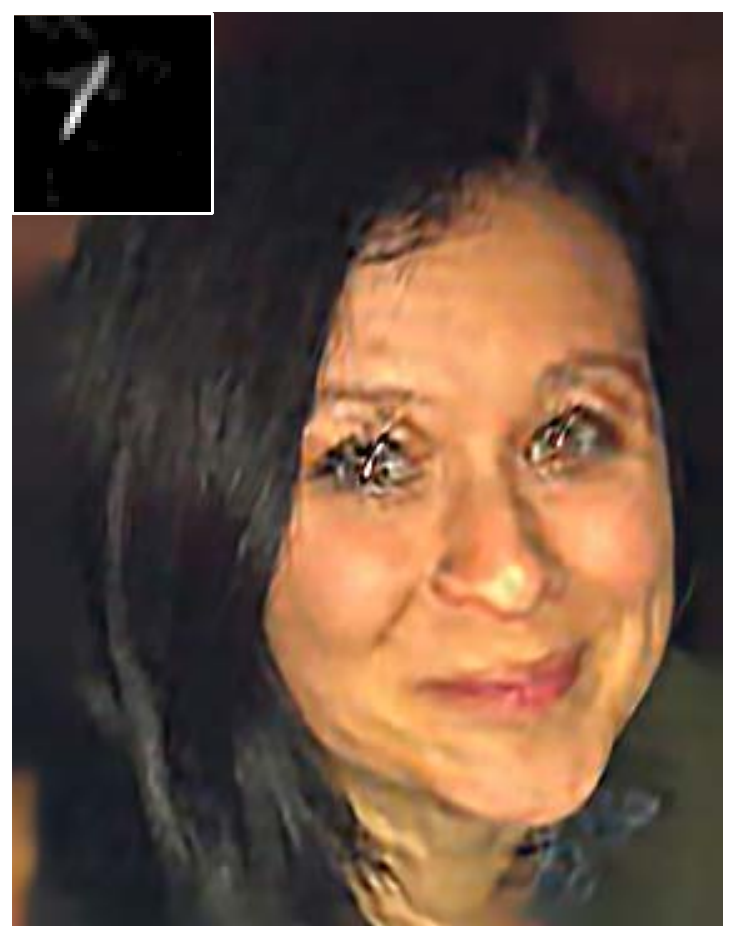} & \hspace{-0.4cm}
			\includegraphics[width=0.19\linewidth]{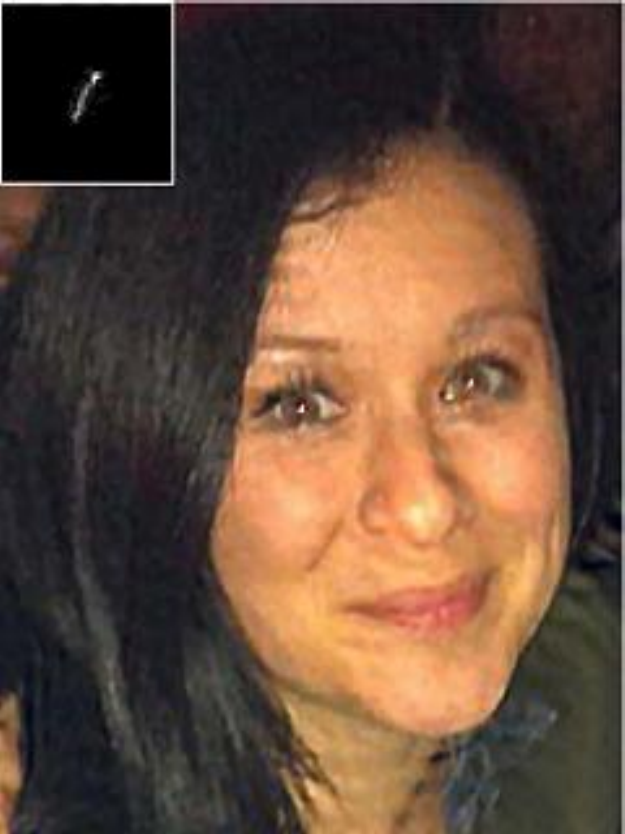} & \hspace{-0.4cm}
			\includegraphics[width=0.19\linewidth]{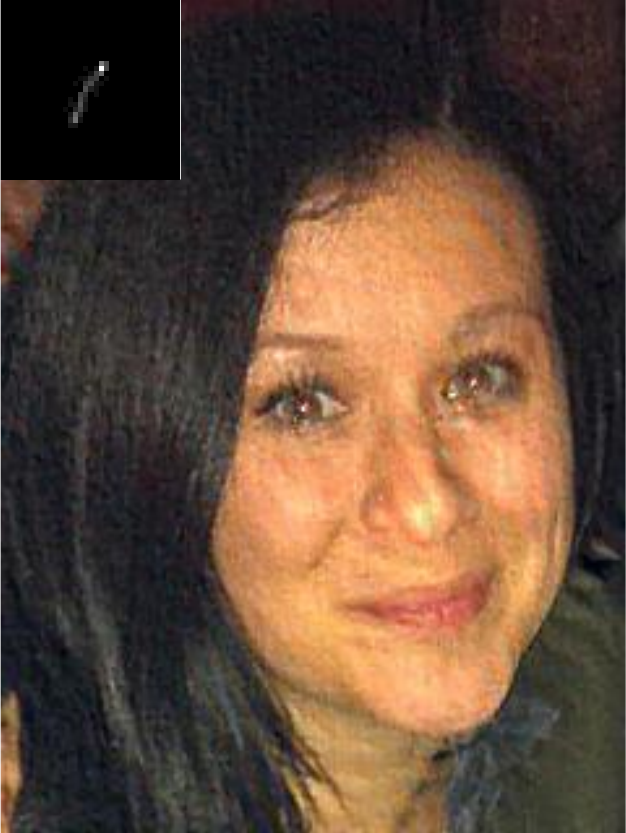} \\
			(f) Zhong~et al.~\cite{zhong/lin_cvpr2013/noise/deblur} &\hspace{-0.4cm} (g) Xu~et al.~\cite{Xu/l0deblur/cvpr2013} &\hspace{-0.4cm} (h) Michaeli and Irani~\cite{tomer/eccv/MichaeliI14}  &\hspace{-0.4cm} (i) Our exemplar-based &\hspace{-0.4cm} (j) Our CNN-based \\
		\end{tabular}
	\end{center}
	\vspace{-0.3cm}
\caption{Real captured example with some noise and saturated pixels.
The estimated kernel is of $35\times35$ pixels.
	}
	\label{fig: real-examples}
\end{figure*}

\vspace{-1mm}
\label{sec: real images}
We evaluate the proposed algorithms with comparisons to the state-of-the-art deblurring
methods  using real blurred images.
%
The input image in Fig.~\ref{fig: real-examples}(a) contains some noise and saturated pixels.
The deblurred results by the state-of-the-art methods~\cite{Cho/et/al,Xu/et/al,libin/sun/patchdeblur_iccp2013,tomer/eccv/MichaeliI14,zhong/lin_cvpr2013/noise/deblur} contain
noticeable noise and ringing artifacts.
In contrast, the proposed exemplar-based method is able to deblur this image with fewer visual artifacts and finer details (Fig.~\ref{fig: real-examples}(i))
despite the best matched exemplar (Fig.~\ref{fig: real-examples}(b)) is significantly different from the input.
Furthermore, the deblurred result by the proposed CNN-based method also contains fewer ringing artifacts as shown in Fig.~\ref{fig: real-examples}(j).
%

\begin{figure*}[!t]\footnotesize
	\begin{center}
		\begin{tabular}{ccccc}
			\includegraphics[width=0.19\linewidth, height=0.23\linewidth]{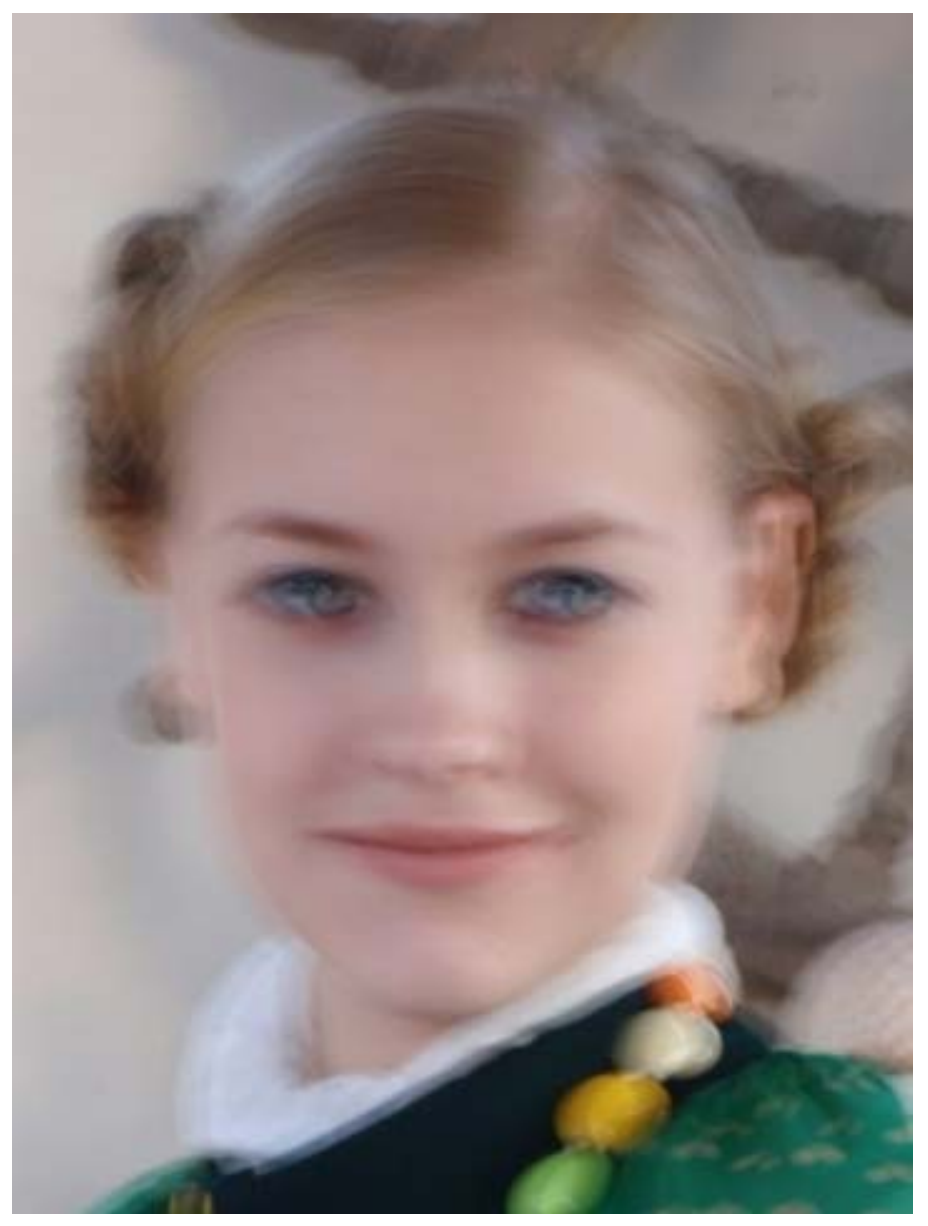} & \hspace{-0.4cm}
			\includegraphics[width=0.19\linewidth, height=0.23\linewidth]{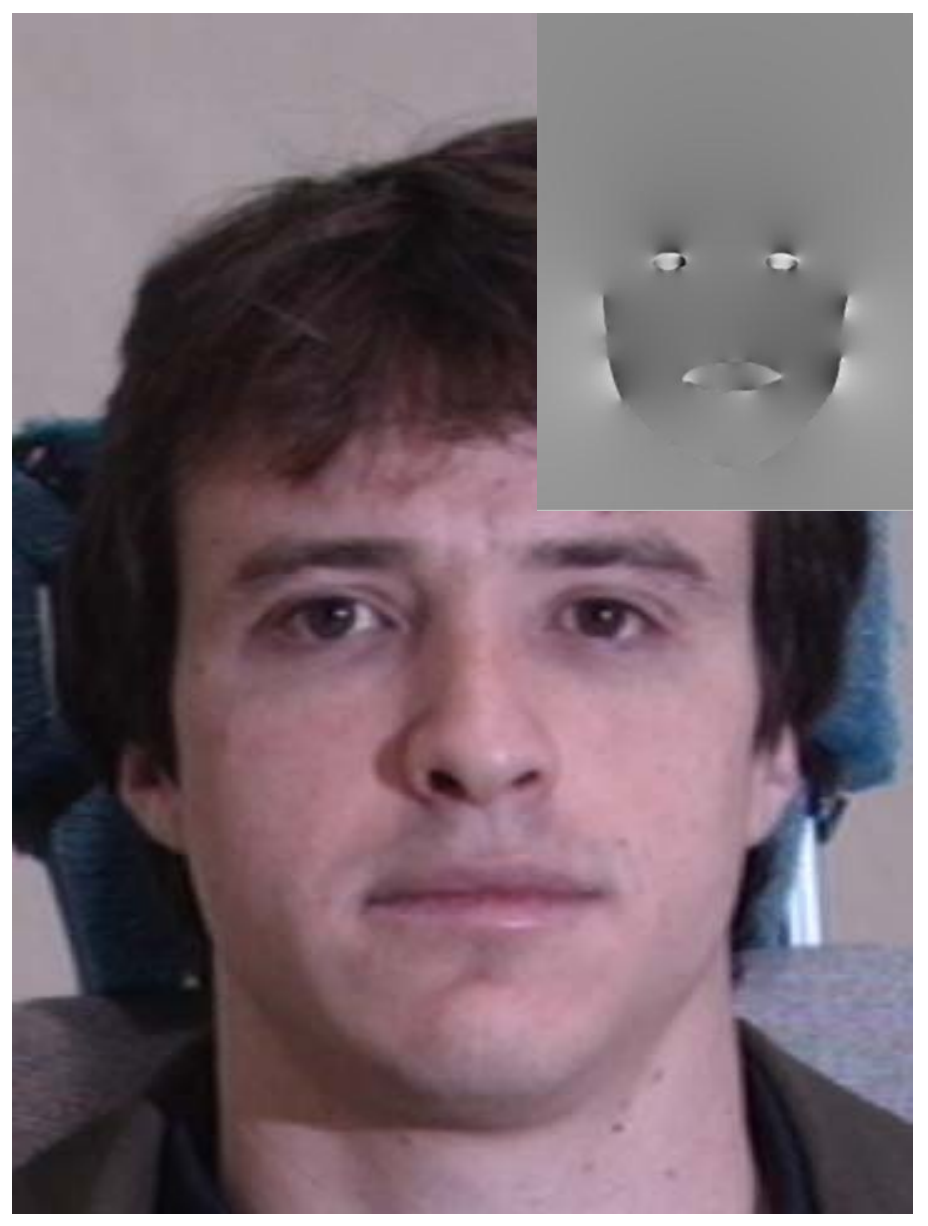} & \hspace{-0.4cm}
			\includegraphics[width=0.19\linewidth, height=0.23\linewidth]{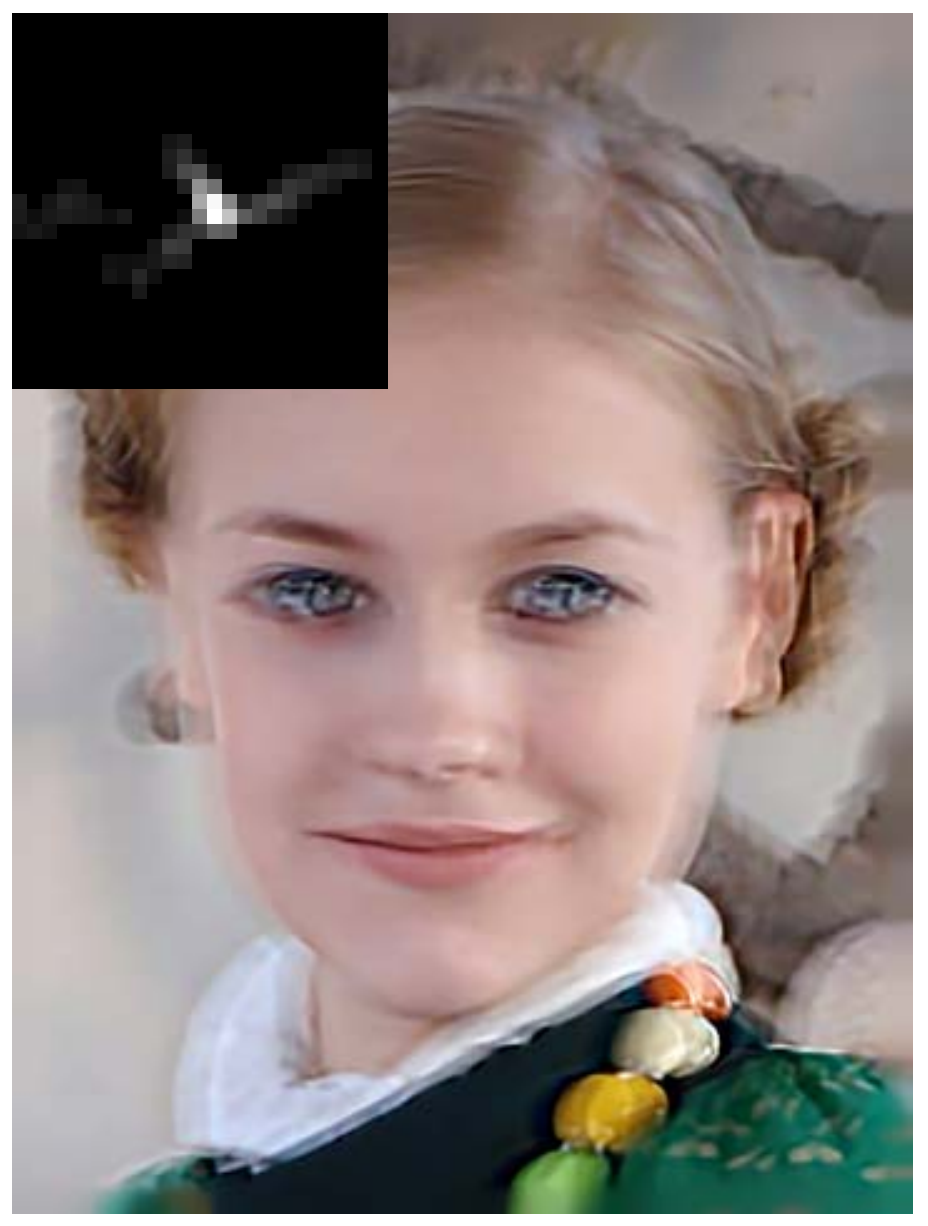} & \hspace{-0.4cm}
			\includegraphics[width=0.19\linewidth, height=0.23\linewidth]{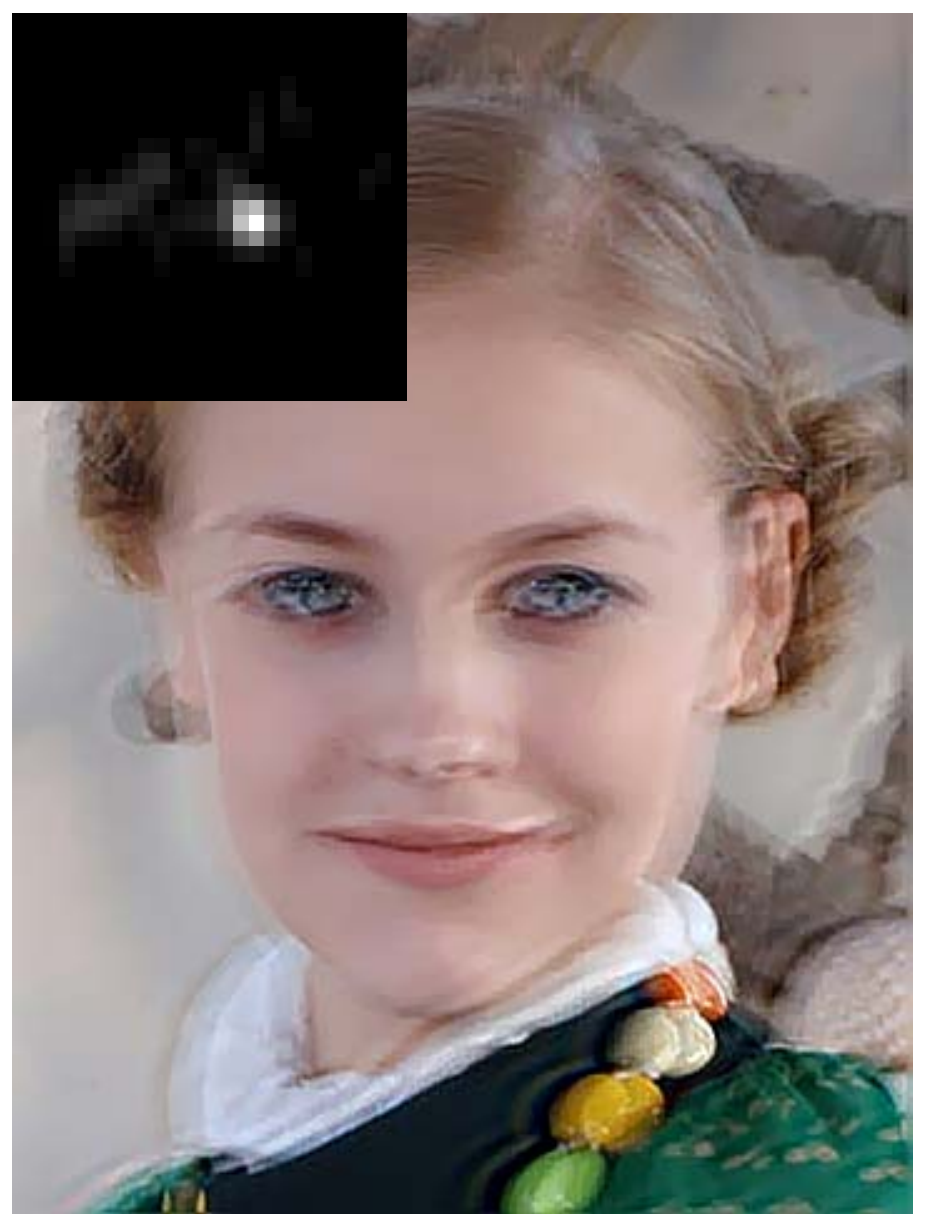} & \hspace{-0.4cm}
			\includegraphics[width=0.19\linewidth, height=0.23\linewidth]{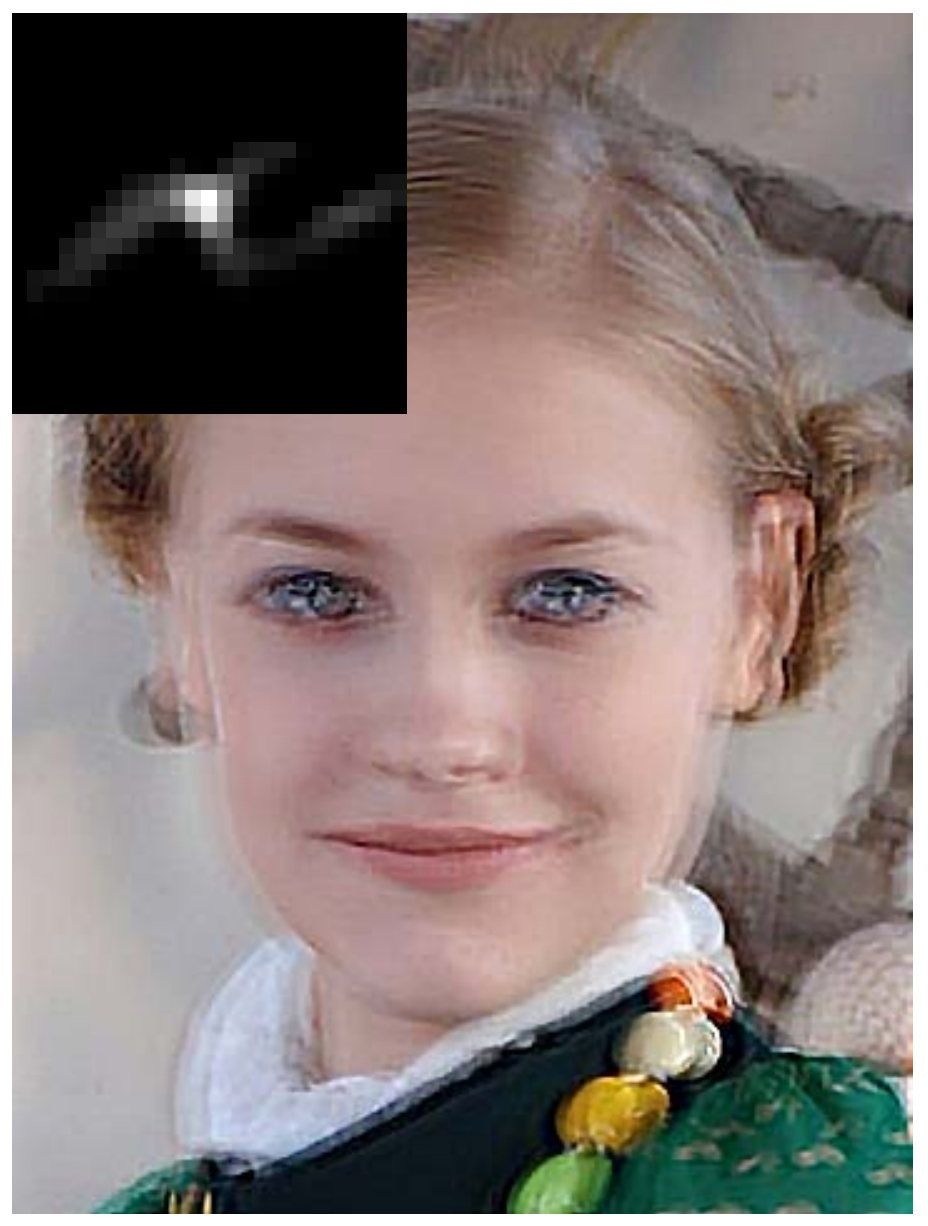} \\
			(a) Input &\hspace{-0.4cm} (b) Exemplar-based $\nabla S$ &\hspace{-0.4cm} (c) Sun~et al.~\cite{libin/sun/patchdeblur_iccp2013} &\hspace{-0.4cm} (d) Cho and Lee~\cite{Cho/et/al} &\hspace{-0.4cm} (e) Xu and Jia~\cite{Xu/et/al} \\
			\hfill
			\includegraphics[width=0.19\linewidth, height=0.22\linewidth]{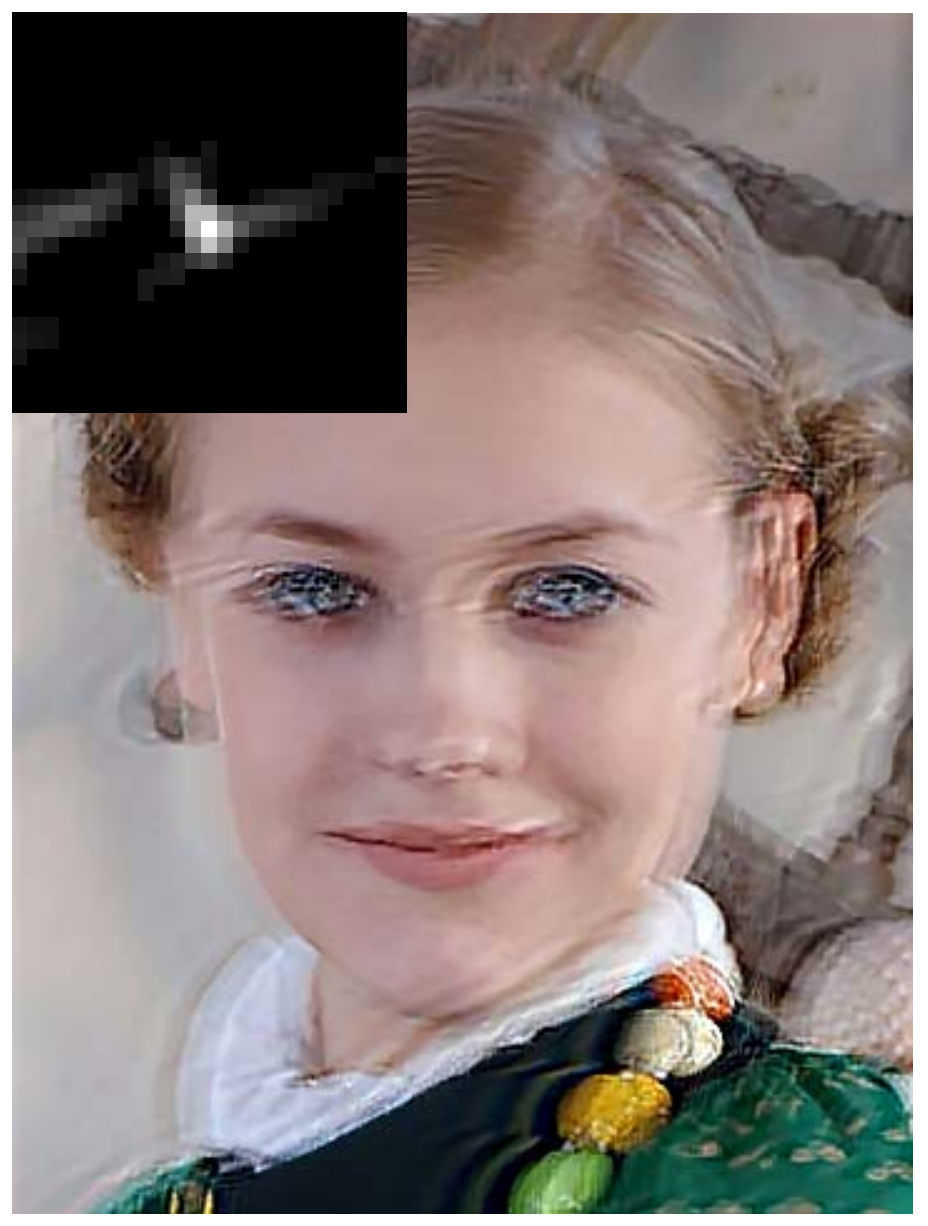} & \hspace{-0.4cm}
			\includegraphics[width=0.19\linewidth, height=0.23\linewidth]{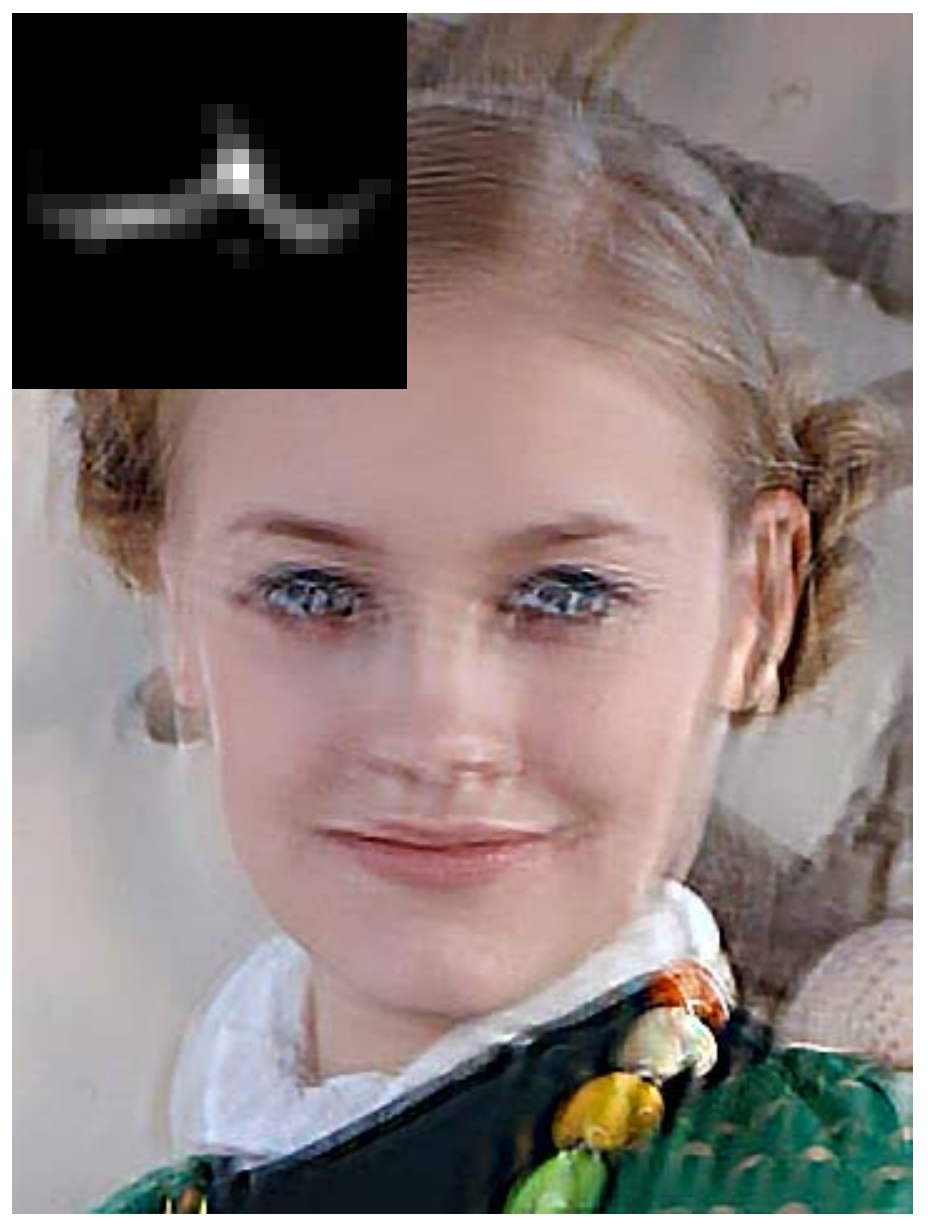} & \hspace{-0.4cm}
			\includegraphics[width=0.19\linewidth, height=0.23\linewidth]{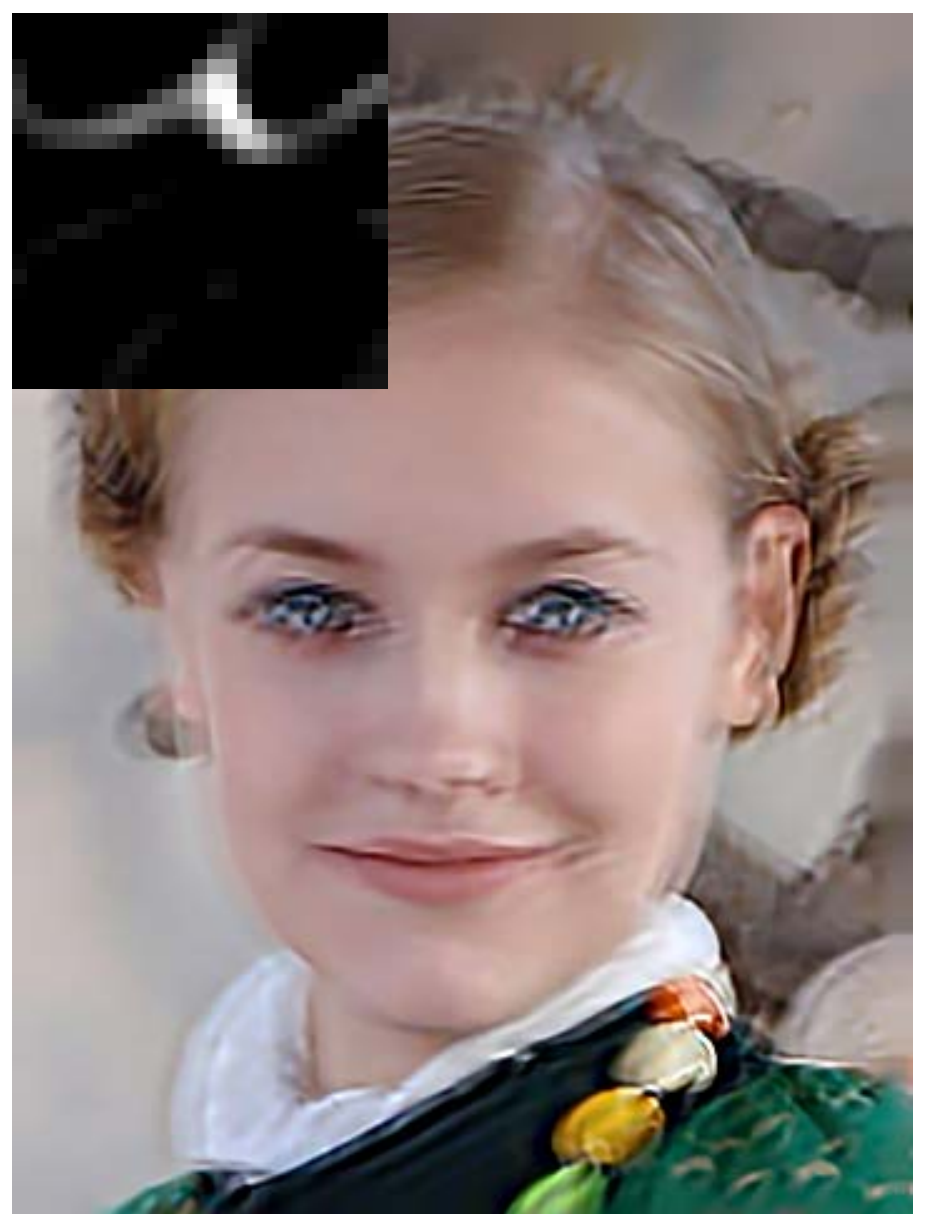} & \hspace{-0.4cm}
			\includegraphics[width=0.19\linewidth, height=0.23\linewidth]{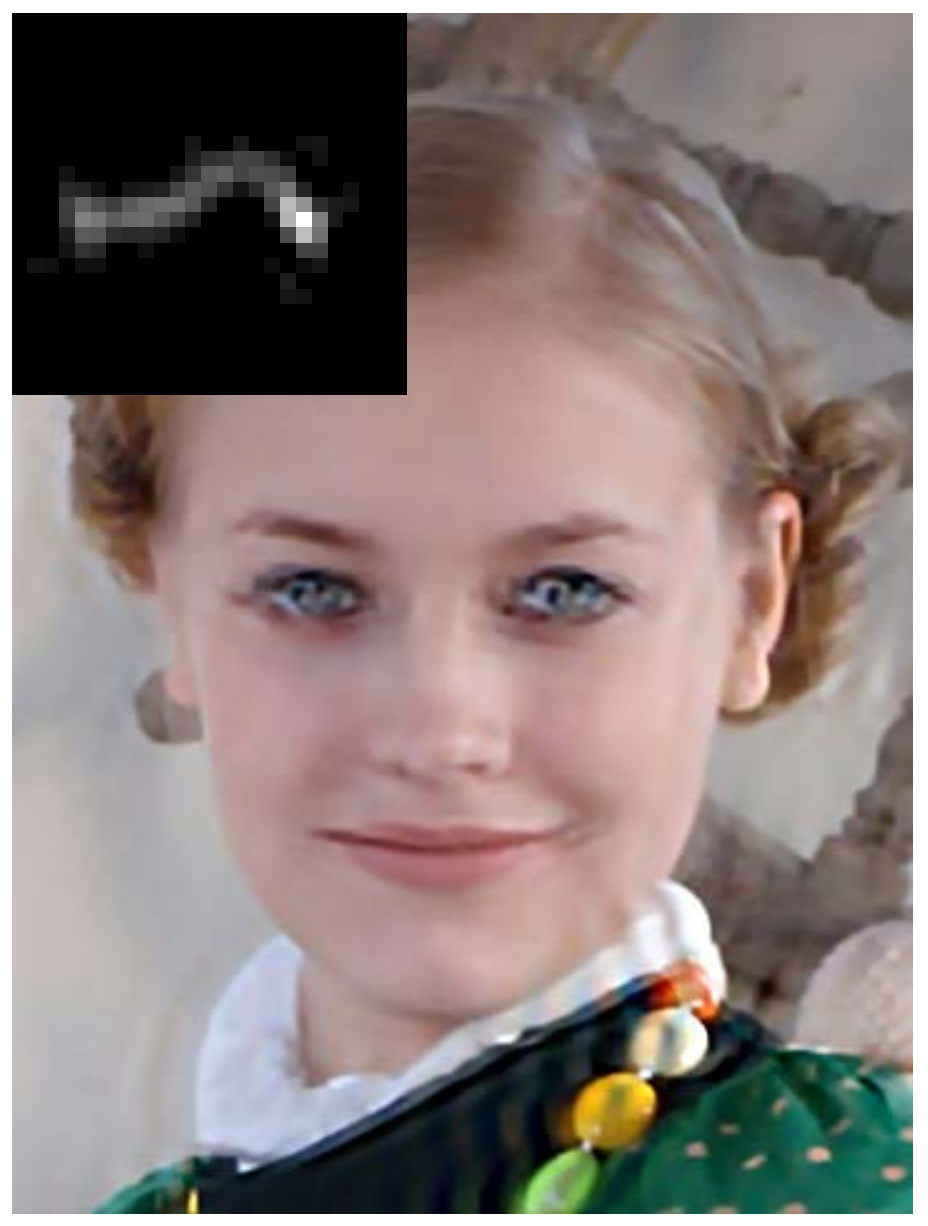} & \hspace{-0.4cm}
			\includegraphics[width=0.19\linewidth, height=0.23\linewidth]{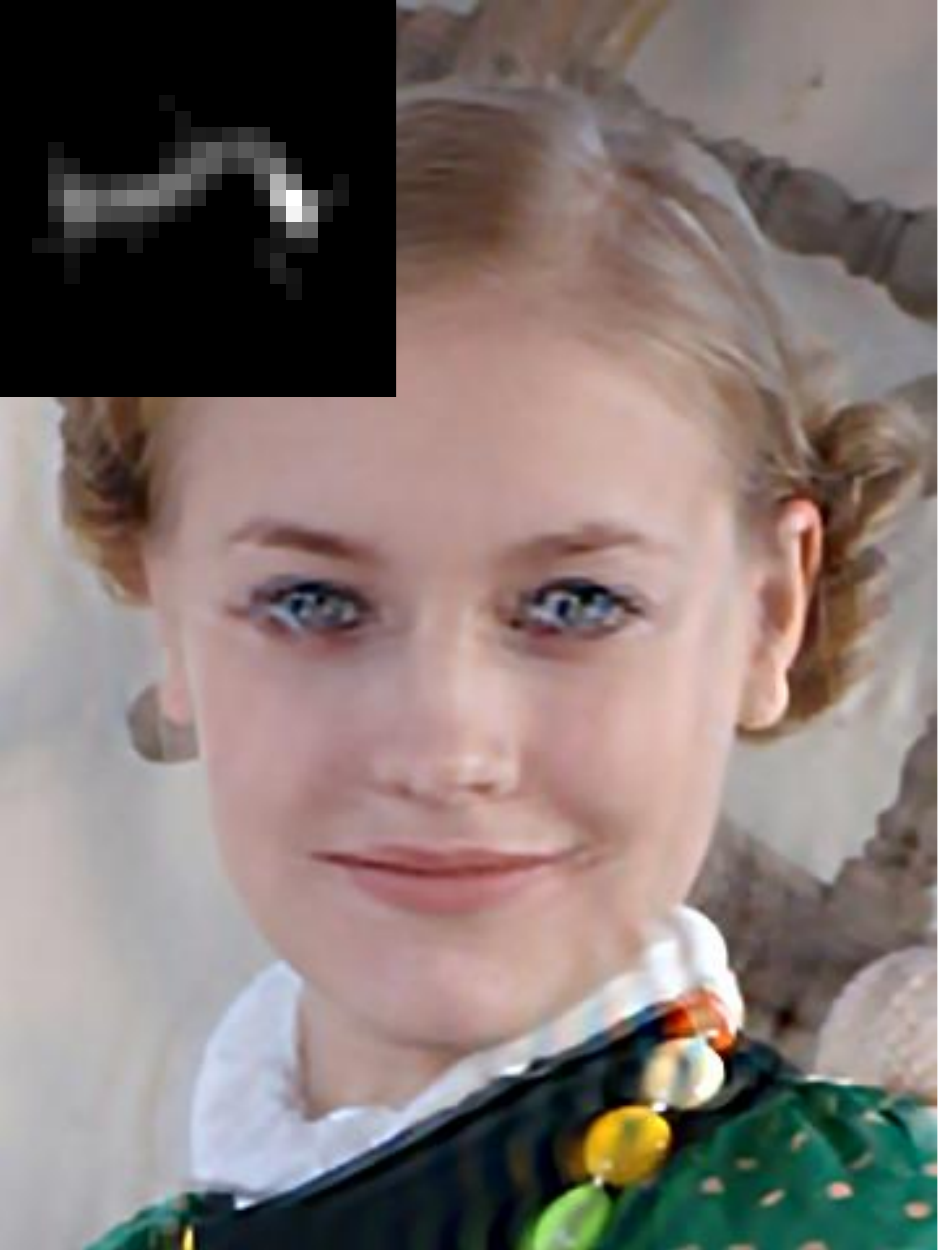} \\
			(f) Zhong~et al.~\cite{zhong/lin_cvpr2013/noise/deblur} &\hspace{-0.4cm} (g) Xu~et al.~\cite{Xu/l0deblur/cvpr2013} &\hspace{-0.4cm} (h) Michaeli and Irani~\cite{tomer/eccv/MichaeliI14} &\hspace{-0.4cm} (i) Our exemplar-based &\hspace{-0.4cm} (j) Our CNN-based\\
		\end{tabular}
	\end{center}
	\vspace{-0.3cm}
\caption{Example of real captured image.
	The estimated kernel is of $25\times25$ pixels.
}
\label{fig: real-imag2}
\end{figure*}

Fig.~\ref{fig: real-imag2}(a) shows another example of a real captured image.
The deblurring methods based on edge selection~\cite{Cho/et/al,Xu/et/al,libin/sun/patchdeblur_iccp2013} do not perform well
as ambiguous edges are selected for kernel estimation.
Similarly, the deblurred images by the methods based on natural priors~\cite{Xu/l0deblur/cvpr2013,tomer/eccv/MichaeliI14}
contain artifacts, while the exemplar-based and CNN-based methods
generate sharper contents as shown in Fig.~\ref{fig: real-imag2}(i) and (j).

%
\begin{figure*}[!t]\footnotesize
\begin{center}
\begin{tabular}{ccccc}
\includegraphics[width=0.19\linewidth, height=0.13\linewidth]{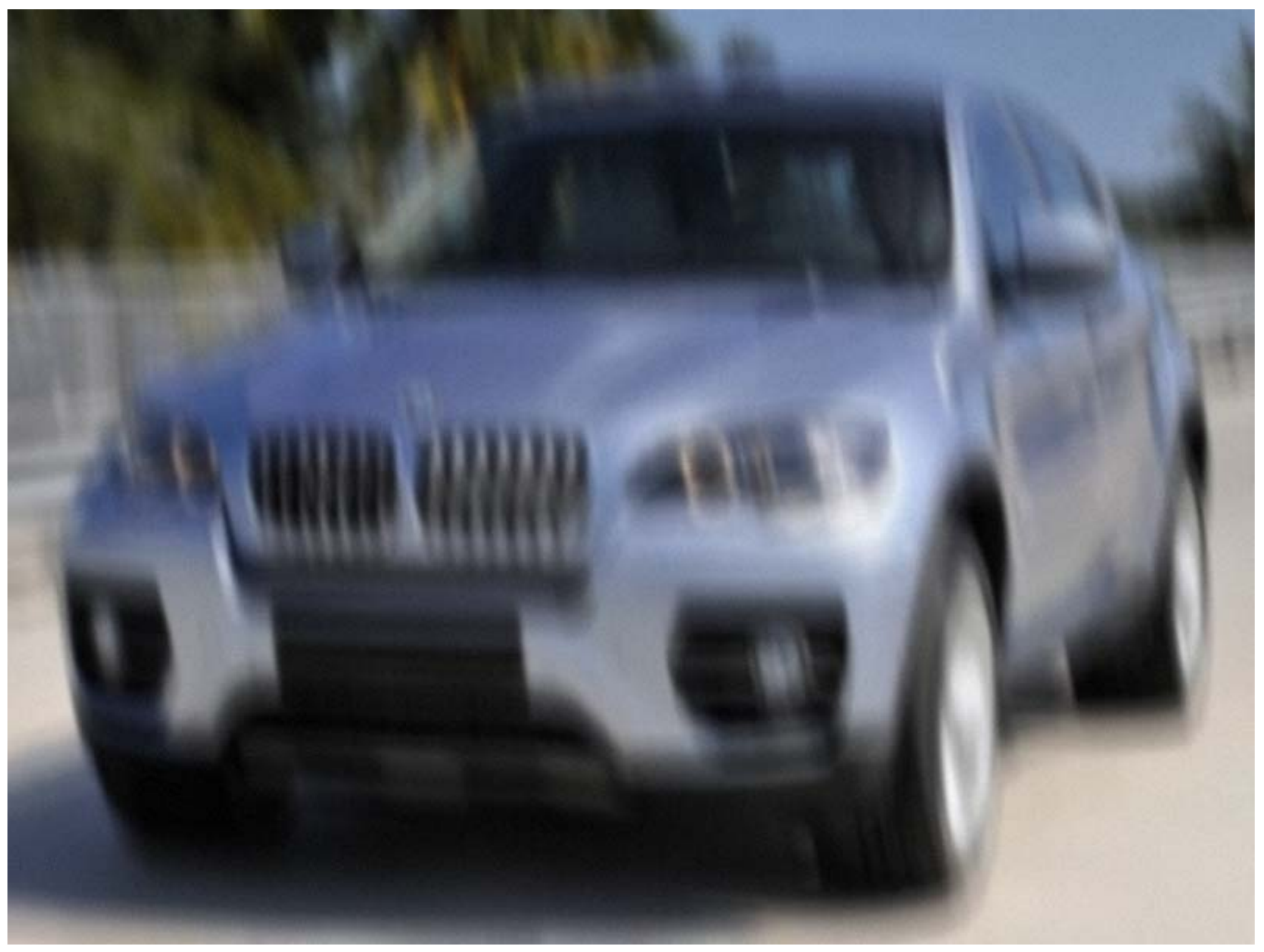} & \hspace{-0.48cm}
\includegraphics[width=0.19\linewidth, height=0.13\linewidth]{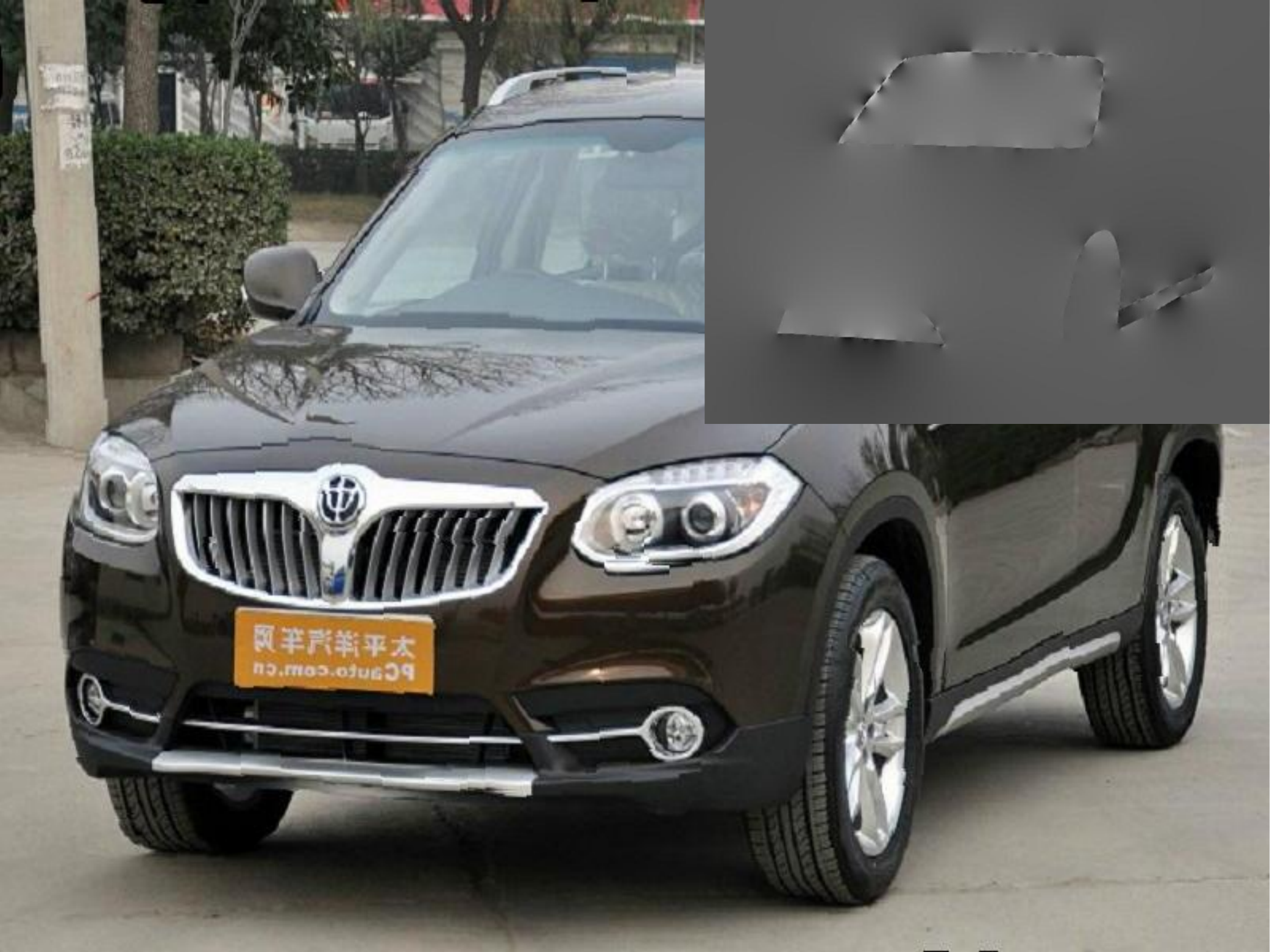} & \hspace{-0.48cm}
\includegraphics[width=0.19\linewidth, height=0.13\linewidth]{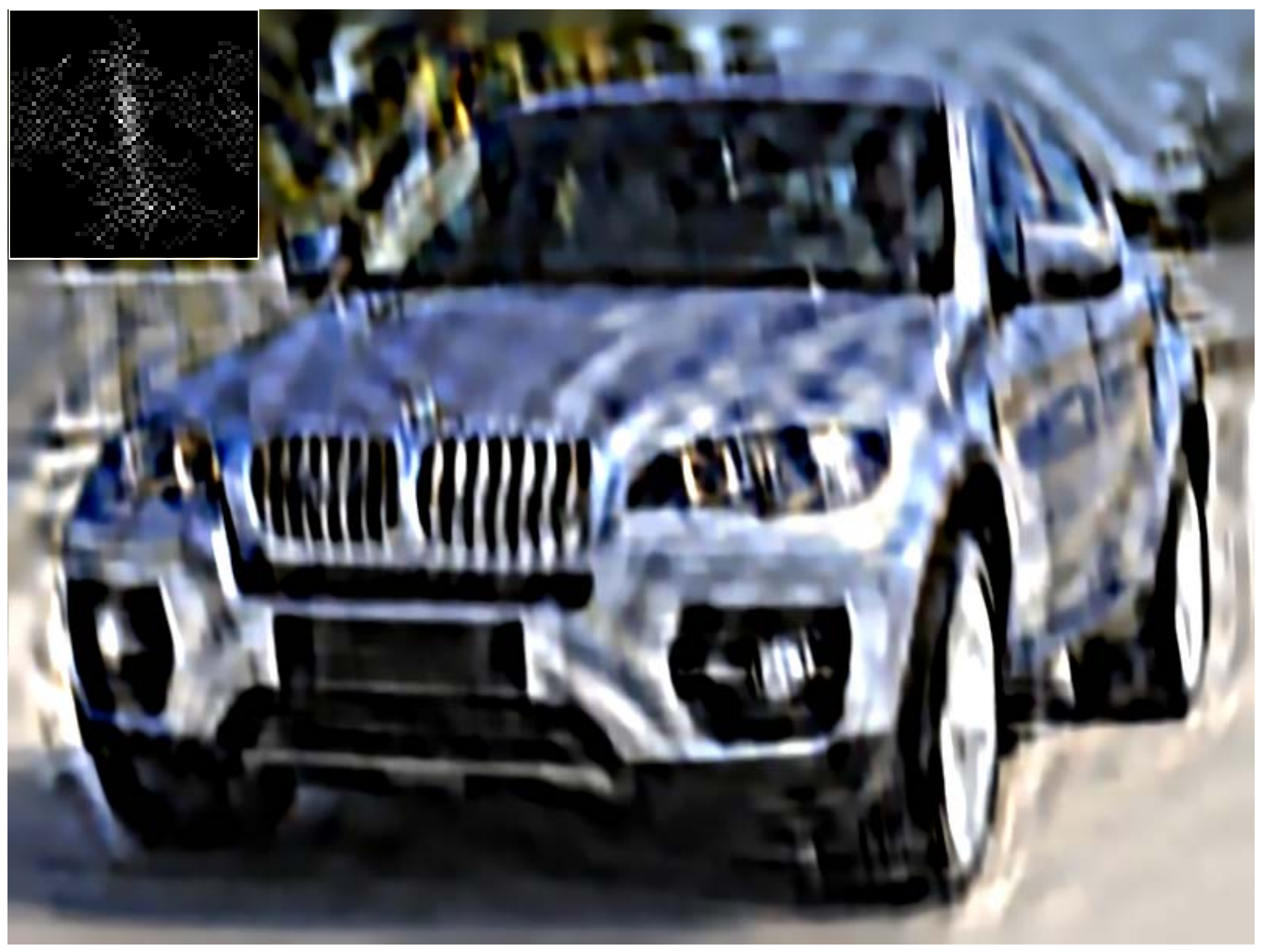} & \hspace{-0.48cm}
\includegraphics[width=0.19\linewidth, height=0.13\linewidth]{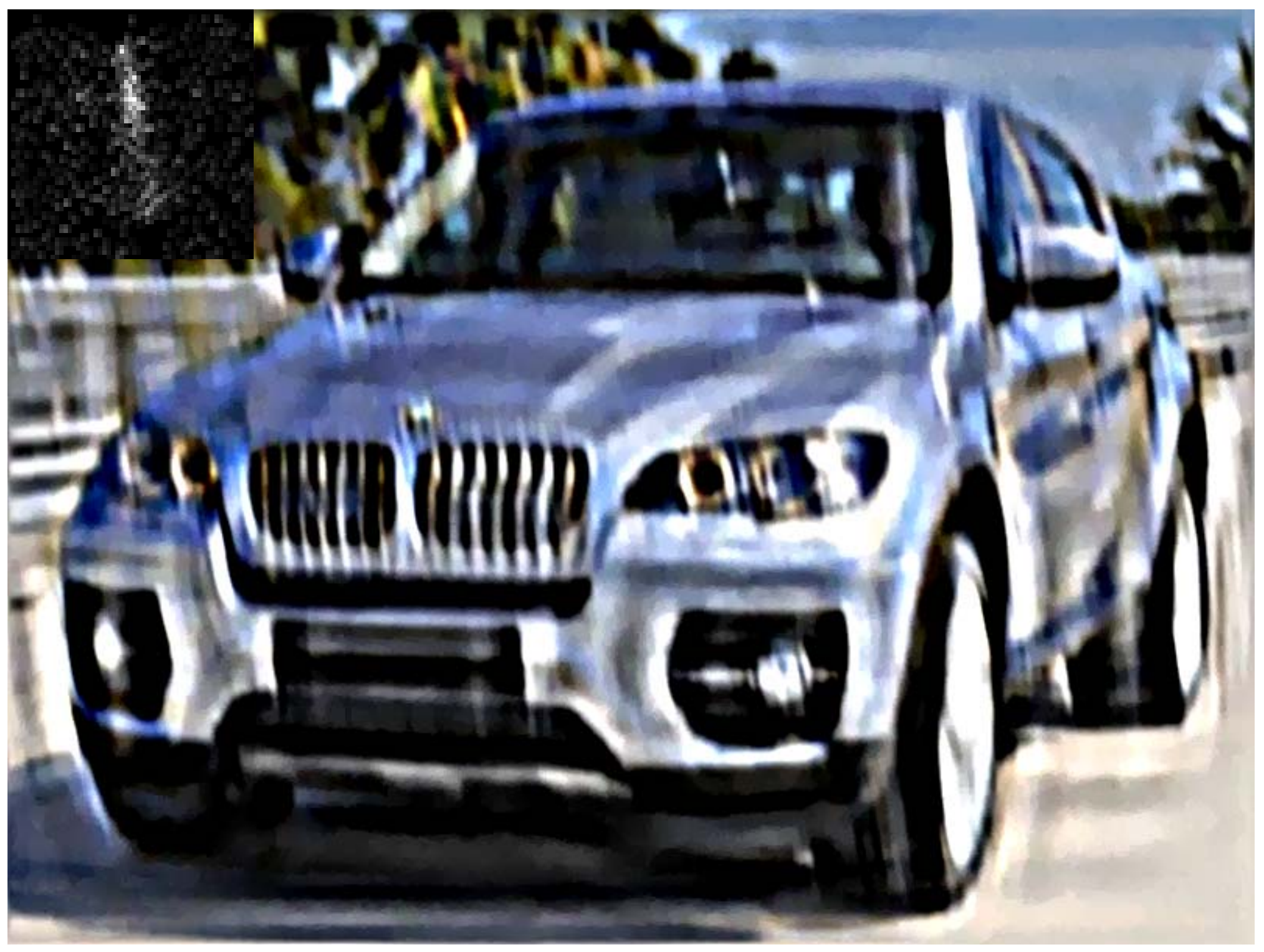} & \hspace{-0.48cm}
\includegraphics[width=0.19\linewidth, height=0.13\linewidth]{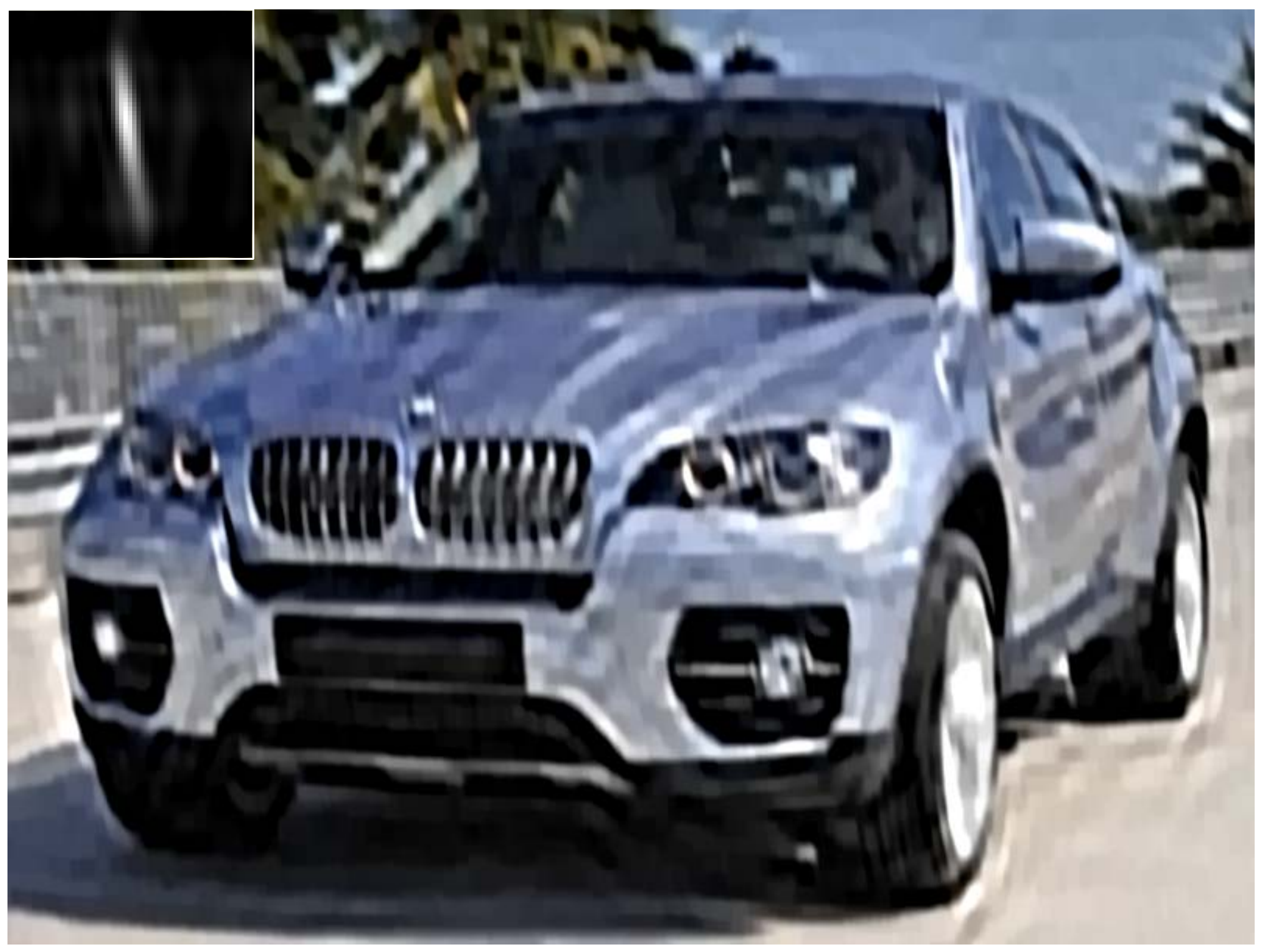} \\
(a) Input &\hspace{-0.45cm} (b) Exemplar-based $\nabla S$ &\hspace{-0.45cm} (c)  Sun et al.~\cite{libin/sun/patchdeblur_iccp2013} & \hspace{-0.45cm}  (d) Cho and Lee~\cite{Cho/et/al} &\hspace{-0.45cm} (e) Krishnan~et al.~\cite{Krishnan/CVPR2011} \\
\hfill
\includegraphics[width=0.19\linewidth, height=0.13\linewidth]{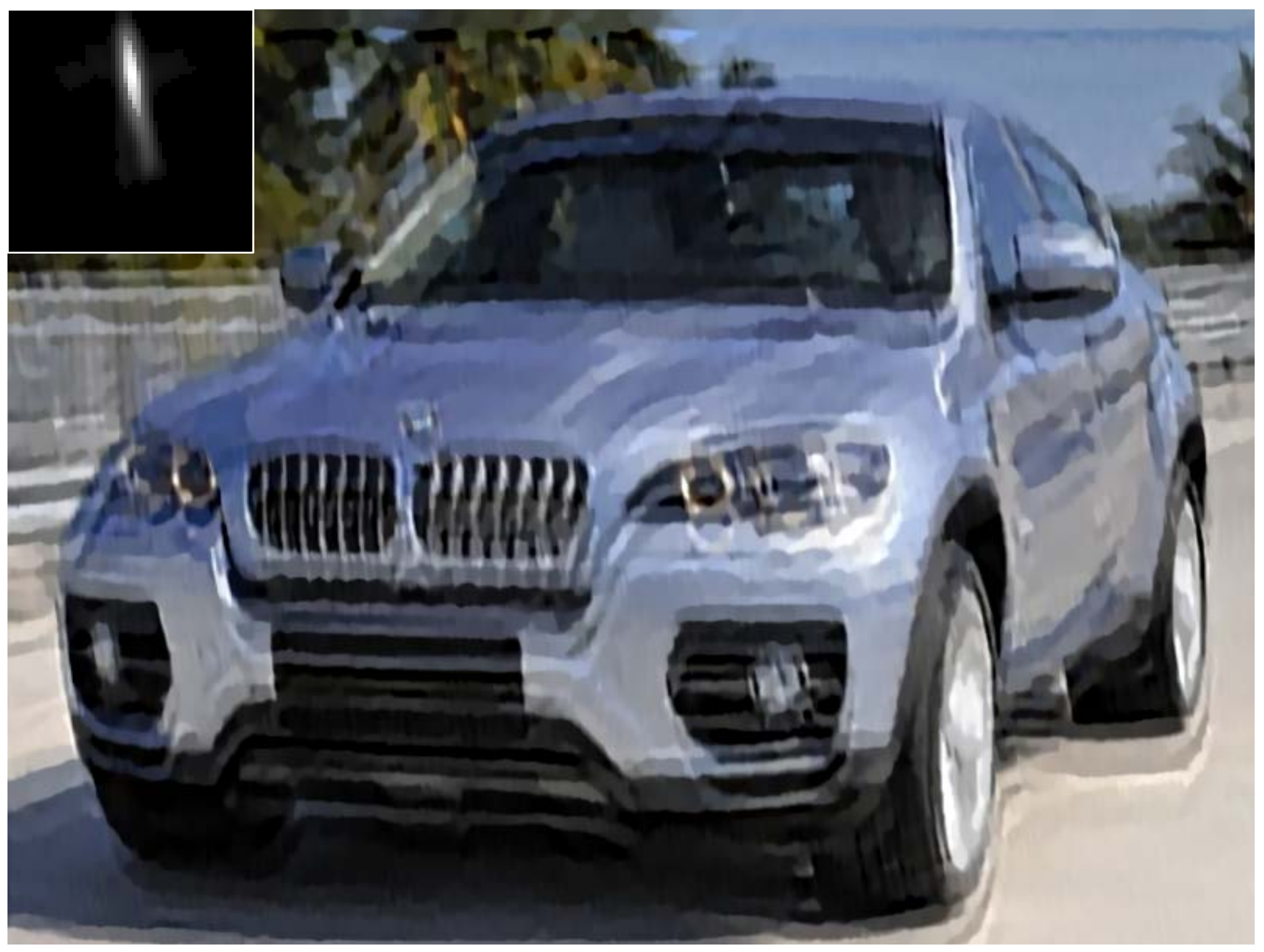} & \hspace{-0.48cm}
\includegraphics[width=0.19\linewidth, height=0.13\linewidth]{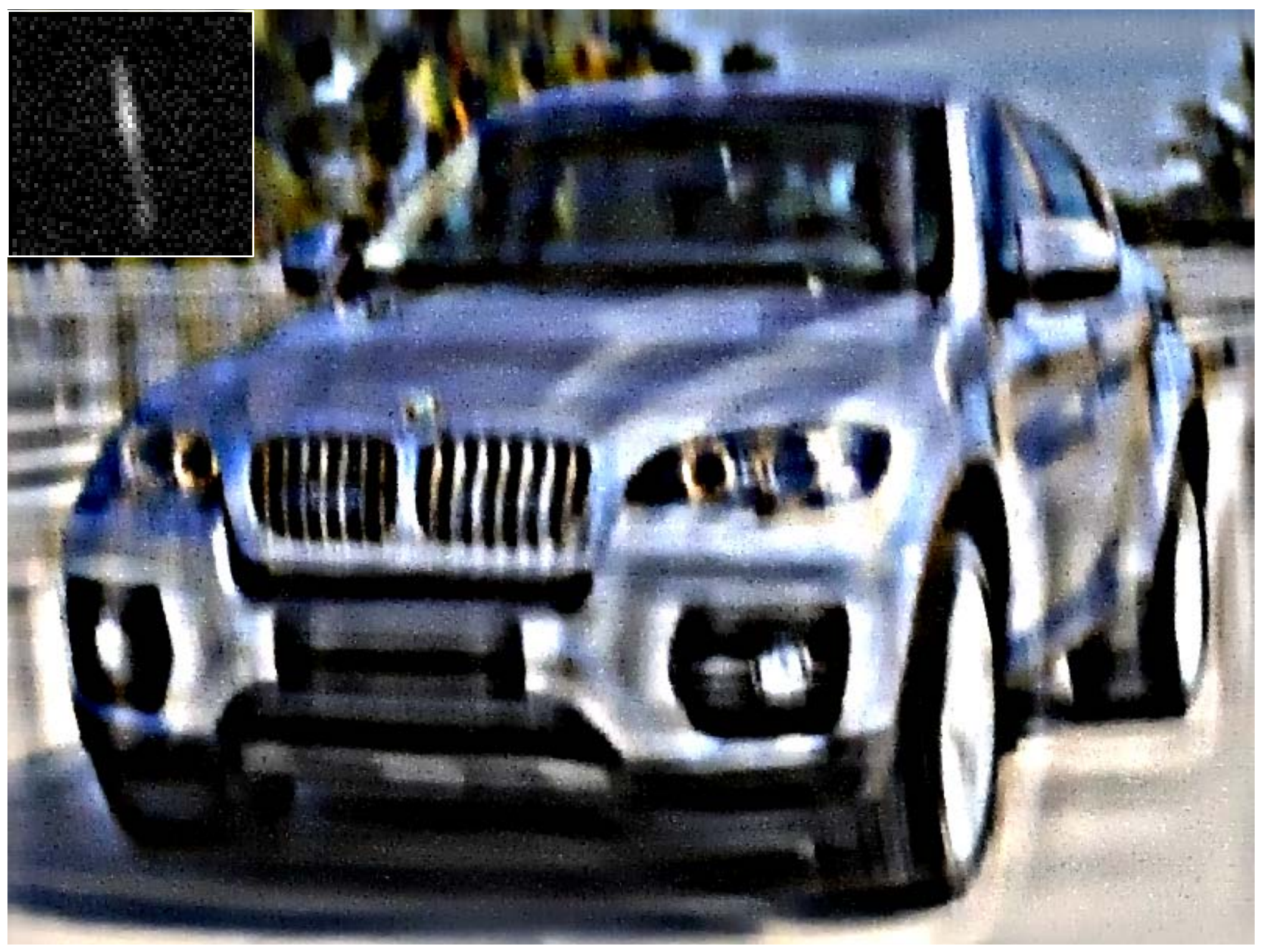} & \hspace{-0.48cm}
\includegraphics[width=0.19\linewidth, height=0.13\linewidth]{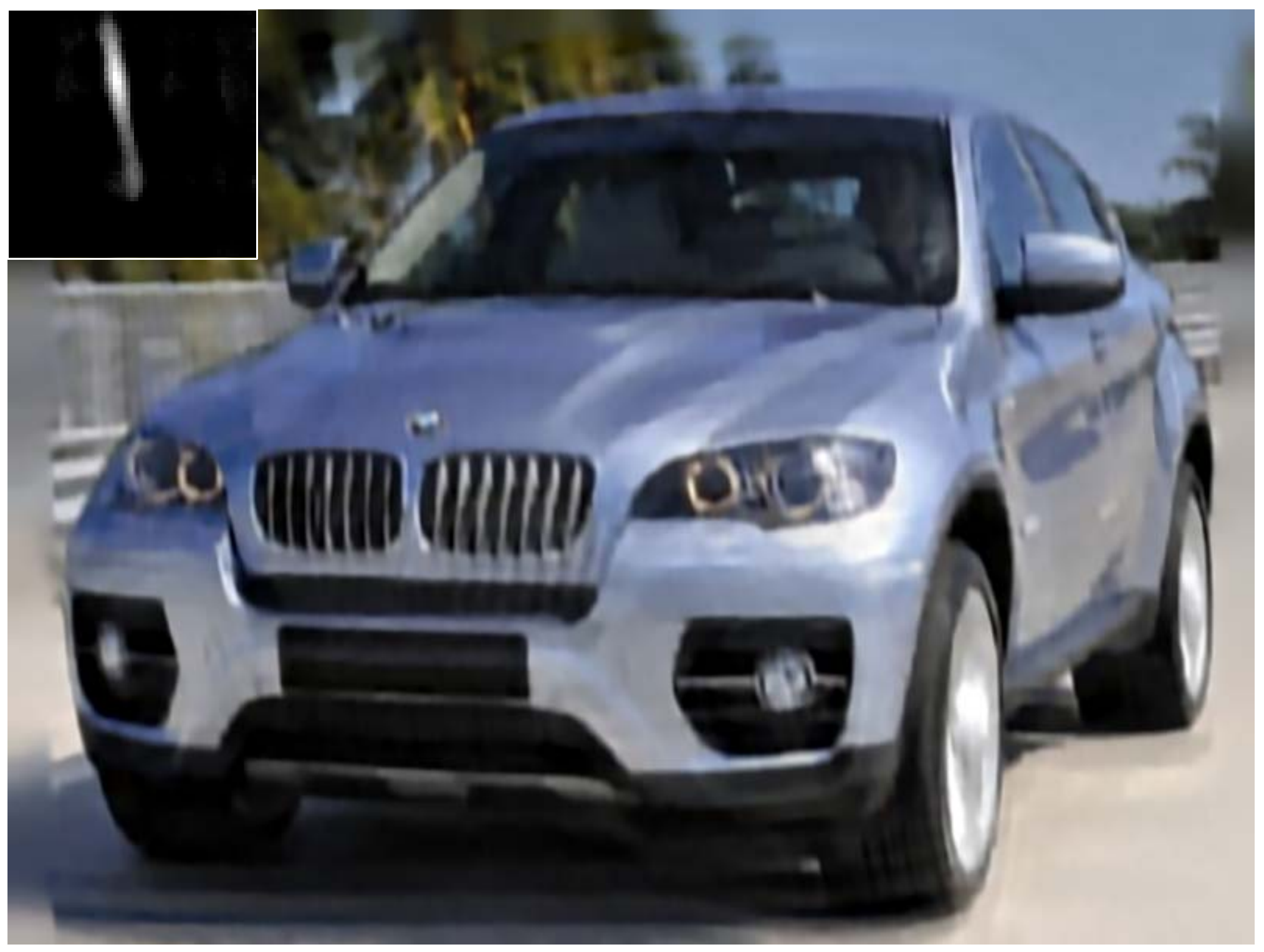} & \hspace{-0.48cm}
\includegraphics[width=0.19\linewidth, height=0.13\linewidth]{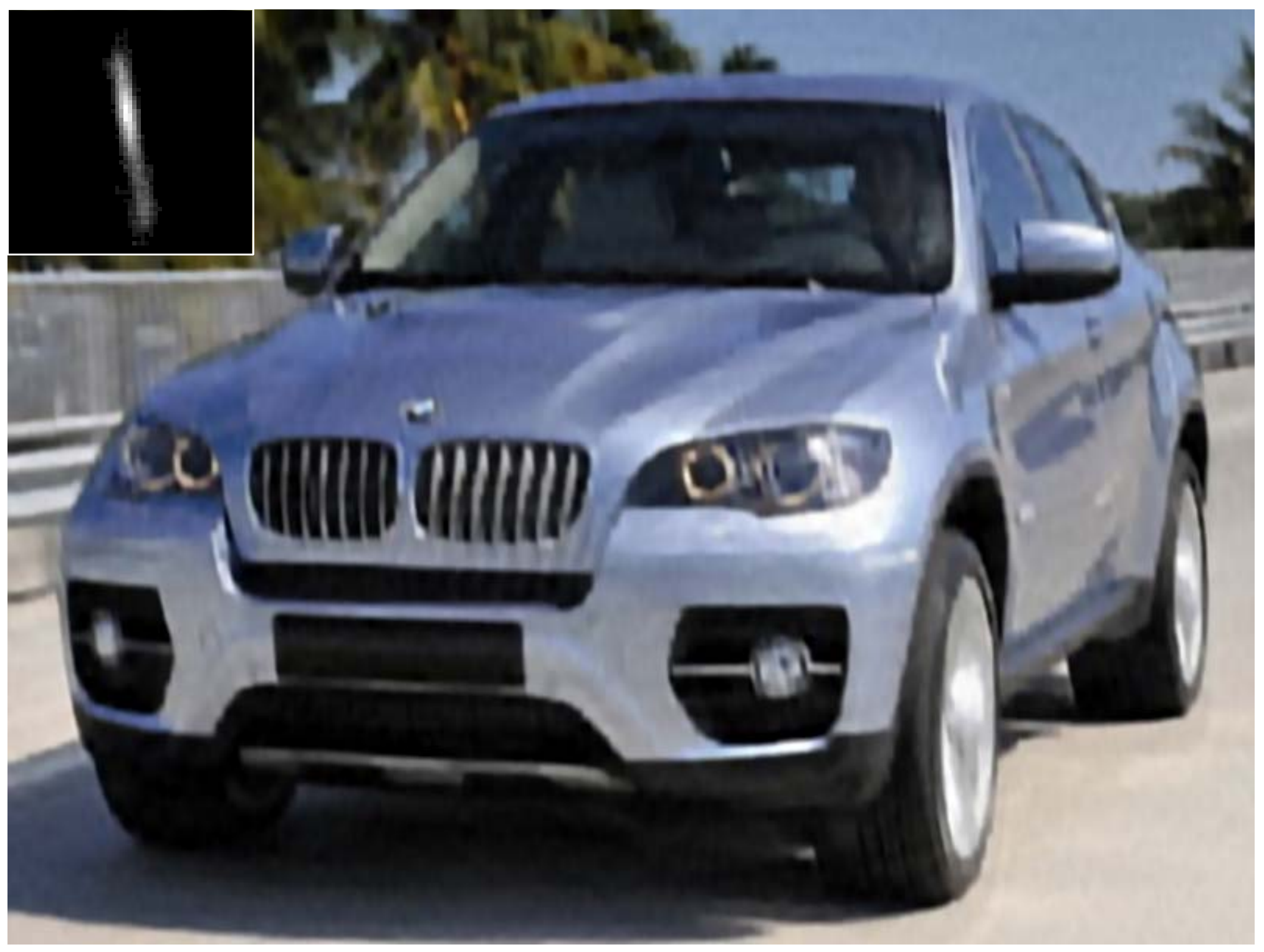} & \hspace{-0.48cm}
\includegraphics[width=0.19\linewidth, height=0.13\linewidth]{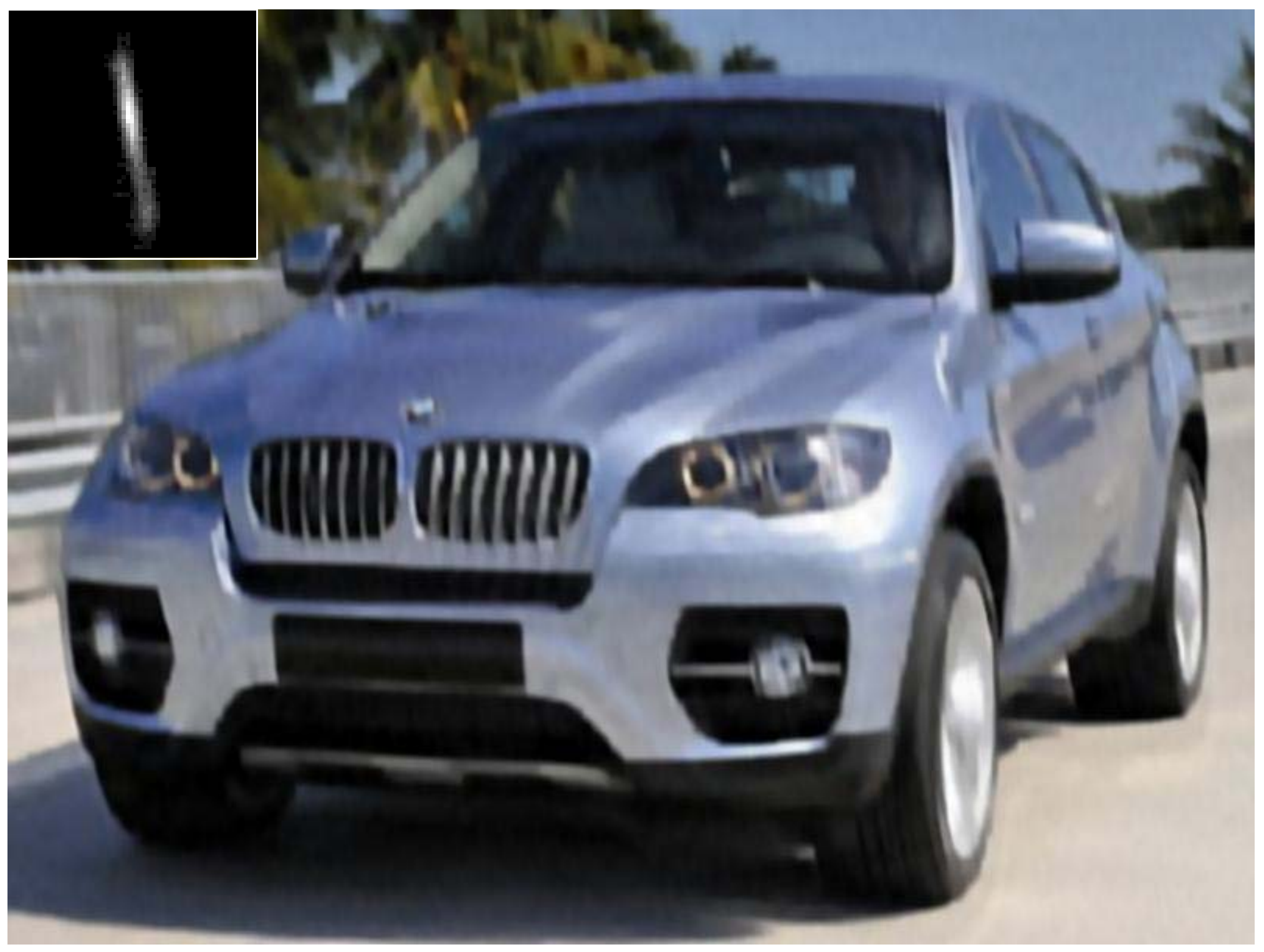} \\
(f) Zhong~et al.~\cite{zhong/lin_cvpr2013/noise/deblur} &\hspace{-0.45cm} (g) Xu~et al.~\cite{Xu/l0deblur/cvpr2013} &\hspace{-0.45cm} (h) Michaeli and Irani~\cite{tomer/eccv/MichaeliI14} &\hspace{-0.45cm} (i) Our exemplar-based &\hspace{-0.45cm} (j) Our CNN-based\\
		\end{tabular}
	\end{center}
	\vspace{-0.3cm}
	\caption{Object deblurring.
	Our method generates the deblurred
		result with fewer ringing artifacts.
	}
	\label{fig: extension-examples-2}
\end{figure*}

\vspace{-3mm}
\subsection{Object Deblurring}
\vspace{-1mm}
In this work, we focus on face image deblurring, as it is of great interest with numerous applications.
However, the proposed methods can be applied to other deblurring tasks
by using exemplars of specific classes with categorical structures.
We use one example in Fig.~\ref{fig: extension-examples-2} to show the proposed methods can be extended to object deblurring.

Similar to face deblurring,
we first collect a set of exemplar images and restore categorical structures (e.g., car body, windows and wheels for car images) using the method described in Section~\ref{ssec: Learning Framework}.
For each test image, we use~\eqref{eq: normalized-cross-correlation} to
find the best exemplar image as shown in Fig.~\ref{fig: extension-examples-2}(b) and compute salient edges according to~\eqref{eq: final-structure}.
Finally, we use the same algorithm (Algorithm~\ref{alg:kernel-estimation-algorithm})
for object deblurring.
For the CNN-based method, we first use the exemplars to generate blurred images and sharp edges using the method in Section~\ref{sec: Deep Learning Framework},
and then train a network based on the synthetic data.

The results generated by~\cite{Cho/et/al,Krishnan/CVPR2011,libin/sun/patchdeblur_iccp2013} contain significant
ringing artifacts as shown in Fig.~\ref{fig: extension-examples-2}(c)-(e) and (g).
In addition, the deblurred results by the state-of-the-art methods~\cite{zhong/lin_cvpr2013/noise/deblur,Xu/l0deblur/cvpr2013,tomer/eccv/MichaeliI14} contain blurry regions as shown in
Fig.~\ref{fig: extension-examples-2}(f) and (h).
%
In contrast, the results generated by our exemplar-based (Fig.~\ref{fig: extension-examples-2}(i)) and CNN-based (Fig.~\ref{fig: extension-examples-2}(j)) methods are sharper with significantly fewer artifacts.

\vspace{-3mm}
\subsection{Natural Image Deblurring}
\begin{figure*}[!t]\footnotesize
\begin{center}
\begin{tabular}{cccccc}
\hspace{-0.4cm}
\includegraphics[width = 0.158\linewidth]{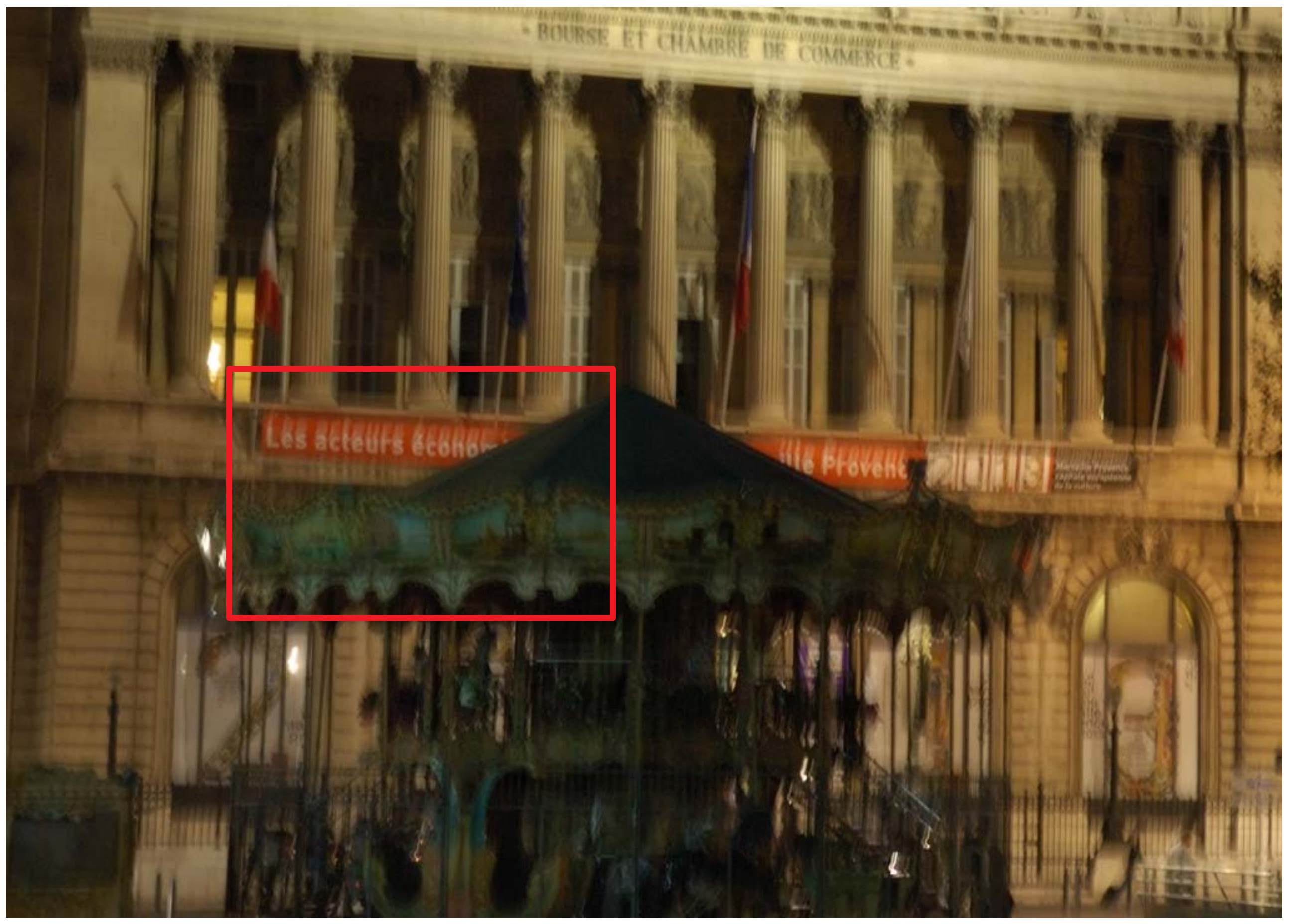} & \hspace{-0.48cm}
\includegraphics[width = 0.158\linewidth]{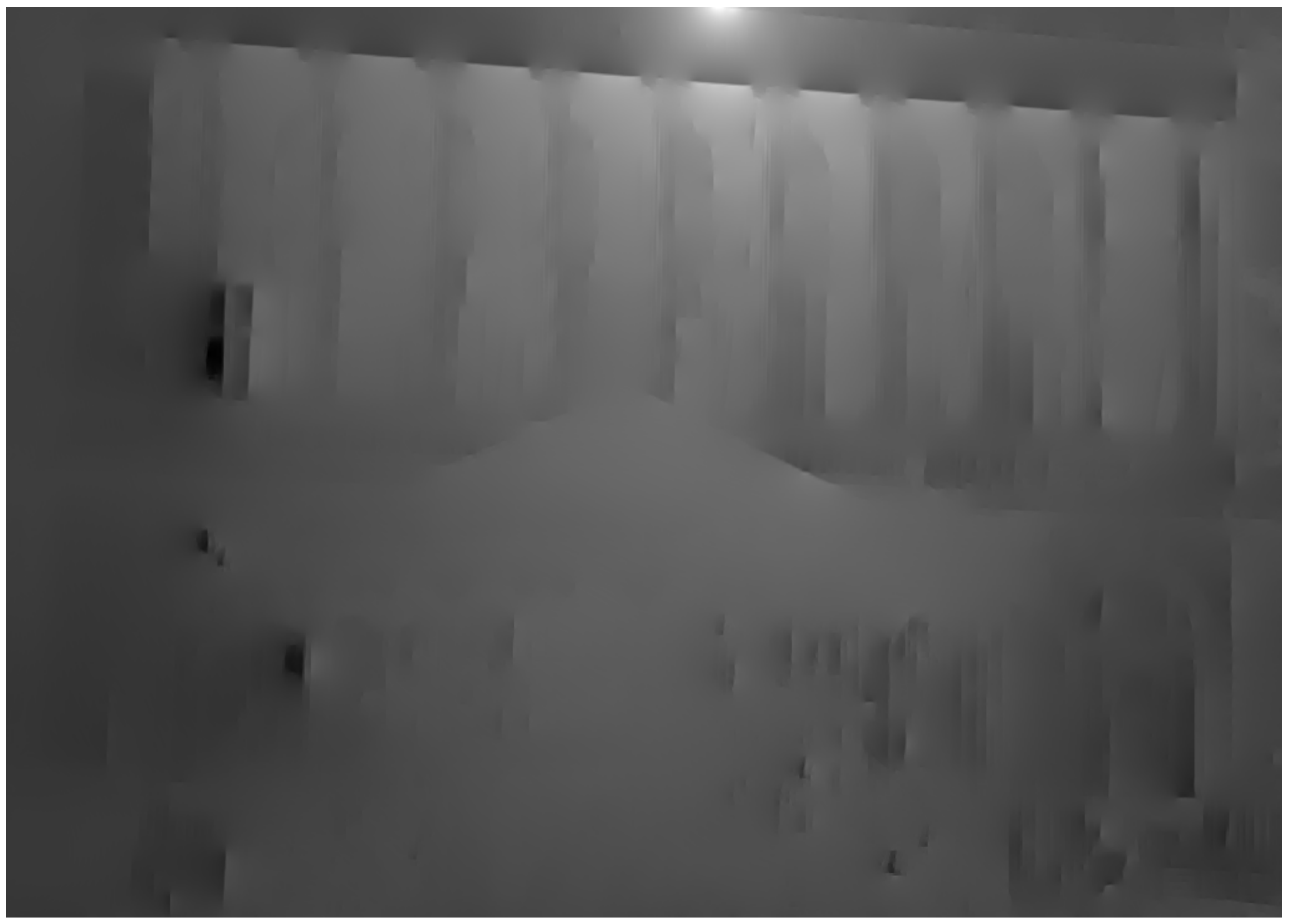} & \hspace{-0.48cm}
\includegraphics[width = 0.168\linewidth]{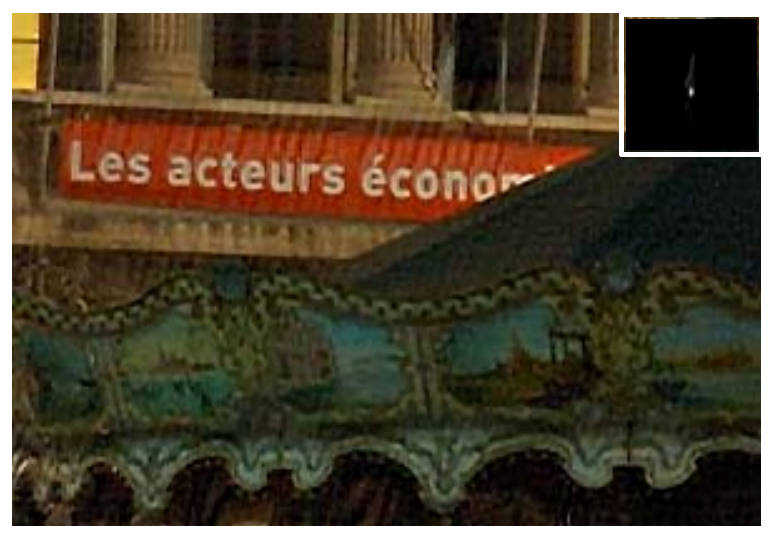} & \hspace{-0.48cm}
\includegraphics[width = 0.168\linewidth]{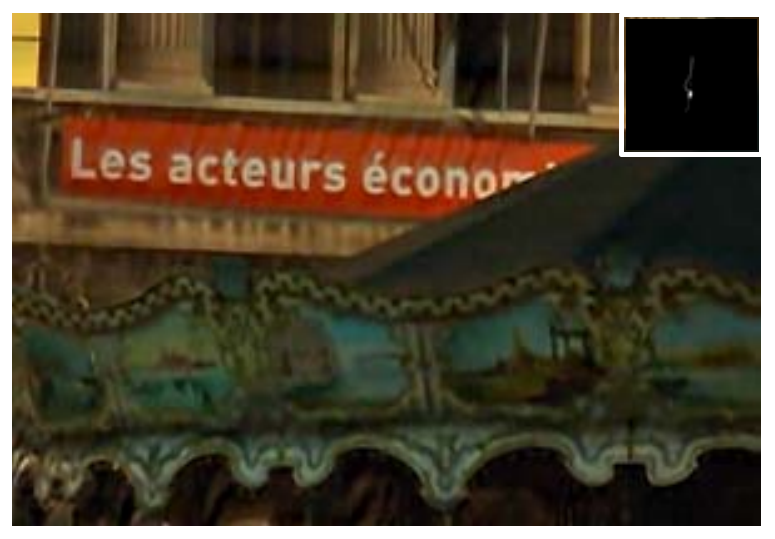} & \hspace{-0.48cm}
\includegraphics[width = 0.168\linewidth]{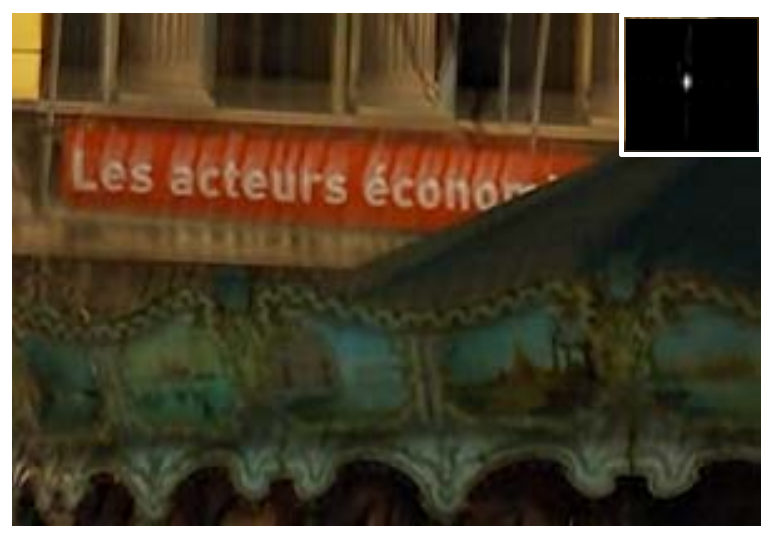} & \hspace{-0.48cm}
\includegraphics[width = 0.168\linewidth]{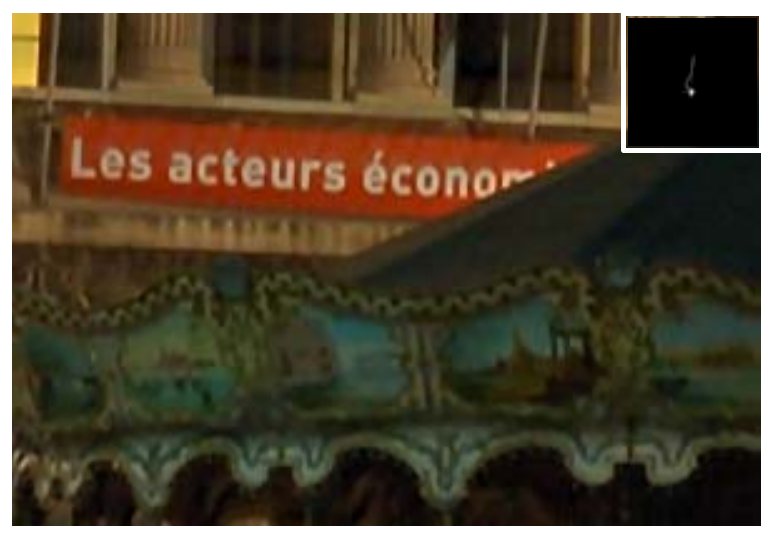} \\
 (a) Input &\hspace{-0.48cm} (b)  Predicted $\nabla S$ &\hspace{-0.48cm} (c) Xu and Jia~\cite{Xu/et/al} &\hspace{-0.48cm} (d)  Sun et al.~\cite{libin/sun/patchdeblur_iccp2013} &\hspace{-0.48cm} (e)  {\scriptsize Michaeli and Irani~\cite{tomer/eccv/MichaeliI14}} &\hspace{-0.48cm} (f)  Ours \\
\end{tabular}
\end{center}
\vspace{-0.3cm}
\caption{Natural image deblurring. Our CNN-based method can be applied to natural image deblurring and generates the image with few ringing artifacts and much clearer characters.}
\label{fig: natural-image-deblurring}
\end{figure*}

\vspace{-1mm}
In contrast to the exemplar-based method, the proposed CNN-based algorithm is not limited
to the structures of specific scenarios (e.g., poses).
Thus, it can be applied to deblur other images of object classes, e.g., natural scenes.
Fig.~\ref{fig: natural-image-deblurring} shows that the proposed CNN-based method is able to deblur natural images effectively.
Overall, the proposed method performs comparably against
the state-of-the-art natural deblurring algorithms~\cite{libin/sun/patchdeblur_iccp2013,tomer/eccv/MichaeliI14}.

\begin{figure*}[t]\footnotesize
	\begin{center}
		\begin{tabular}{ccccc}
			\includegraphics[width=0.19\linewidth, height=0.24\linewidth]{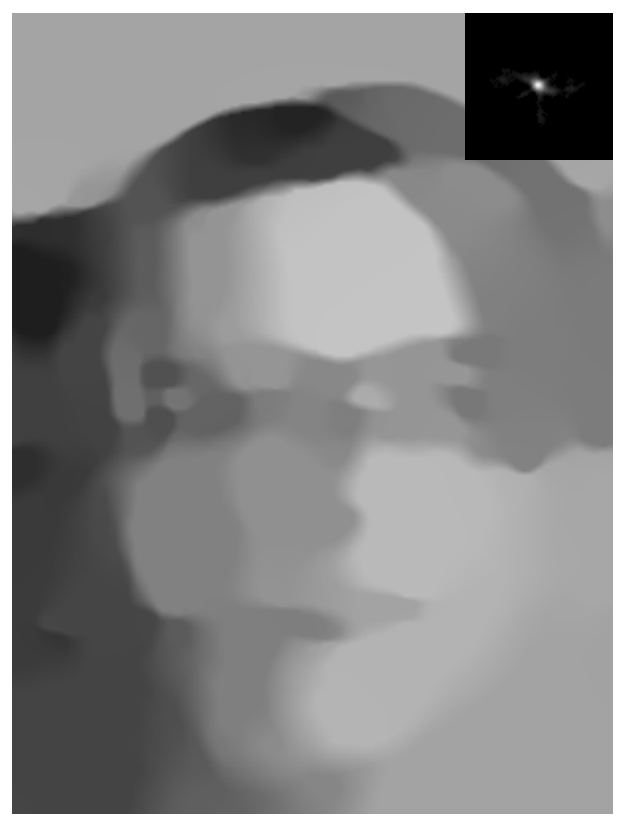} & \hspace{-0.5cm}
			\includegraphics[width=0.19\linewidth, height=0.24\linewidth]{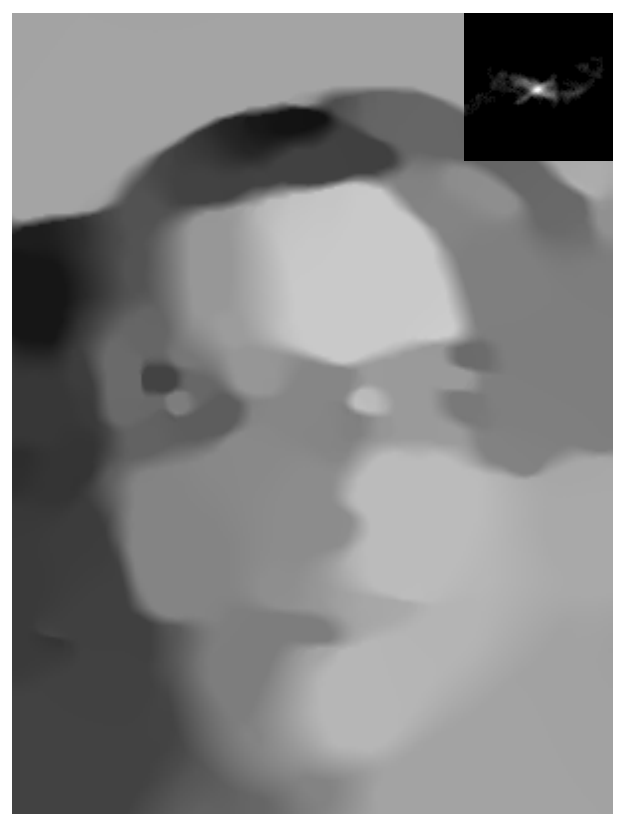} & \hspace{-0.5cm}
			\includegraphics[width=0.19\linewidth, height=0.24\linewidth]{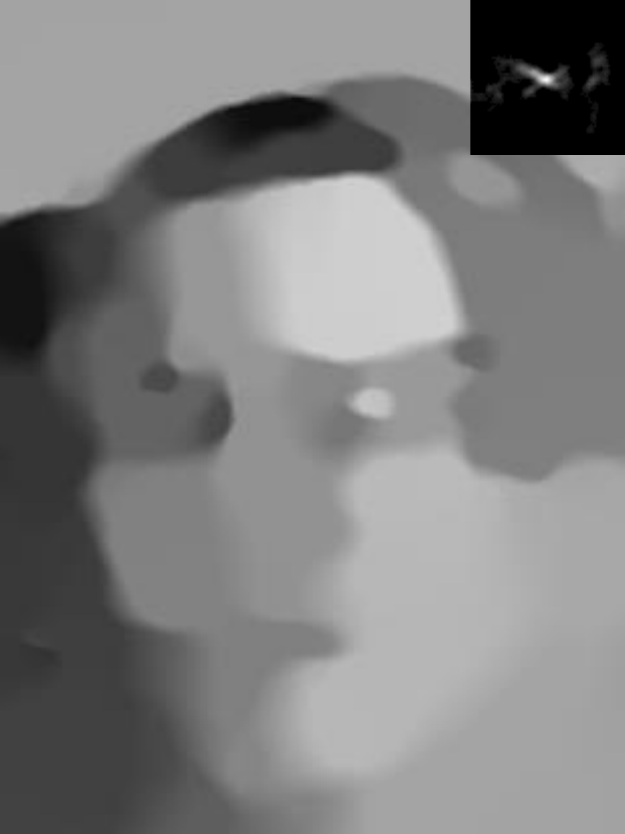} & \hspace{-0.5cm}
			\includegraphics[width=0.19\linewidth, height=0.24\linewidth]{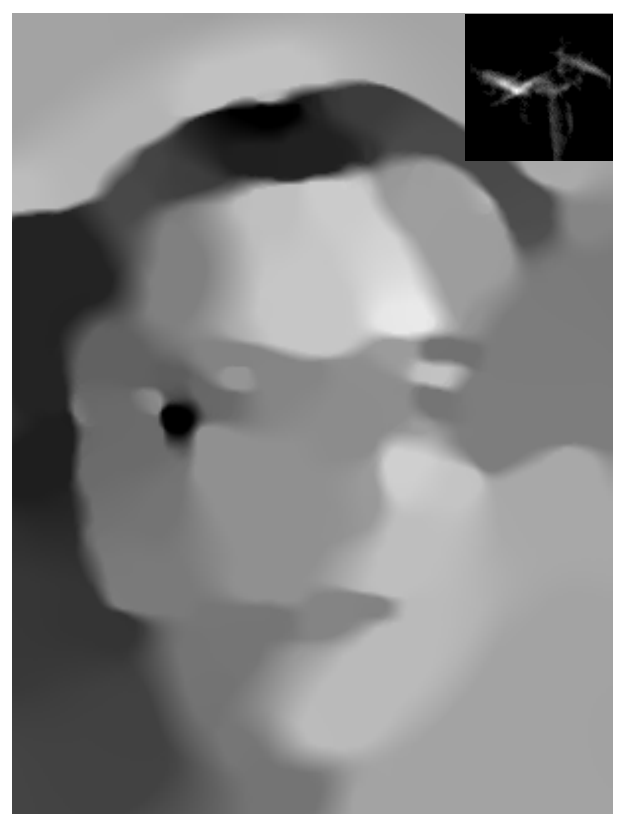} & \hspace{-0.5cm}
			\includegraphics[width=0.19\linewidth, height=0.24\linewidth]{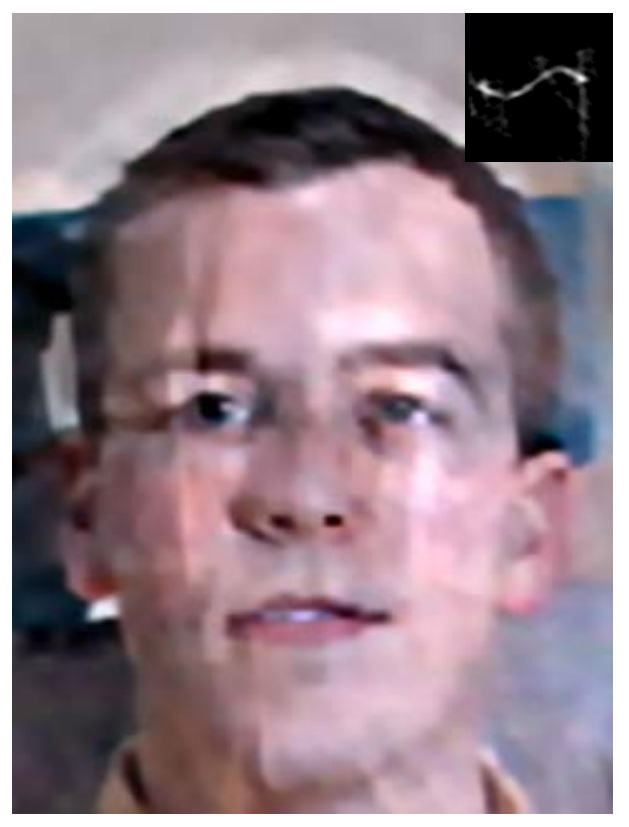}  \\
			(a) & \hspace{-0.5cm}  (b) & \hspace{-0.5cm}  (c)  & \hspace{-0.5cm} (d) & \hspace{-0.5cm} (e)\\
			\includegraphics[width=0.19\linewidth, height=0.24\linewidth]{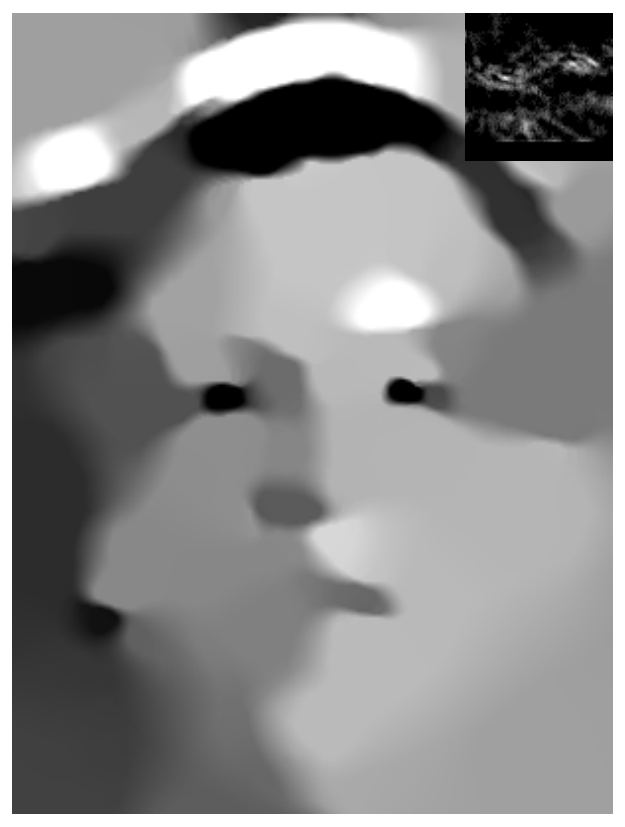} & \hspace{-0.5cm}
			\includegraphics[width=0.19\linewidth, height=0.24\linewidth]{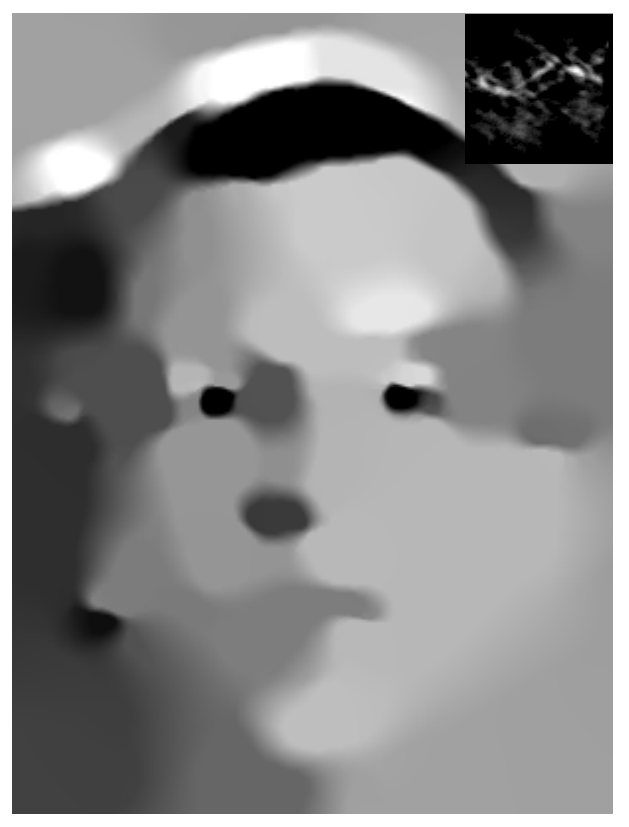} & \hspace{-0.5cm}
			\includegraphics[width=0.19\linewidth, height=0.24\linewidth]{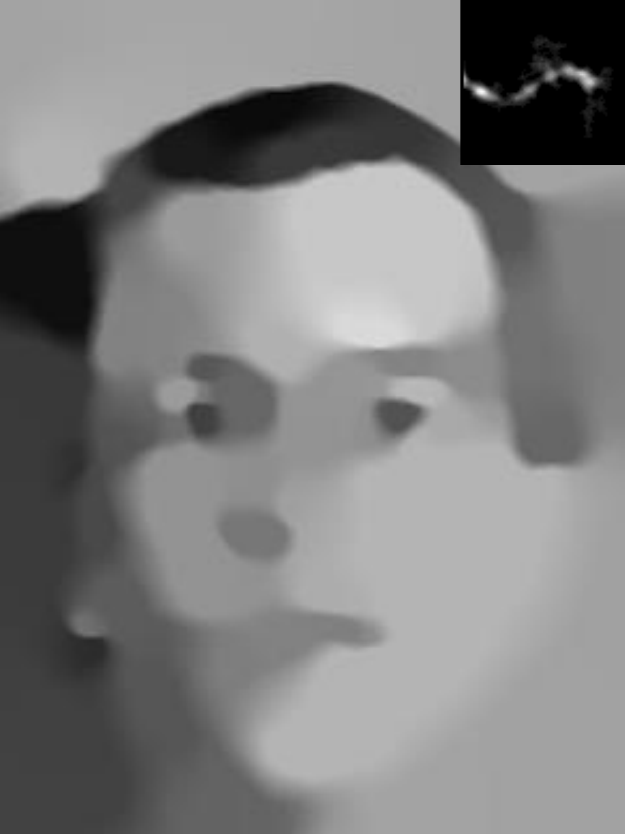} & \hspace{-0.5cm}
			\includegraphics[width=0.19\linewidth, height=0.24\linewidth]{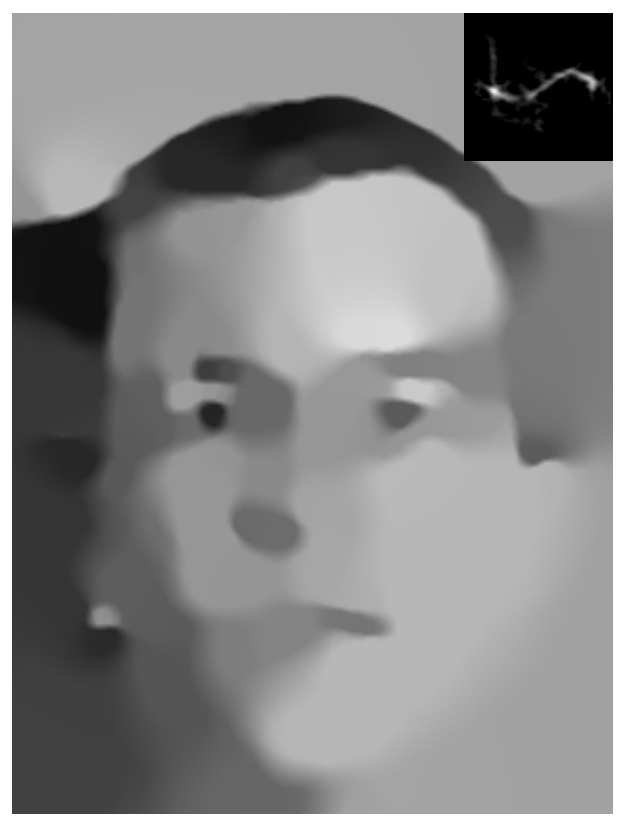} & \hspace{-0.5cm}
			\includegraphics[width=0.19\linewidth, height=0.24\linewidth]{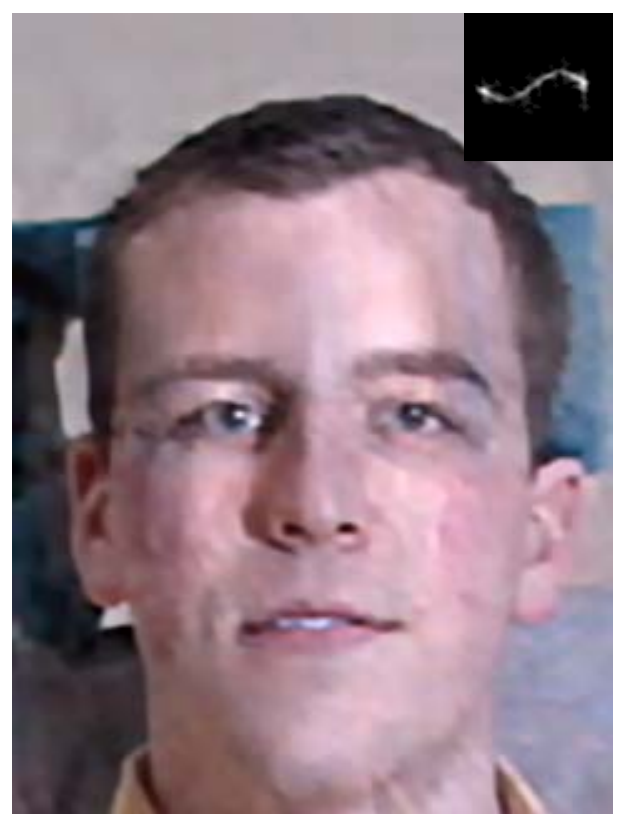} \\
			(f) & \hspace{-0.5cm}  (g) & \hspace{-0.5cm}  (h) & \hspace{-0.5cm} (i) & \hspace{-0.5cm} (j)\\
			\includegraphics[width=0.19\linewidth, height=0.24\linewidth]{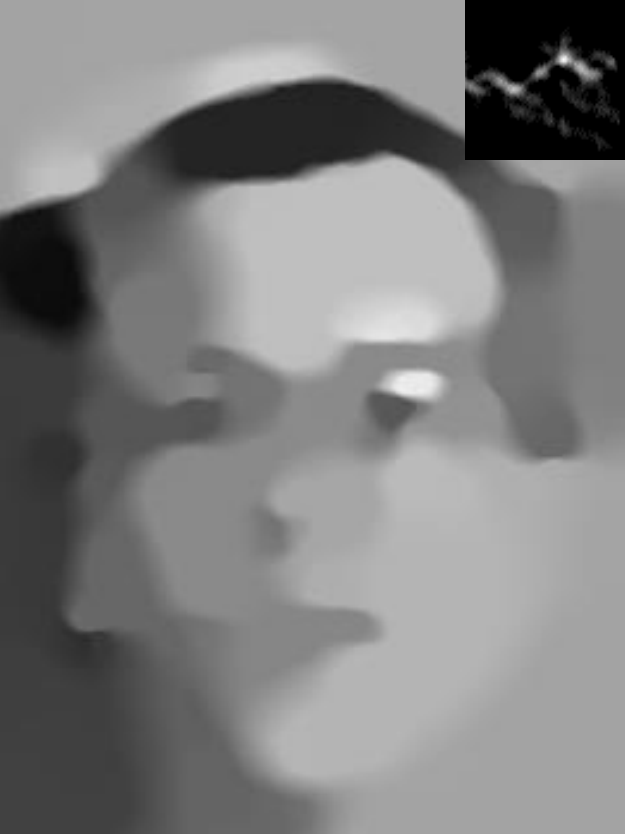} & \hspace{-0.5cm}
			\includegraphics[width=0.19\linewidth, height=0.24\linewidth]{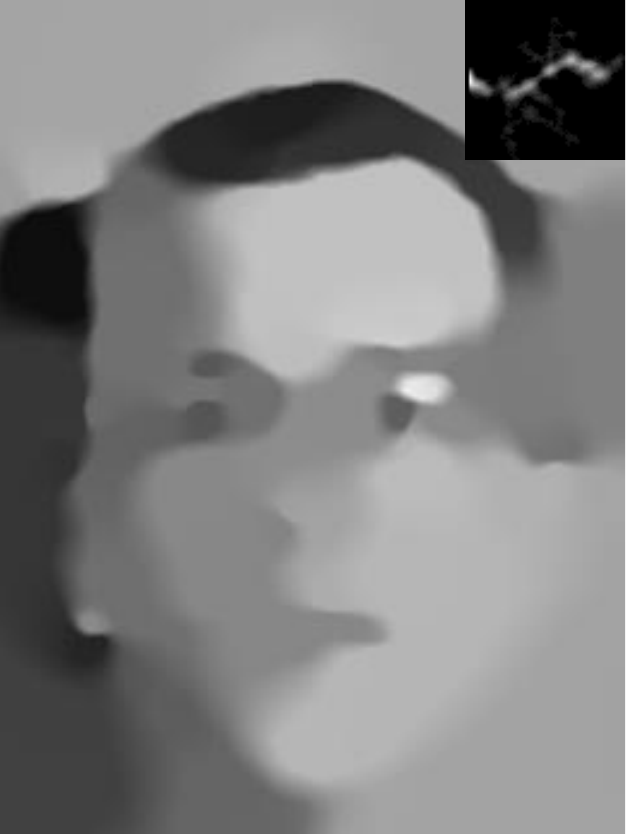} & \hspace{-0.5cm}
			\includegraphics[width=0.19\linewidth, height=0.24\linewidth]{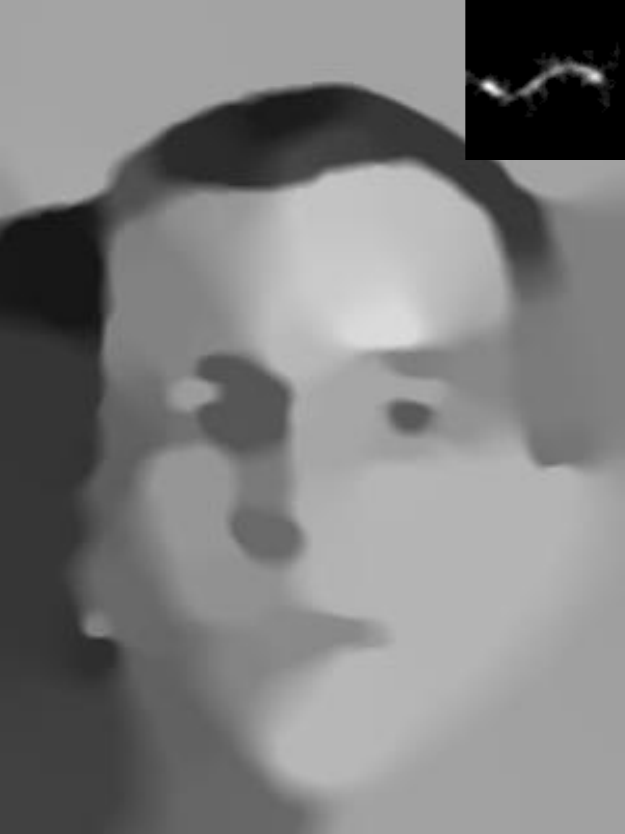} & \hspace{-0.5cm}
			\includegraphics[width=0.19\linewidth, height=0.24\linewidth]{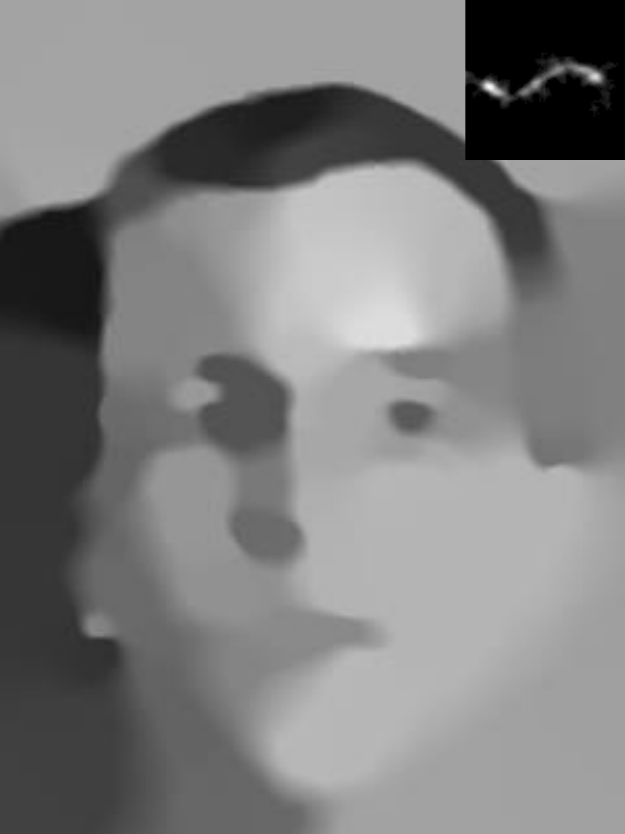} & \hspace{-0.5cm}
			\includegraphics[width=0.19\linewidth, height=0.24\linewidth]{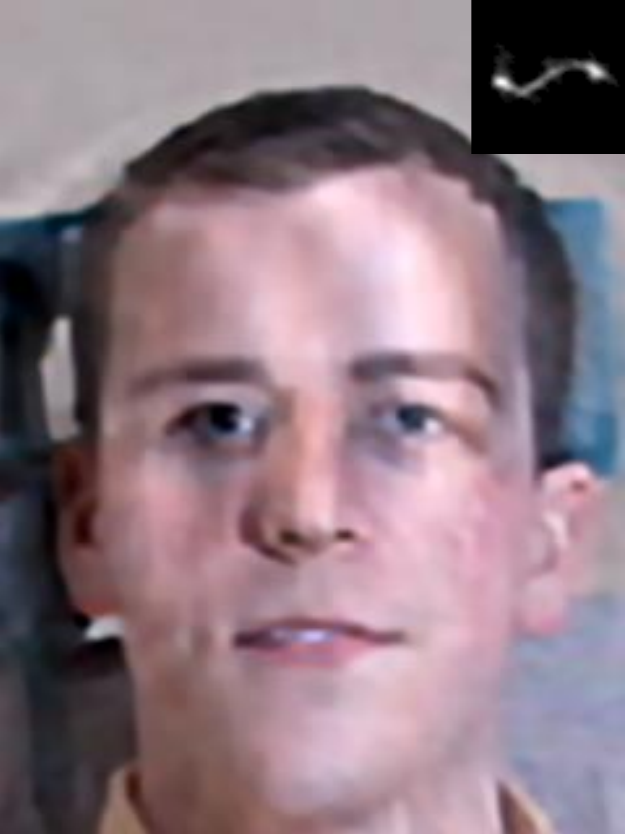} \\
			(k) & \hspace{-0.5cm}  (l) & \hspace{-0.5cm}  (m) & \hspace{-0.5cm} (n) & \hspace{-0.5cm} (o)\\
		\end{tabular}
	\end{center}
	\vspace{-0.2cm}
	\caption{
		Results without and with predicted salient edges $\nabla S$.
		(a)-(d) The 1st, 2nd, 5th, and 9th iteration intermediate
		results, respectively, using the edge selection method~\cite{Cho/et/al} to
		predict salient edges $\nabla S$ in
		Algorithm~\ref{alg:kernel-estimation-algorithm}.
		(e) Deblurred result with the edge selection method~\cite{Cho/et/al} to predict salient edges $\nabla S$ in
		Algorithm~\ref{alg:kernel-estimation-algorithm}.
		(f)-(i) The 1st, 2nd, 5th, and 9th iteration intermediate results, respectively,
		using the proposed exemplar-based method to predict salient edges $\nabla S$ in
		Algorithm~\ref{alg:kernel-estimation-algorithm}.
		(j) The deblurred result based on exemplars.
		(k)-(n) The 1st, 2nd, 5th, and 9th iteration intermediate results, respectively,
		using the proposed CNN-based method to predict salient edges $\nabla S$ in
		Algorithm~\ref{alg:kernel-estimation-algorithm}.
		(o) The deblurred result based on the proposed CNN. 
        The blurred image in this figure is the same as that of
		Fig.~\ref{fig:figure1}.
	}
	\label{fig: with-without-salient-edges}
\end{figure*}

\vspace{-3mm}
\subsection{Analysis and Discussion}
\vspace{-1mm}
\label{sec: Property Analysis}

In this section, we analyze the effect of the proposed edge prediction algorithms.
We show that proposed algorithms are not sensitive to
variation of dataset size, image noise, and parameters.
In addition, we discuss the limitations of the proposed algorithms.

\vspace{-2mm}
{\flushleft \bf Effect of predicted salient edges $\nabla S$.}
The initial predicted salient edges $\nabla S$ play a critical role in kernel estimation.
We use an example to demonstrate the effectiveness of the proposed
algorithm for predicting initial salient edges $\nabla S$.
Fig.~\ref{fig: with-without-salient-edges}(a)-(e) show that the deblurred results
using the edge selection method~\cite{Cho/et/al} contain artifacts as ambiguous edges are selected.
However, the proposed methods using the predicted facial structure
by exemplars (Fig.~\ref{fig: with-without-salient-edges}(f)-(j))
and the CNN (Fig.~\ref{fig: with-without-salient-edges}(k)-(o))
do not include ambiguous edges and thus estimate kernels better.
Fig.~\ref{fig: with-without-salient-edges}(f)-(i) and (k)-(n) also demonstrate that the
predicted salient edges $\nabla S$ by the proposed algorithms lead to fast convergence than the edge selection method~\cite{Cho/et/al}.

We note that the proposed algorithm does not require coarse-to-fine
kernel estimation strategies or heuristic edge selections.
The coarse-to-fine strategy can be viewed as the initialization for the finer levels,
which constrains the solution space and reduces the computational load.
Recent results of several state-of-the-art methods~\cite{Cho/et/al,Krishnan/CVPR2011,Xu/l0deblur/cvpr2013} show that
effective salient edges at the initial stage are important for kernel estimation.
If salient edges can be obtained effectively, it is not necessary to use coarse-to-fine
strategies or specific edge selection, thereby simplifying the kernel estimation process significantly.
Our exemplar-based and CNN-based methods operate on the input image of the original scale only and exploit the sharp structure information
to constrain the solution space.
By exploiting salient edges from the facial structures, the proposed methods perform well
without using coarse-to-fine strategies and achieve fast convergence.
In the method by Cho and Lee~\cite{Cho/et/al}, blur kernels are estimated in a
coarse-to-fine manner based on an heuristic edge selection strategy.
However, it is difficult to select salient edges from heavily blurred images without
exploiting any structural information (Fig.~\ref{fig: with-without-salient-edges}(a)).
Compared to the intermediate results using the $L_0$ prior (Fig.~\ref{fig:figure1}(f)),
our methods based on exemplars and CNN
restore the important facial components effectively
(Fig.~\ref{fig: with-without-salient-edges}(i) and (n)),
thereby facilitating kernel estimation and image restoration.

\vspace{-2mm}
{\flushleft \bf Robustness of exemplar structures.}
In the exemplar-based method, we use~\eqref{eq: normalized-cross-correlation} to find the best matched exemplar in the gradient space.
%
%
The matched exemplar should share similar, although not perfect, structural information
with the input image (e.g., Fig.~\ref{fig:figure1}(g)).
Furthermore, the shared structures should not contain numerous false salient edges caused by blur.
We also note that most mismatched contours caused by facial expressions correspond to the small gradients in the blurred images.
In such cases, these extraneous weak edges are not expected to help estimate kernels
according to the edge based methods~\cite{Cho/et/al,Xu/et/al}.
%
To alleviate this problem, we update exemplar edges iteratively (see Algorithm~\ref{alg:kernel-estimation-algorithm}) to increase
its reliability as shown in Fig.~\ref{fig: with-without-salient-edges}(f)-(i).
Consequently, the matched exemplars help estimate blur kernels and restore latent face images.

\begin{figure}[!t]\footnotesize
	\begin{center}
		\begin{tabular}{c}
			\includegraphics[width=0.95\linewidth]{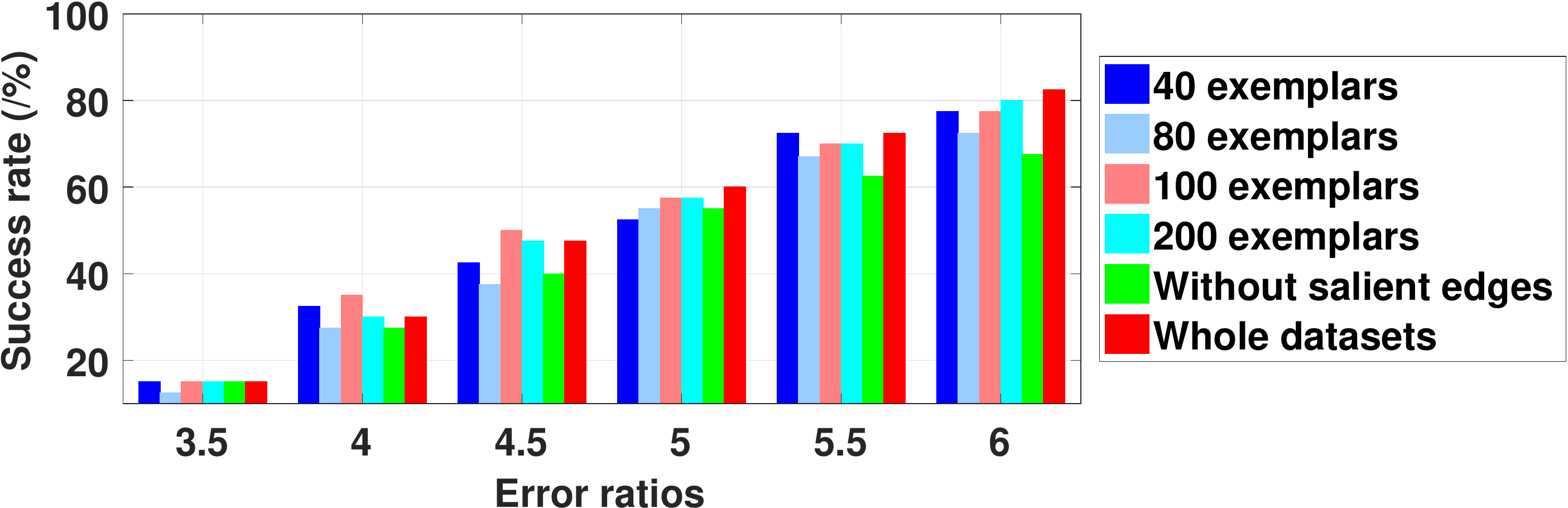} \\
			(a) Exemplar-based deblurring method\\
			\includegraphics[width=0.95\linewidth]{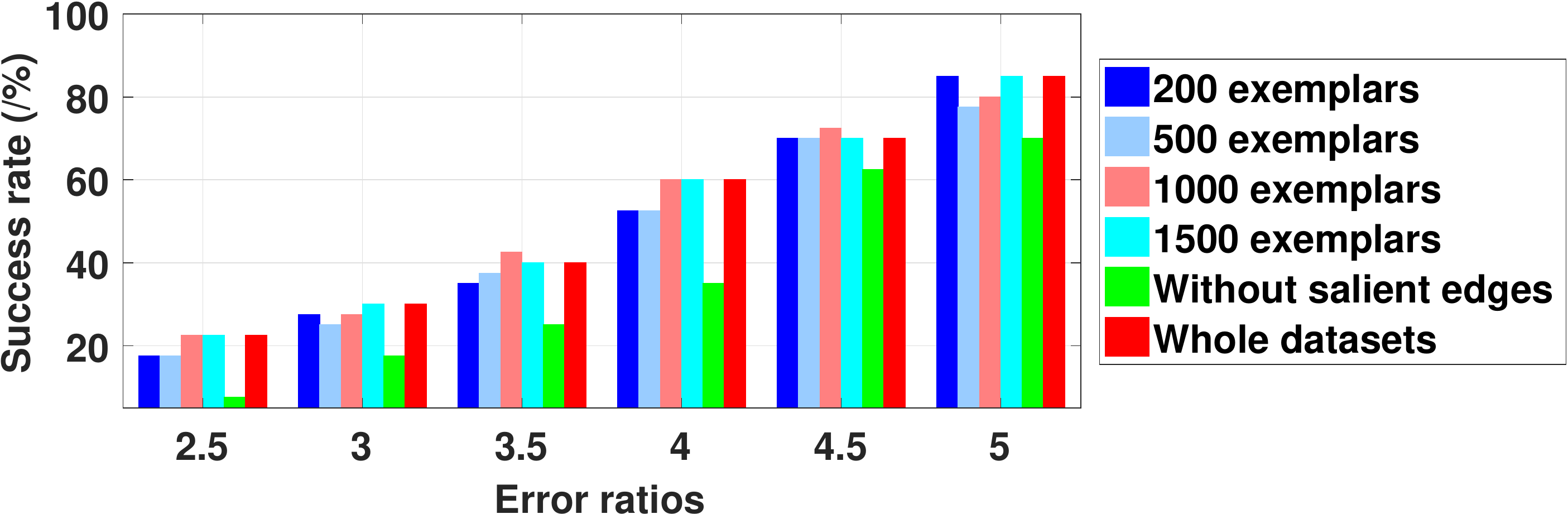} \\
			(b) CNN-based deblurring method
		\end{tabular}
	\end{center}
	\vspace{-0.2cm}
	\caption{Sensitivity analysis of dataset size.
	}
	\label{fig: size-dataset}
\end{figure}

\vspace{-2mm}
{\flushleft \bf Robustness to dataset size.}
Although a larger dataset is likely to contain more diverse exemplars
that facilitates finding the matching process by the proposed method,
the linear search time can be computationally expensive.
Empirically we show that blurry face images can be deblurred well when
coarse matches are available in a small exemplar set.
%
%
We apply the $k$-means clustering method to a set of face images,
and choose 40, 80, 100, and 200 centers as the exemplar datasets, respectively.
Similar to~\cite{Levin/CVPR2009}, we generate 40 blurred images consisting of 5
images (of different identities as the exemplars) with 8 blur kernels for experiments.
The cumulative error ratio~\cite{Levin/CVPR2009} is used to evaluate the method.
Fig.~\ref{fig: size-dataset}(a) shows that the proposed exemplar-based method
performs well with a small set of exemplars (e.g., 40).
With the increasing exemplar dataset size, the estimated results do not change significantly,
which demonstrates the robustness of the proposed method to exemplar dataset size.

To assess the sensitivity of the CNN-based structure prediction method,
we evaluate the proposed method with different numbers of exemplars.
We use 200, 500, 1000, and 1500 training images for the datasets, respectively.
For each clear image in these datasets,
we synthesize the blurred images using the generated kernels in Section~\ref{sec: Deep Learning Framework}.
We use the same 40 test images in the
exemplar-based method to evaluate the sensitivity of dataset size for the CNN-based method.
Fig.~\ref{fig: size-dataset}(b) shows that the proposed CNN-based method performs well
with a small set of exemplars (e.g., 200).
As the number of training images is increased, the performance of the proposed method does not change significantly,  especially when 1500 or all images are used.
The results show the proposed CNN-based method performs robustly against different dataset size.

\vspace{-2mm}
{\flushleft \bf Robustness to noise.}
If the blurred image contains large noise, edge selection~\cite{Cho/et/al,Xu/et/al} and
other state-of-the-art methods (e.g.,~\cite{Levin/CVPR2011,Krishnan/CVPR2011,Xu/l0deblur/cvpr2013}) may not perform well for kernel estimation.
However, the proposed methods perform well in such cases
due to the robust matching criterion
(see analysis in Section~\ref{ssec: Learning Framework}).
We show some examples in Section~\ref{sec: Experimental Results}.

\begin{figure*}[!t]\footnotesize
\begin{center}
\begin{tabular}{ccc}
\includegraphics[width = 0.28\linewidth]{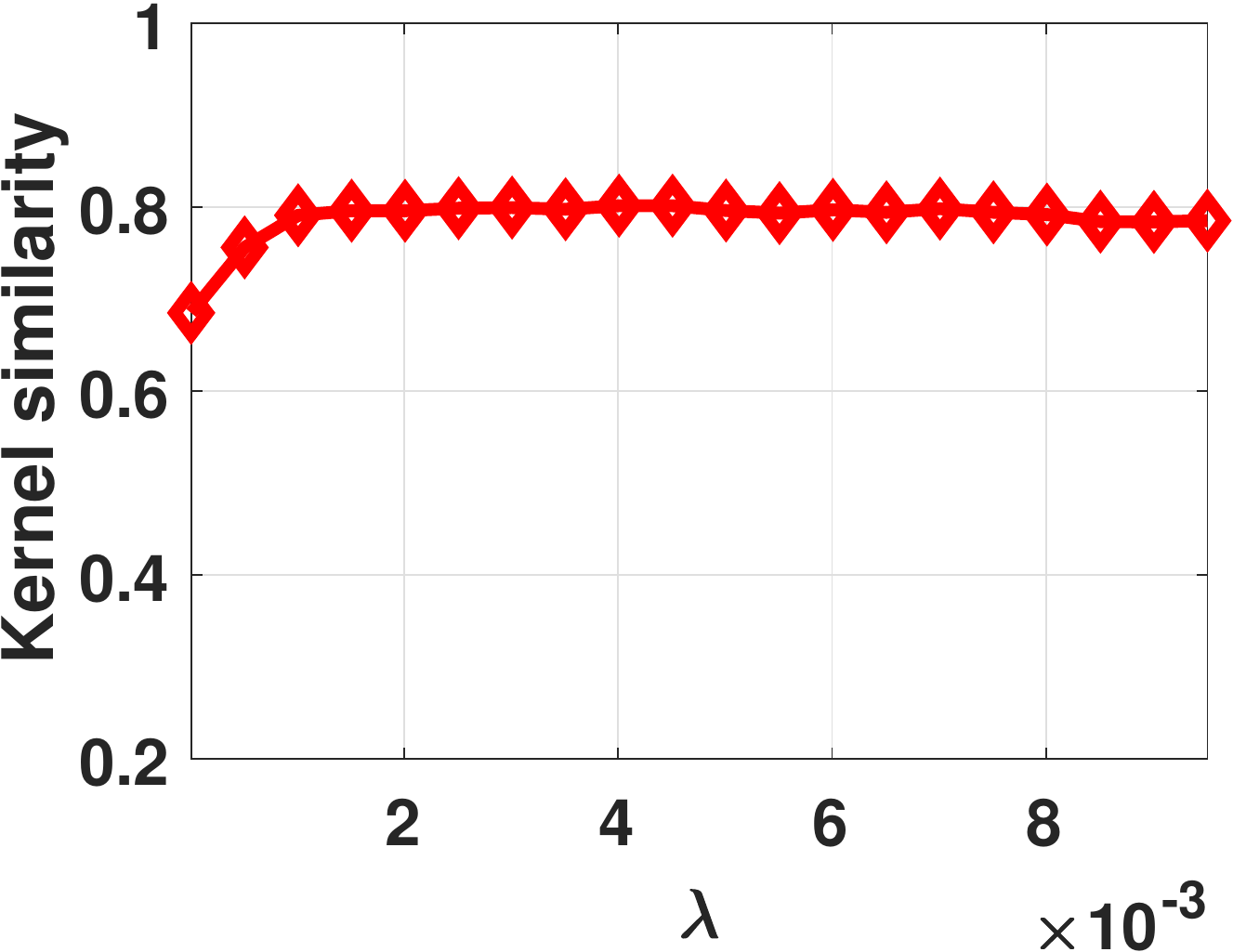} & \hspace{-0.2cm}
\includegraphics[width = 0.28\linewidth]{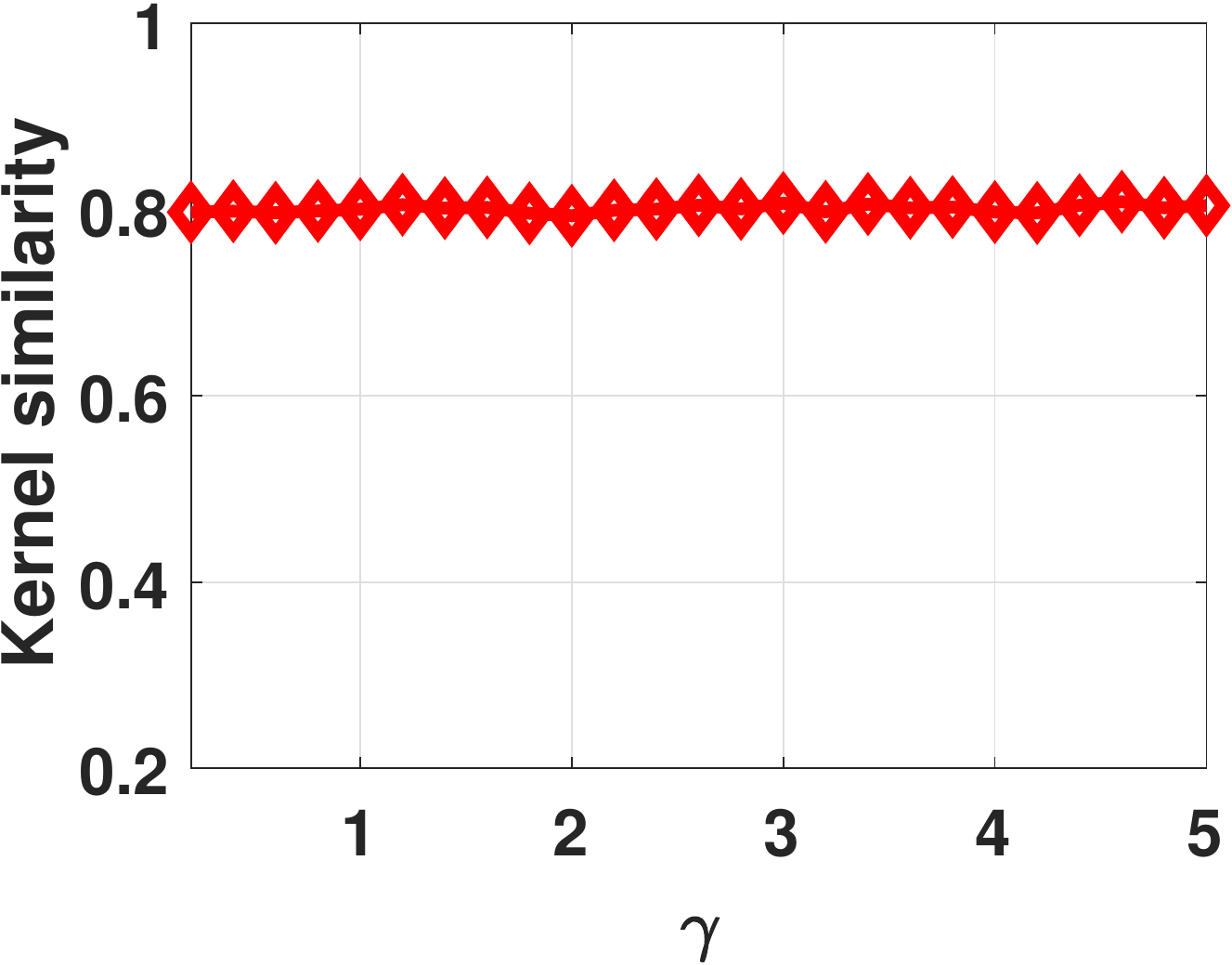} & \hspace{-0.2cm}
\includegraphics[width = 0.28\linewidth]{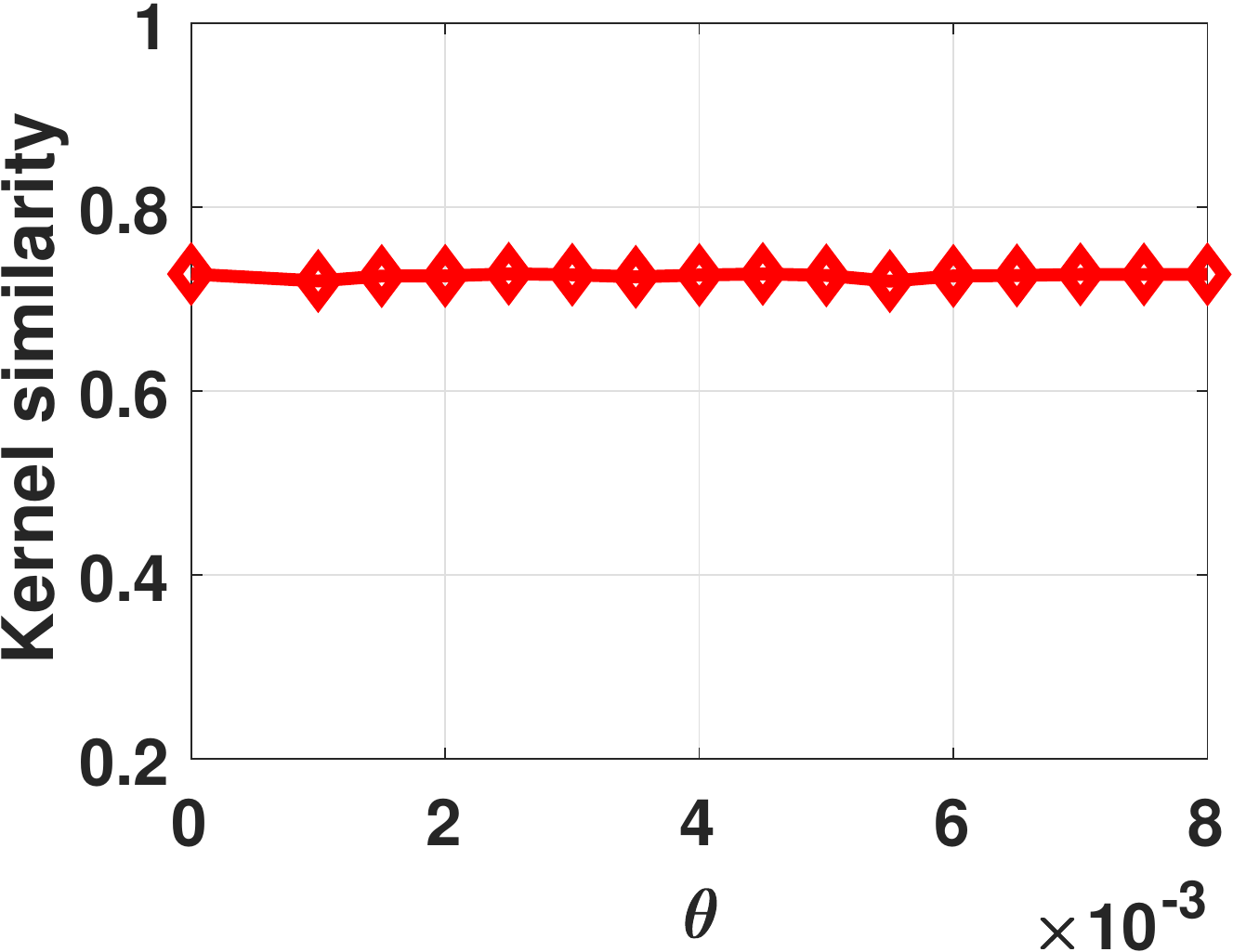}\\
 (a) &\hspace{-0.2cm} (b)  &\hspace{-0.2cm} (c)  \\
\end{tabular}
\end{center}
\vspace{-0.3cm}
\caption{Sensitivity analysis with respect to parameters $\lambda$, $\gamma$, and $\theta$.
}
\label{fig: paramter-analysis}
\end{figure*}

\vspace{-2mm}
{\flushleft \bf Parameter analysis.}
The proposed deblurring model involves three main parameters $\lambda$, $\gamma$, and $\theta$.
We evaluate the effects of these parameters on image deblurring using the dataset with 32 blurred images.
For each parameter, we carry out experiments with different settings by varying one and fixing the others using the kernel similarity metric to measure accuracy of estimated kernels.
Fig.~\ref{fig: paramter-analysis} shows the proposed deblurring algorithm is insensitive to parameter settings.

\begin{table*}[!t]
	\caption{Effect of the filter size in the first layer on image deblurring.}
	\centering
\vspace{-3mm}
\begin{tabular}{cccccccc}
\toprule
\backslashbox{ Filter size}{ Kernel size} & $9\times9$ & $13\times13$ & $15\times15$ & $17\times17$ & $19\times19$ & $21\times21$ & $27\times27$\\
\midrule
		$9\times 9$   & 32.84 & 29.99 & 33.33 & 32.58 & 22.03 & 26.60 & 22.84\\
		$13\times 13$ & 32.58 & 29.21 & 33.60 & 32.60 & 22.19 & 26.00 & 22.93\\
		$15\times 15$ & 33.11 & 29.56 & 32.99 & 32.49 & 22.01 & 26.48 & 24.09\\
		$17\times 17$ & 33.51 & 29.14 & 33.01 & 32.16 & 21.78 & 25.92 & 23.58\\
		$19\times 19$ & 33.11 & 29.72 & 33.48 & 33.01 & 22.16 & 25.75 & 23.89\\
		$21\times 21$ & 32.83 & 29.06 & 32.98 & 32.14 & 21.51 & 25.83 & 23.58\\
\bottomrule
	\end{tabular}
	\label{tab: filter-size}
\end{table*}

\begin{figure}[!t]\footnotesize
\begin{center}
\begin{tabular}{ccccc}
\hspace{-0.4cm}
\includegraphics[width = 0.20\linewidth]{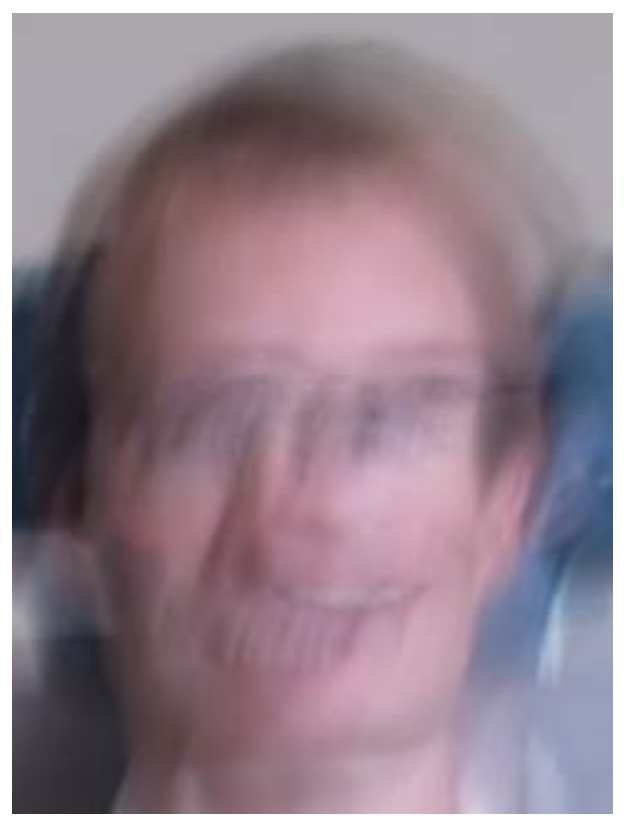} & \hspace{-0.5cm}
\includegraphics[width = 0.20\linewidth]{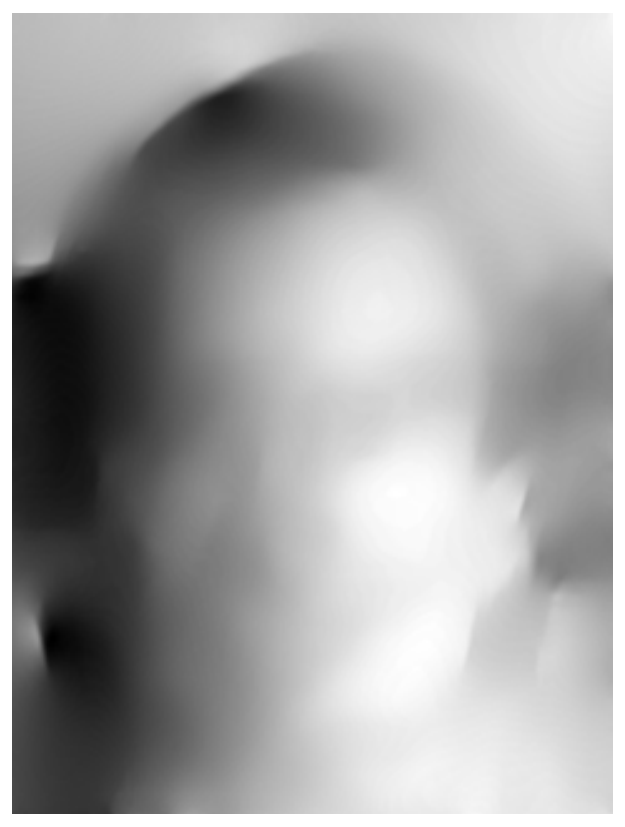} & \hspace{-0.5cm}
\includegraphics[width = 0.20\linewidth]{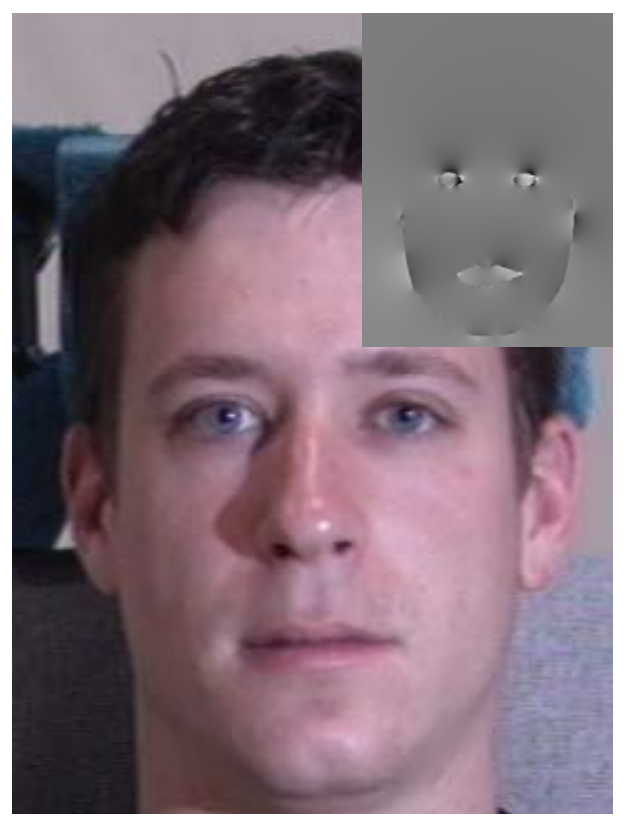} & \hspace{-0.5cm}
\includegraphics[width = 0.20\linewidth]{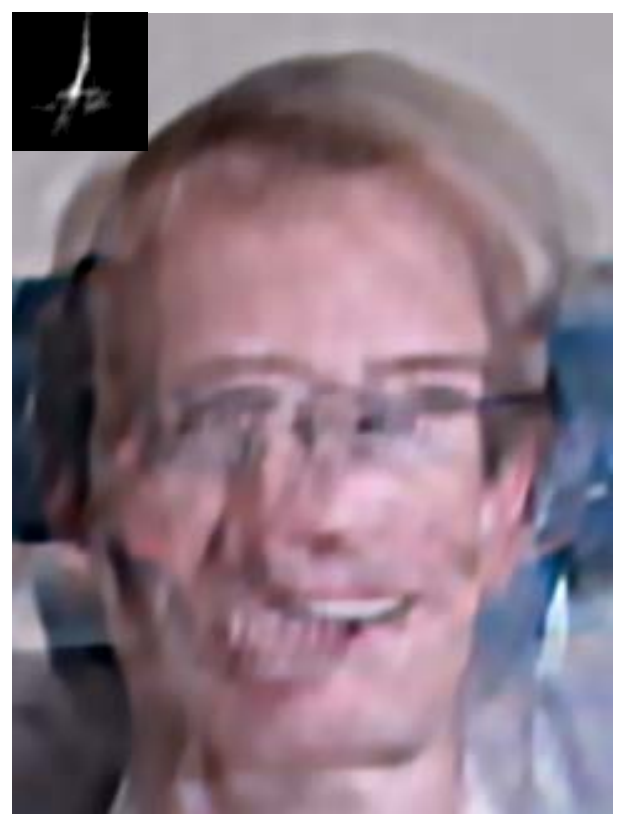} & \hspace{-0.5cm}
\includegraphics[width = 0.20\linewidth]{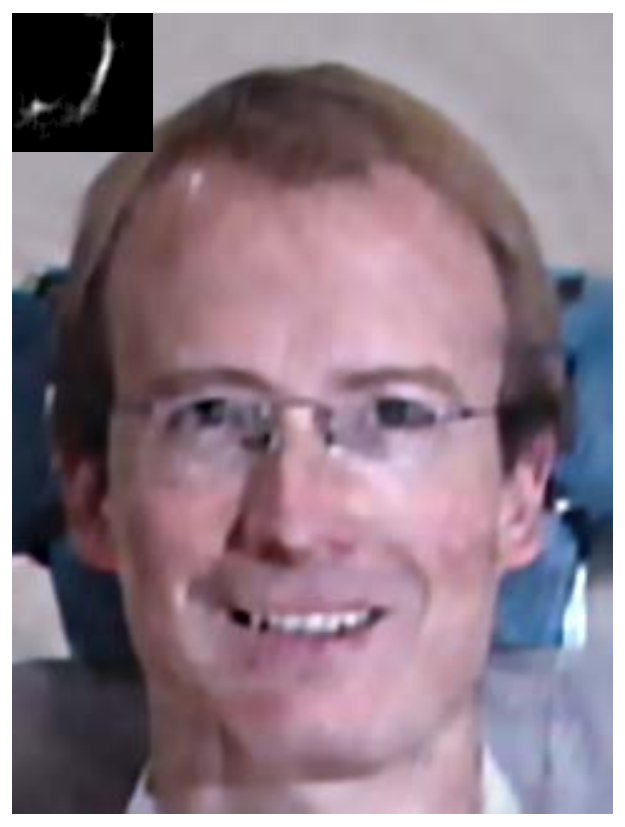} \\
 (a) &\hspace{-0.5cm} (b)  &\hspace{-0.5cm} (c)  &\hspace{-0.5cm} (d) &\hspace{-0.5cm} (e) \\
\end{tabular}
\end{center}
\vspace{-0.4cm}
\caption{The proposed CNN-based method is not able to handle the image with large blur.
(a) Blurred image.
(b) $\nabla S$ by CNN.
(c) Exemplar-based $\nabla S$.
(d) Results by the CNN-based method.
(e) Results by the exemplar-based method.
}
\label{fig: limitation-examples}
\end{figure}

In the proposed network, we note that the filter size of the first layer plays an
important role for predicting sharp edges from blurred images.
We evaluate the effect of this parameter on the proposed test dataset and use the PSNR as the metric in Table~\ref{tab: filter-size} .
Table~\ref{tab: filter-size} demonstrates that the proposed model is insensitive to
filter size change within a certain range.
%

%
%
Note that we use a large filter size in the first layer of the proposed network.
It is interesting to analyze
when the filter sizes of all the convolution layer are the same, e.g., the widely used setting, $3\times 3$ pixels.
For fair comparisons, we use the the same receptive field in the network and train it on the same training dataset.
Table~\ref{tab: filter-size-3-3} demonstrates that the network with this setting generates similar results to the proposed
method using the frontal face dataset.

\begin{table}[!t]
	\caption{Effect of the different filter size settings in the convolution layer of the proposed network. The results are generated by networks whose the receptive fields are the same.}
	\centering
\vspace{-3mm}
\begin{tabular}{cccc}
		\toprule
		Filter size settings & Proposed setting & Filter size with $3\times 3$\\
		\midrule
		Avg. PSNR & $35.33$ & $35.55$ \\
		\bottomrule
	\end{tabular}
	\label{tab: filter-size-3-3}
\end{table}

\vspace{-2mm}
{\flushleft \bf Limitations.}
%
As mentioned in Section~\ref{sec: synthetic frontal}, the exemplar-based edge prediction method is time-consuming (see Table~\ref{tab: run time}) and does not deblur face images well
when the main components cannot be extracted, e.g., profile faces.
Furthermore, it is not able to deblur generic images where the salient structures cannot be extracted.
Although the CNN-based method is more efficient and able to handle profile faces,
it is not able to handle the blurred images with large blur.
Fig.~\ref{fig: limitation-examples} shows an example where the main structures are severely blurred.
It is difficult for the CNN-based method to predict salient edges from such blurred images
(see Fig.~\ref{fig: limitation-examples}(b)), while the exemplar-based method
performs better in such cases with the help of exemplars (with the same pose).
%

\vspace{-3mm}
\section{Conclusions}
\vspace{-1mm}
\label{sec: Conclusion}
We propose an exemplar-based deblurring algorithm for face images that exploits the structural
information.
The proposed method uses facial structures and reliable edges
from exemplars for kernel estimation
without resorting to complex edge predictions.
Our method generates good initialization without using coarse-to-fine optimization strategies to enforce
convergence, and performs well when the blurred images do not contain
rich texture.
In addition, we further propose a
CNN-based deblurring method
which can effectively predict the sharp structure from a blurred input in real time.
Extensive evaluations with state-of-the-art deblurring
methods show that the proposed algorithms are effective
for deblurring face images.
%
%
We also show that that proposed methods can be applied to other object deblurring.


\vspace{-3mm}
\bibliographystyle{IEEEtran}
\bibliography{face-deblur}
\vspace{-15mm}
\begin{IEEEbiography}[{\includegraphics[width=1in,height=1.25in,clip,keepaspectratio]{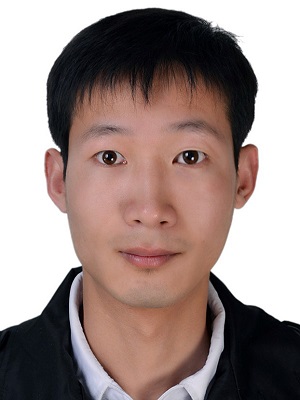}}]{Jinshan
    Pan} is a professor of School of Computer Science and Engineering, Nanjing University of Science and Technology.
    He received the Ph.D. degree in computational mathematics from the Dalian University of Technology, China, in 2017.
    He was a joint-training Ph.D. student in School of Mathematical Sciences at
    and Electrical Engineering and Computer Science at University of California, Merced, CA, USA from 2014 to 2016.
    His research interest includes image deblurring, image/video analysis and enhancement, and related vision problems.
\end{IEEEbiography}
\vspace{-15mm}
\begin{IEEEbiography}[{\includegraphics[width=1in,height=1.25in,clip,keepaspectratio]{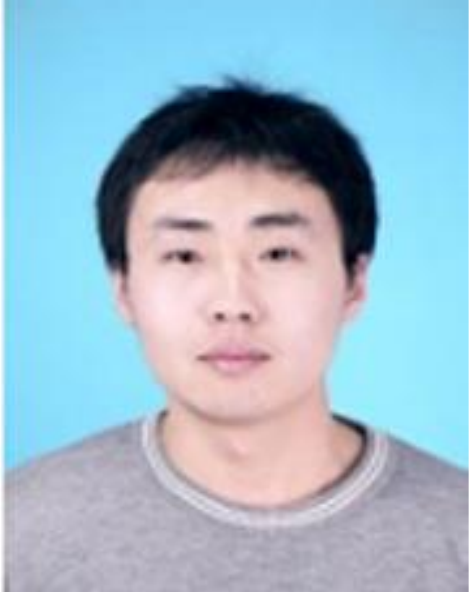}}]{Wenqi Ren} is an assistant professor in Institute of Information Engineering, Chinese Academy of Sciences, China. He received his Ph.D degree from Tianjin University in 2017. During 2015 to 2016, he was a joint-training Ph.D. student in Electrical Engineering and Computer Science at the University of California, Merced, CA, USA.
	His research interest includes image/video analysis and enhancement, and related vision problems.
\end{IEEEbiography}
\vspace{-15mm}
\begin{IEEEbiography}[{\includegraphics[width=1in,height=1.25in,clip,keepaspectratio]{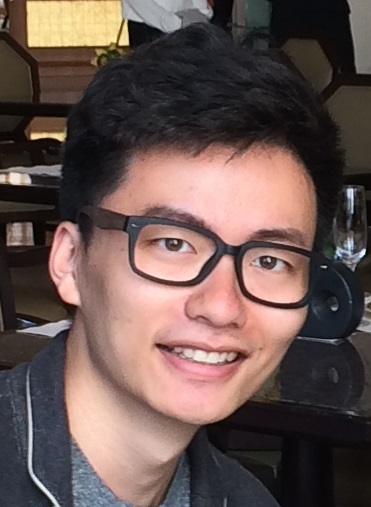}}]{Zhe Hu} is a research scientist at Hikvision. He was a research scientist at Light Labs Inc. from 2015 to 2017. He received the Ph.D. degree in Computer Science from University of California, Merced in 2015, and B.S. degree in Mathematics from Zhejiang University in 2009.
His research interests include computer vision, computational photography and image processing.\end{IEEEbiography}
\vspace{-15mm}
\begin{IEEEbiography}[{\includegraphics[width=1in,height=1.25in,clip,keepaspectratio]{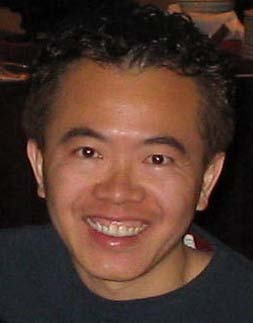}}]{Ming-Hsuan Yang}
is a professor of
Electrical Engineering and Computer Science with
the University of California, Merced, CA, USA. He
received the Ph.D. degree in computer science from
the University of Illinois at Urbana-Champaign,
USA, in 2000.
%
%
%
%
He received the NSF CAREER Award
in 2012,
and the Google Faculty Award in 2009. He is a senior member of the
IEEE and ACM.
\end{IEEEbiography}

%
%
%
%
%
%
%
%
%
%
\end{document}